%% file: main.tex
\documentclass[11pt,a4paper,google]{google}
\usepackage[authoryear,sort&compress,round]{natbib}
\input{math_commands.tex}

\usepackage{float}
\usepackage{wrapfig}
\usepackage{pdflscape}
\usepackage{multirow}
\usepackage{tcolorbox}
\tcbuselibrary{skins,breakable}
\usepackage{placeins}
\usepackage{etoolbox}
\usepackage{fontawesome5}
\newcommand{\projecturl}{https://stream-vibe.github.io/}
\newcommand{\projectlink}{%
  \ifdefempty{\projecturl}{\textcolor{blue!65!black}{\faGlobe\enspace Project page}\enspace\textcolor{gray}{(coming soon)}}%
  {\faGlobe\enspace\href{\projecturl}{\textcolor{blue}{\ttfamily\bfseries stream-vibe.github.io}}}}
\renewcommand{\absfont}{\bfseries\linespread{1.12}\fontsize{10}{12}\selectfont}
\newcolumntype{Y}{>{\raggedright\arraybackslash}X}
\newcolumntype{P}[1]{>{\raggedright\arraybackslash}p{#1}}

\hypersetup{colorlinks=true,linkcolor=blue!65!black,citecolor=blue!65!black,urlcolor=blue!65!black,
 pdftitle={StreamVibe},pdfauthor={Chengzhi Liu, Yuxiang Lai, Yuzhe YANG, Yanqing Liu, Jiheng Hou, Yue Cao, Haonan Wang, Ruixi Fan, Lan Wang, Sheng Liu, Songyou Peng, Yale Song, Kevin Qinghong Lin, Xin Eric Wang, Jui-Hsin Lai, Youbao Tang}}
\newtcolorbox{paperprompt}[1]{enhanced,breakable,
  colback=black!3,colframe=black!48,colbacktitle=black!48,coltitle=white,
  title={#1},title after break={#1 (continued)},fonttitle=\small\bfseries,fontupper=\fontsize{9.5}{12}\selectfont,
  boxrule=0.6pt,arc=2pt,left=10pt,right=10pt,top=9pt,bottom=9pt,
  before skip=10pt,after skip=12pt,parbox=false,
  before upper={\raggedright\interlinepenalty=10000\setlength{\parskip}{6pt}\setlength{\parindent}{0pt}}}
\renewcommand{\topfraction}{0.92}
\renewcommand{\bottomfraction}{0.85}
\renewcommand{\textfraction}{0.08}
\renewcommand{\floatpagefraction}{0.72}
\makeatletter
\let\StreamVibeNatLinkStart\hyper@natlinkstart
\let\StreamVibeNatLinkEnd\hyper@natlinkend
\renewcommand{\hyper@natlinkstart}[1]{%
  \leavevmode\hbox\bgroup\StreamVibeNatLinkStart{#1}}
\renewcommand{\hyper@natlinkend}{\StreamVibeNatLinkEnd\egroup}
\renewcommand{\hyper@natlinkbreak}[2]{%
  \hyper@natlinkend#1\hyper@natlinkstart{#2}}
\makeatother
\makeatletter
\def\wrapfill{\par
  \ifx\parshape\WF@fudgeparshape
    \nobreak
    \ifnum\c@WF@wrappedlines>\@ne
      \advance\c@WF@wrappedlines\m@ne
      \vskip\c@WF@wrappedlines\baselineskip
      \global\c@WF@wrappedlines\z@
    \fi
    \allowbreak
    \WF@finale
  \fi}
\makeatother
\definecolor{cvprblue}{RGB}{33,90,168}
\definecolor{gblue}{RGB}{104,157,246}
\definecolor{gred}{RGB}{238,105,93}
\definecolor{gyellow}{RGB}{252,201,55}
\definecolor{ggreen}{RGB}{93,185,117}
\DeclareRobustCommand{\svname}{\texorpdfstring{\textbf{\textit{%
  \textcolor{gblue}{S}\textcolor{gred}{t}\textcolor{gyellow}{r}\textcolor{gblue}{e}\textcolor{ggreen}{a}\textcolor{gred}{m}%
  \textcolor{gblue}{V}\textcolor{ggreen}{i}\textcolor{gred}{b}\textcolor{gyellow}{e}}}}{StreamVibe}}
\definecolor{tableheader}{RGB}{238,240,243}
\definecolor{tabletolerance}{RGB}{224,229,235}
\definecolor{tableclosed}{RGB}{252,241,234}
\definecolor{tableopen}{RGB}{234,243,250}
\definecolor{tableours}{RGB}{234,246,237}
\definecolor{metricshade}{RGB}{232,242,252}
\definecolor{benchheader}{RGB}{247,248,250}
\definecolor{benchstripe}{HTML}{F8FAFC}
\definecolor{benchours}{RGB}{235,245,255}
\definecolor{benchpartial}{RGB}{91,157,218}
\definecolor{titleblue}{HTML}{669DF6}
\definecolor{titlered}{HTML}{EF6C63}
\definecolor{titleyellow}{HTML}{FCC934}
\definecolor{titlegreen}{HTML}{5BB974}
\newcommand{\benchcmark}{\textcolor{green!60!black}{\ding{51}}}
\newcommand{\benchxmark}{\textcolor{red!75!black}{\ding{55}}}
\newcommand{\benchpmark}{\textcolor{benchpartial}{\raisebox{0.05ex}{\scalebox{0.9}{$\blacktriangle$}}}}
\newcommand{\rqicon}[1]{\raisebox{-0.32em}{\includegraphics[width=1.35em]{#1}}}
\newcommand{\tabgroup}[2]{\rowcolor{#1}\multicolumn{19}{@{}l@{}}{\rule{0pt}{2.25ex}\hspace{0.4em}\textit{#2}}}

\newcommand{\eg}[1]{\textit{e.g.,}}
\newcommand{\ie}[1]{\textit{i.e.,}}

\title{\texorpdfstring{%
  \leavevmode\raisebox{-0.59\height}[0pt][0pt]{%
    \includegraphics[width=2.00cm]{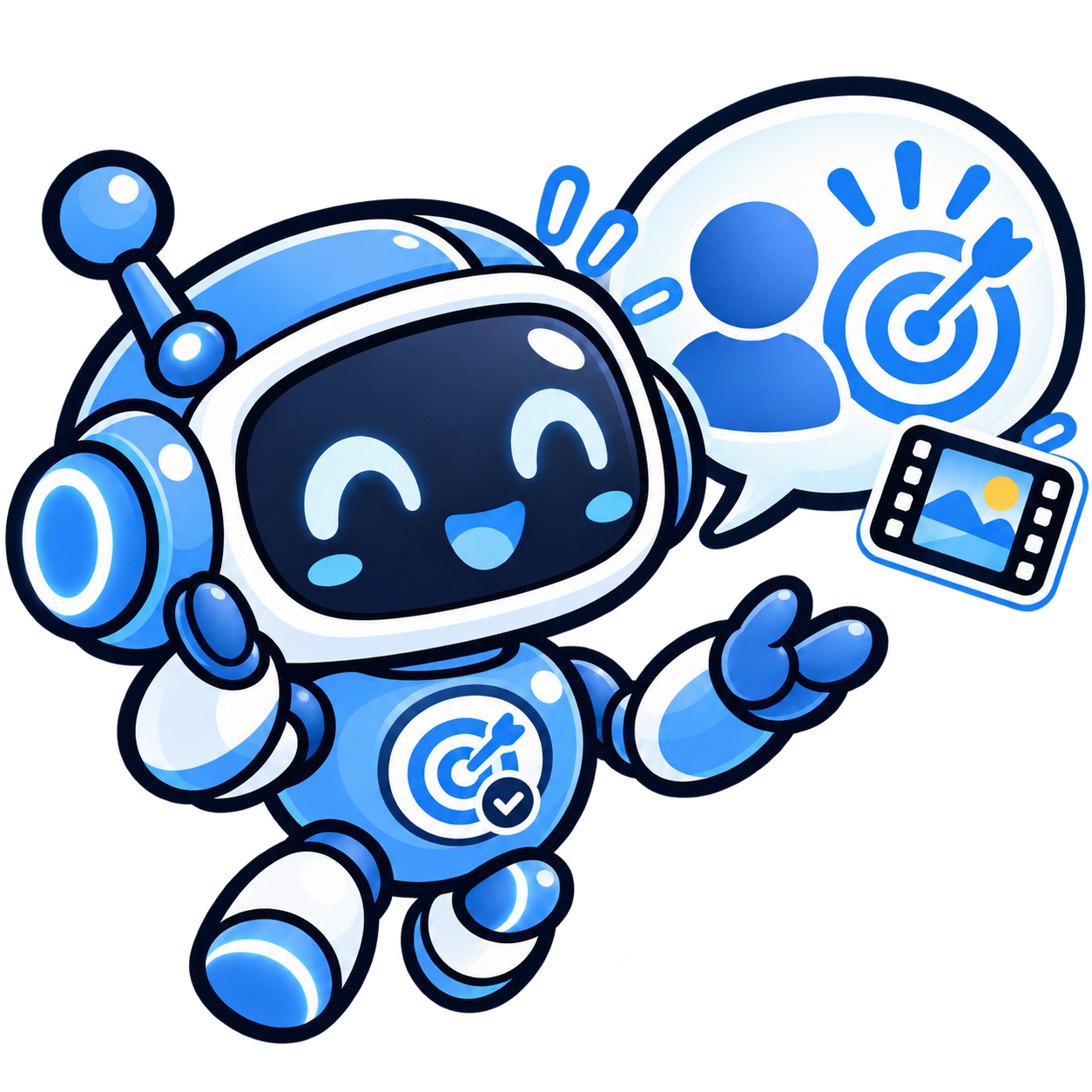}}%
  \hspace{0.35em}%
  \parbox[t]{\dimexpr\linewidth-2.00cm-0.35em\relax}{%
    \svname: Proactive Video Agents for Evolving User Intent}%
}{StreamVibe: Proactive Video Agents for Evolving User Intent}}
\author[1,2,*]{Chengzhi Liu}
\author[1,6]{Yuxiang Lai}
\author[2]{Yuzhe Yang}
\author[3]{Yanqing Liu}
\author[2]{Jiheng Hou}
\author[2]{Yue Cao}
\author[7]{Haonan Wang}
\author[1]{Ruixi Fan}
\author[1]{Lan Wang}
\author[4]{Sheng Liu}
\author[1]{Songyou Peng}
\author[1]{Yale Song}
\author[5]{Kevin Qinghong Lin}
\author[2]{Xin Eric Wang}
\author[1]{Jui-Hsin Lai}
\author[1]{Youbao Tang}
\newsavebox{\googleaffilbox}
\sbox{\googleaffilbox}{\includegraphics[height=8pt,width=10pt,keepaspectratio]{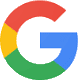}}
\newsavebox{\ucsbaffilbox}
\sbox{\ucsbaffilbox}{\includegraphics[height=8pt,width=10pt,keepaspectratio]{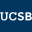}}
\newsavebox{\emoryaffilbox}
\sbox{\emoryaffilbox}{\includegraphics[height=8pt,width=10pt,keepaspectratio]{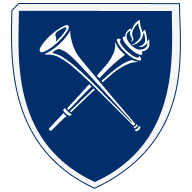}}
\newsavebox{\ucscaffilbox}
\sbox{\ucscaffilbox}{\includegraphics[height=8pt,width=10pt,keepaspectratio]{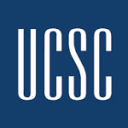}}
\newsavebox{\stanfordaffilbox}
\sbox{\stanfordaffilbox}{\includegraphics[height=8pt,width=10pt,keepaspectratio]{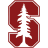}}
\newsavebox{\oxfordaffilbox}
\sbox{\oxfordaffilbox}{\includegraphics[height=8pt,width=10pt,keepaspectratio]{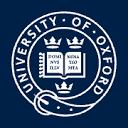}}
\newsavebox{\columbiaaffilbox}
\sbox{\columbiaaffilbox}{\includegraphics[height=8pt,width=10pt,keepaspectratio]{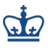}}
\DeclareRobustCommand{\affillogo}[1]{\raisebox{-1.5pt}{\usebox{\csname #1affilbox\endcsname}}\hspace{2pt}}
\affil[1]{\mbox{\protect\affillogo{google}Google}}
\affil[2]{\mbox{\protect\affillogo{ucsb}University of California, Santa Barbara}}
\affil[3]{\mbox{\protect\affillogo{ucsc}University of California, Santa Cruz}}
\affil[4]{\mbox{\protect\affillogo{stanford}Stanford University}}
\affil[5]{\mbox{\protect\affillogo{oxford}University of Oxford}}
\affil[6]{\mbox{\protect\affillogo{emory}Emory University}}
\affil[7]{\mbox{\protect\affillogo{columbia}Columbia University}}
\newcommand{\firstpagenote}{* This work was done while Chengzhi Liu was a Student Researcher at Google.}
\correspondingauthor{chengzhi@ucsb.edu}
\uselogo{}
\begin{abstract}
Vision-language models are shifting from offline passive understanding toward online proactive interaction. However, prevailing methods largely rely on explicit requests while overlooking implicit user intents (\eg, ``where should I click next?'' when a user is navigating a website) and fail to continuously update their understanding as user intents evolve.
These limitations make them less suitable for intent-aware interaction in real-world scenarios.
To address these challenges, we introduce \textbf{StreamVibe-Bench}, a new benchmark that serves as a testbed for user intent understanding and interaction, comprising 4.1K evaluation samples and 85K training samples spanning explicit, implicit, and dynamic intents.
Building on this benchmark, we develop \textbf{StreamVibe-VL}, a streaming proactive model that infers implicit user intents from continuous observations and dynamically adapts its interactions as these intents evolve.
We further introduce \textbf{StreamVibe-Harness}, which maintains user intents as persistent states with flexible lifecycles through hierarchical memory, enabling it to track the latest user intents and environmental states.
Our experiments reveal three key findings: \textit{(i)} StreamVibe-VL preserves strong general video understanding capabilities while improving response timing and answer quality by 10.7\% over existing baselines; \textit{(ii)} existing proactive models remain limited in understanding user intents and determining when to respond, especially under implicit and dynamic intents; and \textit{(iii)} StreamVibe-Harness further improves response consistency and adaptation during sustained interactions, resulting in a 12.3\% gain in proactive interaction performance.
\end{abstract}
\begin{document}
\setlength{\parindent}{0pt}
% Keep the larger overview on the title page by tightening title-block gaps.
\patchcmd{\maketitle}{\vskip11pt}{\vskip7pt}{}{}
\patchcmd{\maketitle}{\vskip20pt}{\vskip10pt}{}{}
% Lower only the title-page footer to use the available bottom margin.
\addtolength{\footskip}{10pt}
\maketitle\par
\vspace{6pt}
{\setlength{\parskip}{0pt}\centering\normalsize\projectlink\par}
\vspace{3pt}
\enlargethispage{44pt}

\begingroup
% Use the available space above the title-page footer for a larger overview.
\vspace{-0.4cm}
\setlength{\intextsep}{8pt}
\begin{figure}[H]
\centering
\captionsetup{font=small,skip=6pt}
\captionsetup{font=small}
\includegraphics[width=0.93\linewidth]{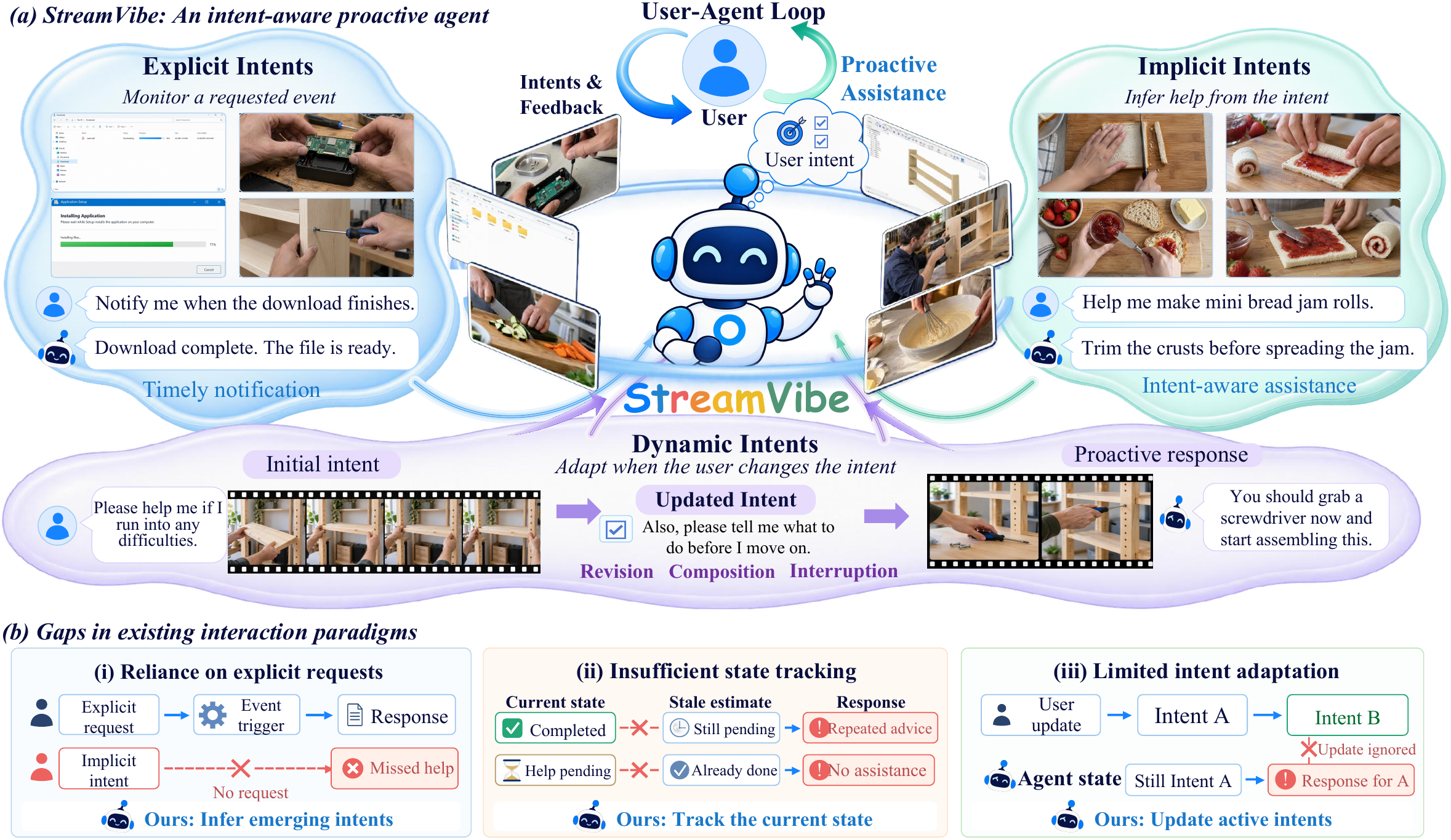}
\caption{\textbf{Overview of StreamVibe.} (a) StreamVibe studies sustained proactive interaction under explicit, implicit, and dynamically evolving user intents, where the agent continuously observes the environment and decides when and how to assist. (b) Existing paradigms rely on explicit intents, with limited tracking of interaction state and evolving intents. StreamVibe addresses these limitations through proactive streaming policy learning and persistent interaction state modeling that tracks task progress and updates active intents.}
\label{fig:intro}

\end{figure}
\endgroup
\clearpage
\addtolength{\footskip}{-10pt}

% Keep the original manuscript content and figures below.
\begingroup
% Keep natural spacing instead of stretching paragraphs to fill each page.
\raggedbottom
% Avoid splitting a hyphenated word across pages.
\brokenpenalty=10000
\setlength{\parskip}{5pt}
\titlespacing*{\section}{0pt}{16pt}{6pt}
\titlespacing*{\subsection}{0pt}{10pt}{4pt}
\titlespacing*{\subsubsection}{0pt}{8pt}{3pt}
\titlespacing*{\paragraph}{0pt}{6pt}{0.7em}
\section{Introduction}

% Opening epigraph: italic quotation and right-aligned attribution, as in Agent S.
\begingroup
\itshape
\noindent ``Mind is activity, a range of dynamic processes of interaction between organism and world.''\par
\nobreak\smallskip
\noindent\hfill --- Stanford Encyclopedia of Philosophy, John Dewey\par
\endgroup

% \kevin{Video is streaming. Thought is streaming. Intent emerges at their intersection. Videos stream. Minds do too.}
% Vision-language models~\citep{lu2026aura, wang2026mossvltechnicalreport} are advancing from offline content understanding~\citep{wang2025internvideo25,zhu2025internvl3} toward continuous  proactive assistance in dynamic\kevin{what does dynamic here means? is diverse?}environments~\citep{cui2026minicpmo,wang2025mmduet2,gander2026}. However, existing proactive interaction approaches are largely designed for explicit queries (“\kevin{example here}”), relatively independent\kevin{what independent means} requests~\citep{ li2026streampro, kundu2026planwatchrecover}, as illustrated in Figure~\ref{fig:intro}(b1). 

Vision-language models~\citep{lu2026aura, wang2026mossvltechnicalreport} are advancing from offline content understanding~\citep{wang2025internvideo25,zhu2025internvl3} toward continuous  proactive assistance in real-world streaming scenarios~\citep{cui2026minicpmo,wang2025mmduet2,gander2026, ding2025streammindunlockingframerate}. However, existing proactive interaction approaches are largely designed for explicit queries (e.g. “Notify me when a goal is scored.”), and isolated, single-turn requests~\citep{ li2026streampro, kundu2026planwatchrecover}, as illustrated in Figure~\ref{fig:intro}(i). 

However, these paradigm is less suited to real-world open-ended settings across three challenges, where \textit{(\textbf{i}) implicit user intents} (e.g. “Help me finish this task whenever I get stuck.”) must be inferred from ongoing observations and interactions.
Beyond identifying user intent, effective assistance requires
% tracking the evolving  interaction state~\citep{pu2025promemassist,tang2026proactiveservice}. 
\textit{(\textbf{ii}) task-relevant state tracking}.
As shown in Figure~\ref{fig:intro}(ii), prior interactions can change the environment and task progress, making a response that was appropriate earlier unnecessary or incorrect later. Although existing streaming models can retain long interaction histories~\citep{yao2026harnessingstreamingvideowild, wang2025streambridge}, they still struggle to maintain an up-to-date state, resulting in repeated, outdated, or missed responses.  
% User intents themselves also change over time~\citep{tack2026evolvingintent,tao2026idss}. As shown in Figure~\ref{fig:intro}(b3), later feedback may introduce new intents, revise existing constraints, or shift the current focus. Existing methods typically treat the initial request as fixed~\citep{liu2026guidememultidomaintaskguidance, joyai2026}, making it difficult to track continuously evolving user intent. 
Lastly, as shown in Figure~\ref{fig:intro}(iii), \textit{(\textbf{iii}) user intents is evolving }as the interaction unfolds~\citep{tack2026evolvingintent,tao2026idss, yang2026magevl}. Later feedback may introduce new goals, revise existing constraints, or shift the current focus, while existing methods often fail to update which intents remain relevant and instead continue to rely on earlier requests. These challenges make proactive interaction more than a problem of real-time long-context modeling: it requires the agent to continuously infer and update user intent while tracking the evolving interaction state. This motivates three key research questions forward “\textbf{Intent-aware interaction}”:

% Give the research-question panel a little more breathing room.
\enlargethispage{10pt}
\begin{tcolorbox}[colframe=cvprblue!35, colback=benchours, boxrule=1.5pt,
arc=2mm, top=2pt, bottom=2pt, left=4pt, right=4pt, boxsep=1pt,
before skip=12pt, after skip=12pt]
\fontsize{9.3}{10.8}\selectfont
\centering
\resizebox{\linewidth}{!}{%
\begin{tabular}{@{}l@{}}
\rqicon{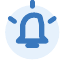}\hspace{0.45em}\textit{How can we  construct a testbed for proactive interaction with implicit user intent?} \\[0.5pt]
\rqicon{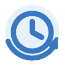}\hspace{0.45em}\textit{How can such intent-aware interaction be internalized as a native model capability?} \\[0.5pt]
\rqicon{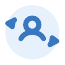}\hspace{0.45em}\textit{How to flexible manage context to track environmental and user intent changes?}
\end{tabular}%
}
\end{tcolorbox}

% \kevin{
% 1. how can we simulate this environment and evaluate it --> data systhsis + bench
% 2. how vlm can manage composition context as human memory --> harness pulggable, free-lunch, is a smart way, dont need to discard 
% 3. 
% }

In this work, we address these three questions through benchmarking, modeling, and agentic design.
Firstly, we introduce {StreamVibe-Bench}, a unified intent-aware benchmark for sustained proactive interaction with 4.1K evaluation samples and 85K training samples spanning {explicit}, {implicit}, and {dynamic} user intents.
Building on this benchmark, we further develop {StreamVibe-VL}, a streaming proactive model that infers implicit user intents and adapts its interactions as these intents evolve. Streaming supervised fine-tuning (SFT) jointly learns when and how to assist, while localized trajectory reinforcement learning (RL) focuses optimization on interaction-critical moments to improve response timing and quality. Lastly, we further develop {StreamVibe-Harness}, which maintains user intents as persistent, updatable states with flexible lifecycles, while hierarchical memory retrieves relevant history to keep subsequent decisions aligned with the current interaction context and user intent. Our main contributions are as follows:

\begin{itemize}[leftmargin=1.5em,itemsep=3pt,topsep=4pt,parsep=0pt,partopsep=0pt]
    % \item \kevin{to the best of our knowledge, we are the first work to study user intent xxx}We construct {StreamVibe-Bench}, a unified benchmark for user-centered proactive interaction, with 4.1K evaluation and 85K training samples spanning explicit, implicit, and dynamic intents.
    \item To the best of our knowledge, we are the first work to study user intents in proactive streaming interaction. To this end, we construct {StreamVibe-Bench}, a unified benchmark with 4.1K evaluation and 85K training samples spanning explicit, implicit, and dynamic intents.

    \item We introduce {StreamVibe-VL}, a proactive streaming video model that jointly learns when and how to assist under implicit and evolving user intents.

    \item We develop the {StreamVibe-Harness}, which maintains updatable intent states with flexible lifecycles and hierarchical memory for consistent behavior over sustained interaction.
\end{itemize}

\section{Related Work}
\setlength{\parskip}{5pt}

\paragraph{Streaming Video Benchmark.}
Early video benchmarks focus on offline understanding~\citep{fu2024videomme,zhao2025mmvu}, while online benchmarks evaluate sequential perception and response before video completion~\citep{li2025ovobench,huang2025ovbench}. Recent proactive benchmarks further require models to decide when to respond, often using annotated trigger points and extending to egocentric or task-guidance settings~\citep{wang2025proactivevideoqa,kundu2026planwatchrecover,ran2026egoprobench}. However, most focus on a single intent type or environment and evaluate responses independently. StreamVibe-Bench instead unifies explicit, implicit, and dynamic intents across physical and digital tasks, with both pointwise and trajectory evaluation.

\paragraph{Proactive Interaction Models.}
Online video-language models support continuous perception and low-latency responses~\citep{zeng2025streamforest,cui2026minicpmo}, while real-time and task-guidance systems further learn when to speak or provide assistance during ongoing interaction~\citep{gong2026vinci2,liu2026guideme,lu2026aura}. However, most existing approaches center on explicitly specified requests or task objectives, with limited support for tracking interaction state and adapting to evolving user intent over time~\citep{tack2026evolvingintent,tao2026idss}. StreamVibe-VL targets open-ended proactive interaction, where intents must be inferred from user goals, environment changes, and task progress, and updated throughout the interaction.

\paragraph{Memory and Agent Harness.} Agentic systems increasingly incorporate persistent memory~\citep{xu2025amem}, planning~\citep{xiong2025mpo}, reflection~\citep{yuan2025agentr}, and reusable skills~\citep{zheng2025skillweaver} to preserve and reuse historical context over long-horizon tasks. Recent memory systems further study how to retrieve, integrate, and abstract historical information into reusable experience for long-term interaction~\citep{yu2026agemem,chhikara2025mem0}. However, these methods mainly organize and access past information without explicitly tracking evolving user intents and task progress, which can cause proactive responses to drift from the current interaction state. StreamVibe-Harness addresses this gap by maintaining user intents and task progress as persistent, updatable interaction states and retrieving relevant history before each decision.

\enlargethispage{12pt}
\section{StreamVibe-Bench: Intent-aware Proactive Benchmark}

\begin{table}[!t]
\centering
\caption{\textbf{Comparison with other proactive interaction benchmarks.}
StreamVibe-Bench covers Explicit, Implicit, and Dynamic Intents across both physical and digital environments, and supports both Pointwise and Trajectory evaluation.
\protect\benchcmark, \protect\benchpmark, and \protect\benchxmark  denote full, partial, and no support, respectively.
Dur. denotes the average sample duration in minutes; Evid. and Traj. denote the numbers of evidence points and evaluation trajectories.}
\label{tab:benchmark_comparison}
\small
\begin{NoHyper}
\setlength{\aboverulesep}{0pt}
\setlength{\belowrulesep}{0pt}
\setlength{\tabcolsep}{4.5pt}
\renewcommand{\arraystretch}{1.20}
\makebox[\textwidth][c]{%
\resizebox{\textwidth}{!}{%
\begin{tabular}{lcccccccccc}
\toprule
\rowcolor{benchheader}
& \multicolumn{3}{c}{\textbf{User Intent Type}}
&
&
& \multicolumn{2}{c}{\textbf{Environment}}
& \multicolumn{2}{c}{\textbf{Eval Setting}}
& \\
\cmidrule(lr){2-4}\cmidrule(lr){7-8}\cmidrule(lr){9-10}
\rowcolor{benchheader}
\multirow{-2}{*}{\textbf{Benchmark}}
& \textbf{Explicit} & \textbf{Implicit} & \textbf{Dynamic}
& \multirow{-2}{*}{\textbf{Dur. (min)}}
& \multirow{-2}{*}{\textbf{Evid.}}
& \textbf{Physical} & \textbf{Digital}
& \textbf{Pointwise} & \textbf{Trajectory}
& \multirow{-2}{*}{\textbf{Traj.}} \\
\midrule
ProactiveVideoQA~\citep{wang2025proactivevideoqa}
& \benchcmark & \benchxmark & \benchxmark & 2.1 & 3.5K & \benchcmark & \benchxmark
& \benchxmark & \benchcmark & 1.4K \\
\rowcolor{benchstripe}
EgoProactive~\citep{kundu2026planwatchrecover}
& \benchpmark & \benchcmark & \benchxmark & 3.5 & 5.4K & \benchcmark & \benchxmark
& \benchcmark & \benchxmark & 700 \\
EgoPro-Bench~\citep{ran2026egoprobench}
& \benchcmark & \benchcmark & \benchxmark & 3.0 & 2.4K & \benchcmark & \benchxmark
& \benchxmark & \benchcmark & 2.4K \\
\rowcolor{benchstripe}
IPIBench~\citep{li2026ipibench}
& \benchcmark & \benchxmark & \benchcmark & 1.0 & 3.7K & \benchcmark & \benchxmark
& \benchcmark & \benchpmark & 1.8K \\
EgoServe~\citep{gong2026vinci2}
& \benchpmark & \benchcmark & \benchpmark & 6.2 & 3.4K & \benchcmark & \benchxmark
& \benchxmark & \benchcmark & 293 \\
\rowcolor{benchstripe}
GuideMe~\citep{liu2026guideme}
& \benchcmark & \benchcmark & \benchxmark & 5.5 & 47.8K & \benchcmark & \benchxmark
& \benchcmark & \benchpmark & 2.5K \\
GUIDE~\citep{yang2026guide}
& \benchcmark & \benchcmark & \benchpmark & 33.8 & 0.7K & \benchxmark & \benchcmark
& \benchcmark & \benchxmark & 120 \\
\midrule
\rowcolor{benchours}
\textbf{StreamVibe-Bench (Ours)}
& \benchcmark & \benchcmark & \benchcmark & \textbf{8.3} & \textbf{20.5K}
& \benchcmark & \benchcmark & \benchcmark & \benchcmark & \textbf{4.1K} \\
\bottomrule
\end{tabular}%
}
}
\end{NoHyper}
\end{table}

\begin{figure}[!t]
\centering
\captionsetup{skip=5pt}
\includegraphics[width=\linewidth]{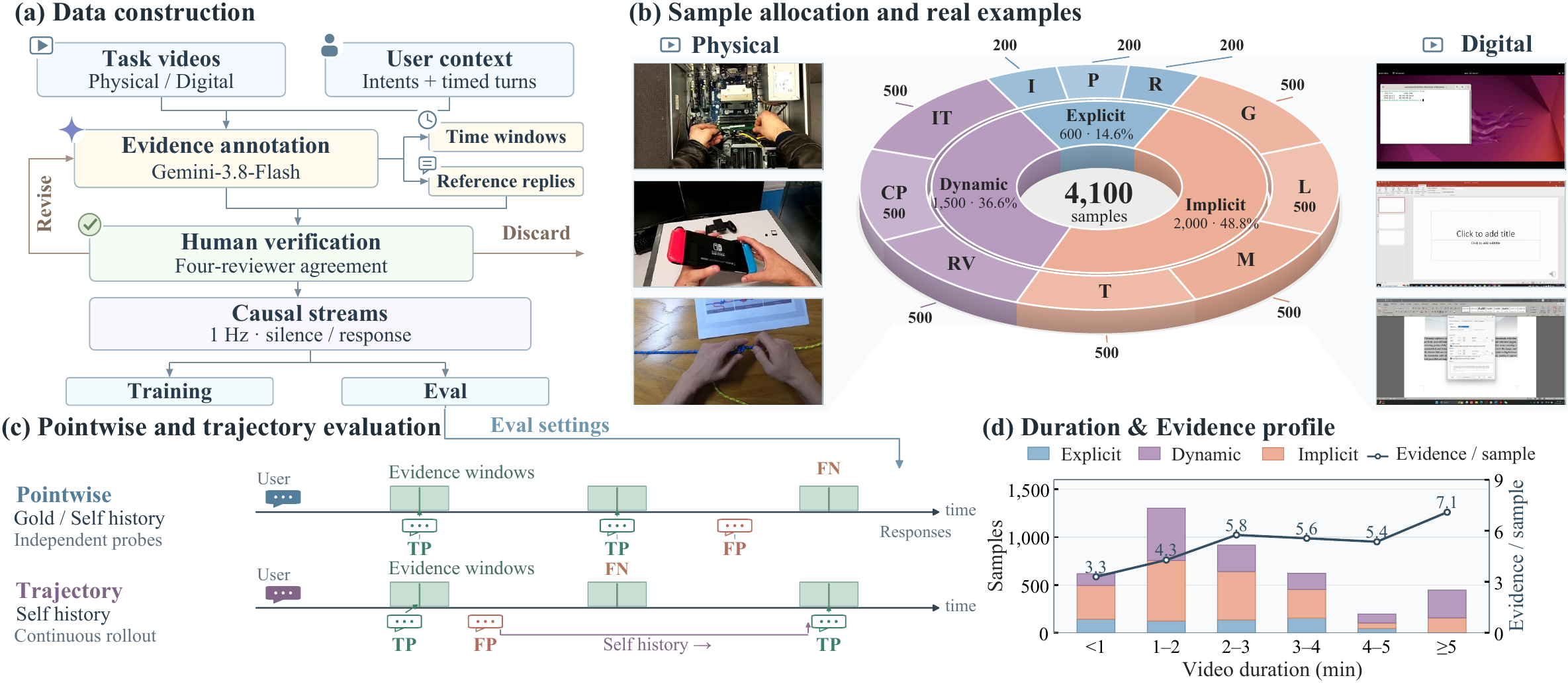}
\caption{\textbf{Construction and statistics of StreamVibe-Bench.}
(a) Data construction from task videos and user context to evidence annotation, human verification, and causal stream generation.
(b) Sample allocation across three requirement types, with examples from physical and digital environments.
(c) Pointwise and trajectory evaluation settings for local response decisions and continuous interaction behavior. TP: true positives; FP: false positives; FN: false negatives.
(d) Joint distribution of video duration, sample count, and average number of evidence points per sample.}
\label{fig:data}
\end{figure}

\subsection{Task Formulation}
\label{sec:problem-setup}
We formulate proactive assistance as a causal streaming decision process. The agent observes a video stream $\mathcal{V}=\{v_t\}_{t=1}^{T}$ while the user may express one or more intents $i_j$ over time, explicitly or implicitly, possibly before earlier intents are fulfilled. Each intent becomes available to the agent at time $\tau_j$. Let $\mathcal{I}_{\leq t}$ denote all intents available by time $t$, and $\mathcal{A}_t\subseteq\mathcal{I}_{\leq t}$ the currently active ones. At each timestep, the agent accesses only $c_t=\langle\mathcal{V}_{\leq t},\mathcal{I}_{\leq t},\mathcal{H}_{<t}\rangle$ and predicts $a_t=\pi(c_t)\in\{\textsc{silence},\textsc{response}\}$. If responding, it generates $y_t$ for some $i_j\in\mathcal{A}_t$, without access to future video or user inputs.

Unlike retrospective VideoQA, relevant evidence may emerge only after an intent is expressed. Each intent $i_j$ can correspond to one or more evidence points $\xi_{jk}$, each with a valid response window $[s_{jk},e_{jk}]$ and reference response $\hat{y}_{jk}$. An evidence point is covered if the agent responds within its window with a semantically correct response to the corresponding intent. An intent remains active until all its evidence points are covered. The agent must therefore determine both when to respond and what to respond, while avoiding premature, missed, repeated, or irrelevant responses.

\enlargethispage{18pt}
\subsection{Hierarchical Difficulty}
\label{sec:interaction-settings}

StreamVibe-Bench characterizes proactive interaction along two dimensions:
\textit{intent explicitness} and \textit{intent dynamics}.
Existing benchmarks primarily focus on \textbf{(i) Level 1: Explicit Intents}, where the user specifies a condition, event, or state to monitor, and the agent responds when it occurs, e.g. Let me know when the water starts boiling.

We further introduce \textbf{(ii) Level 2: Implicit Intents},
where the user provides only a high-level goal, requiring the agent to infer when and how to assist from ongoing observations and task progress,
e.g. Help me keep an eye on this dish.

Finally, \textbf{(iii) Level 3: Dynamic Intents} considers intents that evolve throughout the interaction.
As the task unfolds, the user may provide multiple new inputs, requiring the agent to continually update the currently valid task state.

We consider three common forms:
\textit{Revision (RV)}, which modifies an existing intent;
\textit{Composition (CP)}, which adds a new intent or constraint while preserving existing ones; and
\textit{Interruption (IT)}, which introduces an immediate request while the ongoing task remains active.
These changes may occur repeatedly and in combination within a single interaction.

\subsection{Benchmark Construction}

\paragraph{Task Collection.} We collect task videos from both physical and digital environments to cover two complementary forms of proactive interaction. In (i) \textit{physical} environments, assistance mainly relies on understanding continuous visual processes, including user activities, task progress, and emerging intents. In contrast, user intent in (ii) \textit{digital} environments is often implicit in interaction trajectories such as navigation, content editing, and tool use, requiring the agent to infer user goals, constraints, and task state from the current interface and interaction history. The two settings thus capture user intent through physical behavior and digital interaction, respectively, covering proactive assistance across different interaction environments. Specifically, physical tasks mainly include egocentric daily activities and task execution, while digital tasks focus on GUI and computer-use scenarios.

 % \kevin{scope: what make digital important? physical v.s. digital difference in user-mental udnerstanding}

% \textbf{Help Point Annotation.}
% \kevin{evidence}
% Given each video-request pair, we use Gemini-3.7-Flash~\citep{google2026gemini3} to clean the data and identify candidate help points. Each candidate includes an acceptable intervention window, a preferred intervention time, and a reference response. For tasks with Dynamic Intents, we additionally track how the active intents change after each user turn.

% \textbf{Human Verification.}  Four reviewers independently verify the user requests, assistance content, and intervention timing, correcting or removing candidates that lack sufficient visual or task evidence and retaining only samples approved by all four reviewers. The validated annotations are then aligned to 1\,FPS streams and converted into causal interaction sequences with timestamped user turns and per-second \textsc{silence}/\textsc{response} labels for supervised fine-tuning, reinforcement learning, and evaluation.

% Allow flexible line breaking around the long linked author name.
\begin{sloppypar}
\paragraph{Annotation and Verification.} For each video-request pair, we use Gemini-3.8-Flash \citep{google2026gemini3} to clean the data and identify candidate help points with response windows, preferred times, and reference answers. For dynamic intents, we further construct timestamped multi-turn interactions in which intents may be revised, extended, or interrupted as the task unfolds. Four reviewers then independently verify the user requests and response timing, correcting or removing samples without sufficient evidence and retaining only those approved by all reviewers. The validated annotations are aligned to 1 FPS streams and converted into causal interaction sequences for training and evaluation. Details are provided in Appendix~\ref{app:human-review}.
\end{sloppypar}

\subsection{Data Statistics}

\paragraph{Categories and Scale.} Building on the three proactive interaction settings in Section~\ref{sec:interaction-settings}, we organize StreamVibe-Bench into finer-grained task subtypes, as illustrated in Figure~\ref{fig:data}(b). The 600 \textit{Explicit Intents} samples span Single Event (I), Transition (P), and Recurring Event (R). For \textit{Implicit Intents}, 2,000 samples cover Guidance (G), Learning (L), Monitoring (M), and Tool (T). The remaining 1,500 \textit{Dynamic Intents} samples consist of Revision (RV), Composition (CP), and Interruption (IT). Subtypes are evenly distributed within each category. In total, StreamVibe-Bench contains 4,100 evaluation samples, 20,546 help points, and 183.4 hours of video. Additional statistics and detailed explanations are provided in the Appendix~\ref{app:dataset-details}.

\paragraph{Training Data Curations.} We construct the StreamVibe-85K training set, consisting of 70K samples for supervised fine-tuning and 15K tasks for reinforcement learning. During supervised fine-tuning, we further incorporate 768K general streaming-video samples from sources including TimeChat-Online~\citep{src_timechatonline} and VideoChat-Flash~\citep{src_videochatflash} to improve general streaming-video modeling. Detailed components are provided in Appendix~\ref{training}.

\subsection{Evaluation Protocol}
\label{protocol}
\paragraph{Settings.} As illustrated in Figure~\ref{fig:data}(c), two complementary evaluation settings are adopted to distinguish local proactive decision-making from sustained interaction behavior. 

\textit{\textbf{(i) Pointwise Evaluation}} samples positive and negative timestamps around annotated help points and independently predicts whether the model should respond. Both \textit{Gold History} and \textit{Self History} are considered, where the former isolates local decision ability and the latter measures the effect of model-generated history. Independent evaluation avoids error accumulation across time. 

\textit{\textbf{(ii) Trajectory Evaluation}} runs the model continuously over the full video stream, with each decision conditioned on prior observations and self-generated interaction history. This setting captures error propagation, requirement tracking, and repeated or missed responses. Together, the two settings evaluate local decisions under a given history and sustained interaction under self-generated history.

\paragraph{Metrics.} We evaluate models along two dimensions: response timing and response quality. Timing is measured by Precision, Recall, F1, and False Positive Rate (FPR), capturing response accuracy, coverage, and false activations. Quality is evaluated using Answer Score for overall correctness and relevance, and Content F1 for key-content alignment with reference answers.

% Separate the section heading from the preceding discussion.

\section{StreamVibe: Grounding Evidence, Tracking Intent}
% Keep the method blocks distinct while fitting the complete main text.
\setlength{\parskip}{5pt}
% https://lilianweng.github.io/posts/2026-07-04-harness/
\paragraph{Overview.}
As shown in Figure~\ref{fig:agent}, the formulation in Section~\ref{sec:problem-setup} highlights two key challenges in proactive interaction. \textbf{(i) Evidence Grounding.} The policy $\pi$ must ground each active intent $i_j$ in emerging visual evidence $\xi_{jk}$ and determine when and how to respond. We address this with a dedicated training framework that directly improves the streaming video policy. \textbf{(ii) State Tracking.} As intents evolve, maintaining an up-to-date active set $\mathcal{A}_t$ and interaction history $\mathcal{H}_{<t}$ becomes challenging for the policy alone. We therefore introduce the StreamVibe-Harness $\phi$, which explicitly maintains these interaction states and provides $\pi$ with continuously updated context, $\tilde{c}_t=\phi(c_t,\mathcal{A}_t)$. This separates video-grounded response decisions from persistent state tracking: the Harness tracks dynamic intents and task progress, while response decisions and generation remain with $\pi$. Details of the framework and workflow are provided in Appendix~\ref{app:harness-details}.

\suppressfloats[t]
\begin{figure}[t]
\centering
\captionsetup{skip=6pt}
\includegraphics[width=\linewidth]{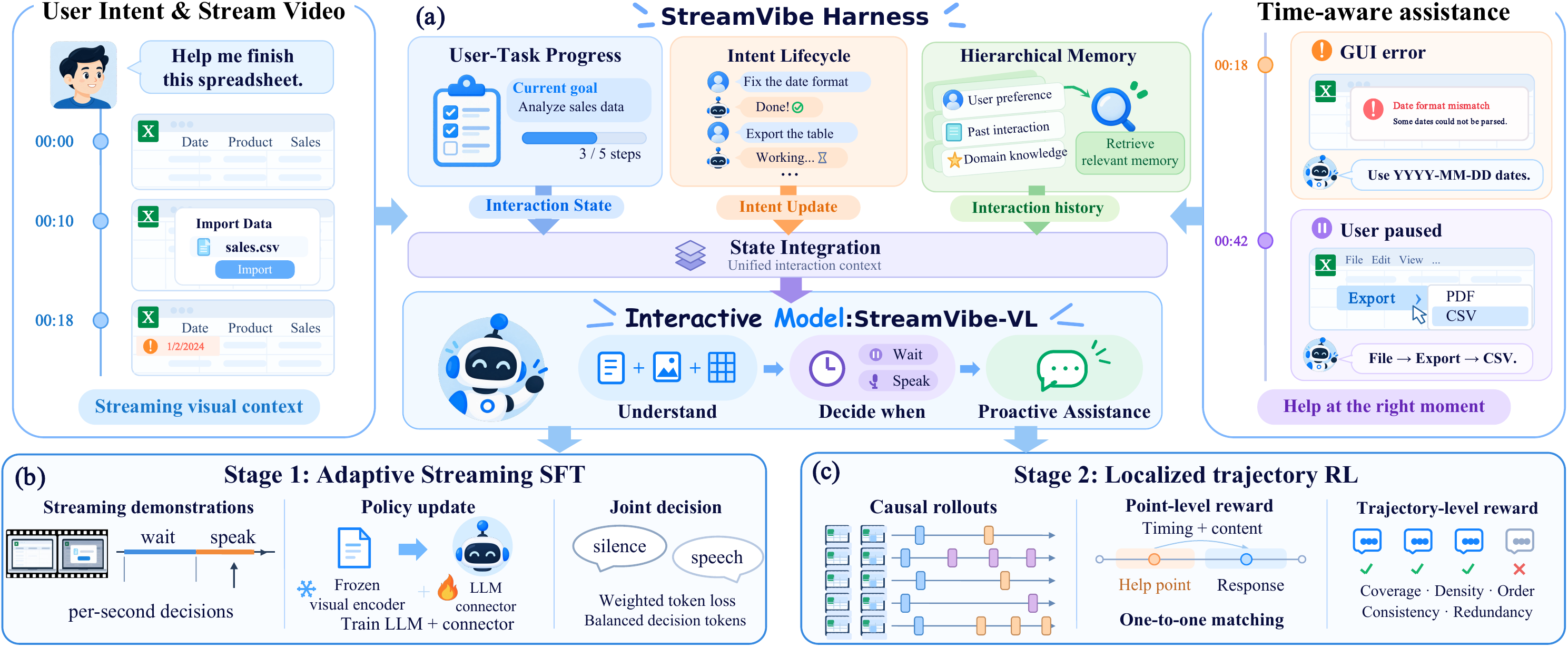}
\caption{\textbf{Overview of StreamVibe.} (a) During inference, StreamVibe-Harness maintains persistent interaction state, tracks each intent through an independent lifecycle, and retrieves relevant history from hierarchical memory, providing StreamVibe-VL with a unified context for response decisions. (b) Adaptive streaming SFT balances wait/speak decisions and jointly supervises timing and content. (c) Localized trajectory RL samples causal rollouts around help points, combining point-level timing and content rewards with trajectory-level rewards on coverage, density, order, consistency, and redundancy.}
\label{fig:agent}
\end{figure}

\subsection{StreamVibe-Harness Framework}
\label{sec:streamvibe-workflow}

\paragraph{Interaction State.} The Harness maintains an explicit interaction state $\mathcal{S}_t=\langle\mathcal{A}_t,\mathcal{P}_t,\mathcal{R}_t\rangle$, where $\mathcal{A}_t$ denotes the currently active intents, $\mathcal{P}_t$ captures task progress and environment state, and $\mathcal{R}_t$ records previously delivered responses. As new user inputs and observations arrive, the Harness incrementally updates $\mathcal{S}_{t-1}$ to $\mathcal{S}_t$ by adding, revising, or terminating intents and updating task progress. State changes are applied only when supported by sufficient interaction or perceptual evidence, avoiding premature updates from uncertain observations.

\paragraph{Intent Lifecycle.} Each intent $i_j$ is maintained as an independent stateful instance, allowing multiple intents to remain active and evolve concurrently. The Harness tracks its current status, relevant evidence, and completion state, and updates the lifecycle as the interaction unfolds. Persistent intents may remain active across multiple events, whereas one-time intents are closed once satisfied. Importantly, the lifecycle only determines which intents remain valid; when and how to respond are still decided by the agent policy $\pi$.

\paragraph{Hierarchical Memory.} The Harness organizes past observations and interactions into hierarchical memory $\mathcal{M}_{<t}$, preserving relevant history without continuously expanding the agent context. Recent information is retained in fine-grained form, while older content is progressively compressed to preserve task-relevant entities, actions, constraints, and state changes. Given the active intents $\mathcal{A}_t$ and task state $\mathcal{P}_t$, the Harness retrieves relevant history as $m_t=\operatorname{Retrieve}(\mathcal{M}_{<t},\mathcal{A}_t,\mathcal{P}_t)$, prioritizing recent and task-relevant information. The retrieved memory serves as additional evidence for the current decision without directly modifying the persistent interaction state.

\paragraph{State Integration.} Before each decision, the Harness integrates the persistent interaction state and retrieved memory with the causal context to obtain $\tilde{c}_t=\phi(c_t,\mathcal{S}_t,m_t)$, which is then passed to the agent policy $\pi$ for response decision and generation. After a response is delivered, the resulting interaction is written back to the state and memory for subsequent decisions.

% Keep the training heading and opening lines in the available page space.
\enlargethispage{12pt}
\subsection{Evidence-Grounded Policy Learning for Streaming Interaction}
% Give the training stages and display equations a little more space.
\setlength{\parskip}{5pt}

\label{sec:training}

\paragraph{\textit{Stage 1}: Adaptive Streaming SFT.} We introduce adaptive streaming supervision, which adjusts decision-token weights to the current training distribution while separately optimizing response generation. This helps balance varying $\mathrm{silence}/\mathrm{response}$ ratios across training sources and reduces bias toward excessive silence or over-response. Our SFT objective is
\begingroup
\begin{equation}
\mathcal{L}_{\mathrm{SFT}}
=
-\frac{1}{|\mathcal{Y}|}
\sum_{j\in\mathcal{Y}}
\omega_j
\log
\pi_{\theta}
\left(
z_j \mid c_{t(j)}, z_{<j}
\right),
\qquad
\omega_j=
\begin{cases}
\gamma_{B,c(j)}, & j\in\mathcal{A},\\
\lambda_{\mathrm{text}}, & j\in\mathcal{R},
\end{cases}
\label{eq:adaptive-sft}
\end{equation}
\endgroup
where $\mathcal{Y}$ denotes supervised assistant tokens, $\mathcal{A}$ and $\mathcal{R}$ denote decision and response tokens, and $c(j)\in\{\mathrm{response},\mathrm{silence}\}$ is the decision class. The adaptive weight $\gamma_{c(j)}^{(t)}$ follows the current training distribution, while $\lambda_{\mathrm{text}}$ preserves response-generation learning.

\paragraph{\textit{Stage 2}: Localized Trajectory RL.} Building on SFT, we introduce localized trajectory reinforcement learning. Full-trajectory optimization mixes correct and erroneous behaviors, leading to diffuse credit assignment and reward interference. We therefore construct short causal rollouts around annotated help points and optimize them with GRPO~\citep{grpo} using complementary point-level and trajectory-level rewards.

\textit{\textbf{(i) Point-Level Reward.}} For each evidence point $g$ and predicted response $p$, the temporal offset is defined as $\Delta_{g,p}=t_p-t_g$. The timing score $\tau(\Delta_{g,p})\in[0,1]$ measures alignment within an asymmetric response window, while $\rho(g,p)\in[0,1]$ measures LLM-judged content quality. Their product $\tau(\Delta_{g,p})\rho(g,p)$ serves as the matching score, and one-to-one matching yields the set $\mathcal{M}$, where each evidence point and response can be matched at most once. The matched scores are then aggregated while accounting for missed evidence points and excessive responses.
\begingroup
\begin{equation}
R_{\mathrm{align}}
=
\frac{
(1+\alpha)
\sum_{(g,p)\in\mathcal{M}}
\tau(\Delta_{g,p})\rho(g,p)
}{
N_{\mathrm{pred}}+\alpha N_{\mathrm{gt}}
}
\label{eq:align-reward}
\end{equation}
\endgroup
Here, $N_{\mathrm{pred}}$ and $N_{\mathrm{gt}}$ denote the numbers of predicted responses and evidence points, respectively, while $\alpha$ controls the trade-off between missed evidence and over-response.  

\Needspace*{3\baselineskip}
\textit{\textbf{(ii) Trajectory-Level reward.}} We further introduce a trajectory-level reward $R_{\mathrm{glob}}$ to evaluate each rollout as a whole. Since point-level reword alone does not ensure coherent interaction, $R_{\mathrm{glob}}$ captures the overall consistency and naturalness of the response sequence, discouraging redundant or conflicting responses and favoring coherent streaming behavior. The final reward is defined as $R=\lambda R_{\mathrm{align}}+(1-\lambda)R_{\mathrm{glob}}$, with $\lambda=0.7$ to place greater emphasis on response timing and content quality while preserving trajectory-level coherence. Since response formatting is already well learned during SFT, the RL stage optimizes only these two objectives without introducing an additional format reward.

\section{Experiments}
\setlength{\parskip}{5pt}
\setlength{\textfloatsep}{14pt plus 2pt minus 1pt}
\setlength{\intextsep}{10pt plus 2pt minus 1pt}
\captionsetup[figure]{skip=5pt}

In this section, we conduct extensive experiments to answer the following research questions: 

\begin{itemize}[leftmargin=*, itemsep=0pt, label=\small$\spadesuit$]

    \item \textbf{RQ1}: How top-performing baseline behave and can StreamVibe-VL outperform them in intent-aware interaction setting?

    \item \textbf{RQ2}: What limitations do current proactive agents exhibit in sustained proactive interaction?
    
    \item \textbf{RQ3}: What gains does the StreamVibe-Harness provide beyond the end-to-end policy?
\end{itemize}

\subsection{Setup}
\paragraph{Benchmarks.}
We evaluate StreamVibe across two settings: \textit{\textbf{(i) Proactive Interaction}} on StreamVibe-Bench, General Timing~\citep{joyai2026}, EgoProactive~\citep{kundu2026planwatchrecover}, and IPIBench~\citep{li2026ipibench}; and \textit{\textbf{(ii) General Video Understanding}} on MVBench~\citep{li2023mvbench}, Video-MME~\citep{fu2024videomme}, OVO-Bench~\citep{li2025ovobench}, and OVBench~\citep{huang2025ovbench}. External suites retain their official protocols.  Further details are provided in Appendix~A.

\paragraph{Baselines.} We benchmark StreamVibe against representative models from the three categories used in Table~\ref{tab:main}: \textit{(i) Proprietary Models}, including Gemini-3.1-Pro and Gemini-3.8-Flash~\citep{google2026gemini3}, and GPT-Realtime-2.1~\citep{openai2026gptrealtime}; \textit{(ii) Open-Source Offline Models}, including Qwen3-VL-8B~\citep{bai2025qwen3vl}, Qwen3.5-9B~\citep{qwen2026qwen35}, and JoyAI-VL-Interaction~\citep{joyai2026}; and \textit{(iii) Open-Source Streaming Models}, including MMDuet2~\citep{wang2025mmduet2}, Mage-VL~\citep{yang2026magevl}, LiveStar~\citep{yang2025livestar}, ViSpeak~\citep{fu2025vispeak}, and VideoLLM-Online~\citep{chen2024videollmonline}.

\paragraph{Evaluation Settings.} For StreamVibe-Bench, we use the protocol and metrics defined in Section~\ref{protocol}, with Claude Sonnet 4.6~\citep{anthropic2026sonnet46} as a shared judge and response quality scored on a 0--4 scale. For external benchmarks, we strictly follow their official evaluation protocols and metrics. See Appendix~\ref{app:evaluation-protocols} for evaluation details.

\begin{table*}[!t]
\centering
\captionsetup{font=normalsize,position=top,skip=4pt}
\caption{Main results on StreamVibe-Bench ($\pm2$\,s tolerance). \textit{Pointwise evaluation} scores each decision given the gold interaction history, whereas \textit{trajectory evaluation} runs the model over the full stream with its self-generated history. Streaming models do not support pointwise evaluation and are thus reported only under trajectory evaluation. \textbf{bold} and \underline{underline} mark the best and second-best results within each block.}
\label{tab:main}
\setlength{\tabcolsep}{1.8pt}
\renewcommand{\arraystretch}{1.06}
\arrayrulecolor{gray!38}
\resizebox{\linewidth}{!}{%
\begin{tabular}{@{}l!{\color{gray!38}\vrule width 0.3pt}*{5}{c}>{\columncolor{metricshade}}c!{\color{gray!38}\vrule width 0.3pt}*{5}{c}>{\columncolor{metricshade}}c!{\color{gray!38}\vrule width 0.3pt}*{5}{c}>{\columncolor{metricshade}}c@{}}
\toprule
\rowcolor{tableheader}
\rule{0pt}{2.4ex}\textbf{Model} & \multicolumn{6}{c}{\textbf{Explicit}} & \multicolumn{6}{c}{\textbf{Implicit}} & \multicolumn{6}{c}{\textbf{Dynamic}} \\
\rowcolor{tableheader}
\rule{0pt}{2.2ex} & P$\uparrow$ & R$\uparrow$ & F1$\uparrow$ & FPR$\downarrow$ & ANS$\uparrow$ & C-F1$\uparrow$ & P$\uparrow$ & R$\uparrow$ & F1$\uparrow$ & FPR$\downarrow$ & ANS$\uparrow$ & C-F1$\uparrow$ & P$\uparrow$ & R$\uparrow$ & F1$\uparrow$ & FPR$\downarrow$ & ANS$\uparrow$ & C-F1$\uparrow$ \\
\midrule
\multicolumn{19}{@{}l}{\cellcolor{gray!12}\rule{0pt}{2.3ex}\textbf{\textit{Pointwise Evaluation}} \quad \textit{(gold interaction history)}} \\
\tabgroup{tableclosed}{Proprietary Models} \\
Gemini-3.1-Pro & \textbf{62.34} & 68.52 & 65.28 & \textbf{41.40} & 45.22 & 29.52 & \underline{70.92} & 85.58 & \underline{77.56} & 43.57 & 67.46 & 52.32 & 62.87 & \underline{84.59} & 72.13 & 58.81 & 51.15 & 36.89 \\
GPT-Realtime-2.1 & 57.78 & 74.09 & 64.93 & 64.87 & 51.10 & 33.18 & 69.67 & 70.95 & 70.30 & \underline{29.83} & 55.92 & 39.31 & \textbf{62.99} & 84.47 & \textbf{72.17} & \underline{55.37} & 45.64 & 32.94 \\
Gemini-3.8-Flash & \underline{60.34} & 77.40 & \textbf{67.81} & 50.69 & \underline{51.37} & \textbf{34.84} & \textbf{81.55} & 80.45 & \textbf{81.00} & \textbf{27.09} & 82.63 & \textbf{66.93} & \underline{62.96} & 84.52 & \underline{72.16} & 60.25 & \textbf{60.56} & \textbf{43.70} \\
\tabgroup{tableopen}{Open-Source offline Models} \\
Qwen3-VL-8B & 51.25 & 82.28 & 63.16 & 89.05 & 40.91 & 25.84 & 56.38 & 57.85 & 57.11 & 83.71 & 80.51 & 45.98 & 54.37 & 87.56 & 67.08 & 93.88 & 45.01 & 30.19 \\
Qwen3.5-9B & 52.95 & \textbf{91.36} & \underline{67.04} & 81.45 & 44.59 & 29.89 & 57.42 & 61.23 & 59.26 & 49.17 & 78.37 & 46.45 & 56.74 & \textbf{90.93} & 69.88 & 78.89 & 47.50 & 33.19 \\
JoyAI-VL-Interaction  & 52.50 & \underline{90.10} & 66.34 & 81.88 & 40.06 & 26.58 & 58.38 & 79.21 & 67.22 & 67.32 & 81.63 & 54.87 & 57.48 & 72.20 & 64.00 & 62.60 & 45.23 & 28.95 \\
\tabgroup{tableours}{\textbf{Ours}} \\
\textbf{StreamVibe-VL (SFT)} & 57.14 & 62.90 & 59.88 & 47.90 & 49.89 & 29.88 & 59.71 & \underline{85.64} & 70.36 & 71.34 & \textbf{85.92} & 60.46 & 58.56 & 58.85 & 58.70 & \textbf{49.13} & 56.68 & 33.27 \\
\textbf{StreamVibe-VL (RL)} & 57.33 & 64.89 & 60.88 & \underline{46.19} & \textbf{55.32} & \underline{33.68} & 61.59 & \textbf{86.66} & 72.01 & 67.91 & \underline{85.11} & \underline{61.28} & 58.94 & 75.86 & 66.34 & 62.93 & \underline{59.25} & \underline{39.31} \\
\midrule
\multicolumn{19}{@{}l}{\cellcolor{gray!12}\rule{0pt}{2.3ex}\textbf{\textit{Trajectory Evaluation}} \quad \textit{(self-generated interaction history)}} \\
\tabgroup{tableopen}{Open-Source offline Models} \\
Qwen3-VL-8B & 3.56 & 40.05 & 6.54 & 19.30 & 36.78 & 2.40 & 3.99 & 35.54 & 7.17 & 26.46 & 40.34 & 2.89 & 2.88 & 48.89 & 5.44 & 18.64 & 25.31 & 1.38 \\
Qwen3.5-9B & 9.37 & 47.62 & 15.66 & 19.30 & 16.63 & 2.60 & 7.94 & \underline{52.48} & 13.79 & 14.56 & 44.98 & 6.20 & 4.87 & 51.65 & 8.90 & 20.95 & 27.67 & 2.46 \\
JoyAI-VL-Interaction & 4.67 & \textbf{95.44} & 8.90 & 30.87 & 20.21 & 1.80 & 4.39 & 37.96 & 7.87 & 18.87 & 46.83 & 3.69 & 3.15 & \textbf{76.70} & 6.05 & 64.52 & 21.64 & 1.31 \\
\tabgroup{tableopen}{Open-Source Streaming Models} \\
MMDuet2 & 19.45 & 38.65 & 25.88 & 7.90 & 22.71 & 5.88 & 15.98 & 36.50 & 22.23 & 4.13 & \textbf{74.26} & \underline{16.51} & 13.06 & 35.51 & 19.10 & 5.24 & \underline{39.10} & \underline{7.47} \\
Mage-VL & \underline{20.70} & 50.25 & \underline{29.32} & 6.11 & 10.23 & 3.00 & 8.56 & 51.94 & 14.70 & 5.50 & 8.19 & 1.20 & 11.97 & 52.88 & 19.52 & 6.20 & 6.10 & 1.19 \\
LiveStar & 14.96 & 19.95 & 17.10 & 4.81 & 3.60 & 0.62 & 6.84 & 24.95 & 10.74 & 6.27 & 25.43 & 2.73 & 7.29 & 12.92 & 9.32 & \underline{2.86} & 8.05 & 0.75 \\
ViSpeak & 20.66 & \underline{74.71} & \textbf{32.37} & 12.75 & 10.19 & 3.30 & 8.41 & \textbf{74.03} & 15.10 & 12.67 & 62.98 & 9.51 & 13.51 & \underline{72.32} & \textbf{22.77} & 14.19 & 16.85 & 3.84 \\
VideoLLM-Online & 12.29 & 5.04 & 7.15 & \textbf{1.44} & 17.89 & 1.28 & 6.33 & 17.66 & 9.32 & 5.32 & 31.88 & 2.97 & \textbf{17.86} & 14.75 & 16.16 & \textbf{1.28} & 8.69 & 1.40 \\
\tabgroup{tableours}{\textbf{Ours}} \\
\textbf{StreamVibe-VL (SFT)} & 18.20 & 29.39 & 22.48 & 6.33 & \underline{38.93} & \underline{8.75} & \underline{18.49} & 28.51 & \underline{22.43} & \underline{3.80} & 61.79 & 13.86 & 10.01 & 43.20 & 16.25 & 12.25 & 38.98 & 6.34 \\
\textbf{StreamVibe-VL (RL)} & \textbf{21.53} & 34.48 & 26.51 & \underline{2.02} & \textbf{40.86} & \textbf{10.83} & \textbf{24.57} & 28.20 & \textbf{26.26} & \textbf{1.49} & \underline{72.83} & \textbf{19.13} & \underline{14.80} & 30.65 & \underline{19.96} & 5.68 & \textbf{44.62} & \textbf{8.91} \\
\bottomrule
\end{tabular}}
\end{table*}

% Use the remaining space for the analysis heading and its opening lines.
\enlargethispage{12pt}
\Needspace{5\baselineskip}
\subsection{Analysis}
\setlength{\parskip}{5pt}

\noindent\textbf{$\spadesuit$ RQ1: \textit{StreamVibe-VL achieves stronger proactive interaction while preserving general video understanding.}} Table~\ref{tab:main} shows that StreamVibe-VL consistently outperforms open-source baselines on StreamVibe-Bench. Under Pointwise Evaluation, it achieves the highest Content F1 while remaining competitive with proprietary models in response quality. RL further improves over SFT, with larger gains on dynamic intents. This trend persists under Trajectory Evaluation, where RL yields higher Content F1 and response quality, indicating stronger sustained interaction when decisions depend on prior history. The gains also transfer to external benchmarks. Figure~\ref{fig:general_bench_radar}(a-b) shows competitive results on EgoProactive and consistent gains over other models on IPIBench. Detailed results and analysis are provided in Appendix~\ref{sec:additional_results}.

\Needspace{15\baselineskip}
\begin{wrapfigure}[14]{r}{0.62\linewidth}
\centering
\captionsetup{skip=4pt}
\setlength{\belowcaptionskip}{4pt}
\includegraphics[width=\linewidth]{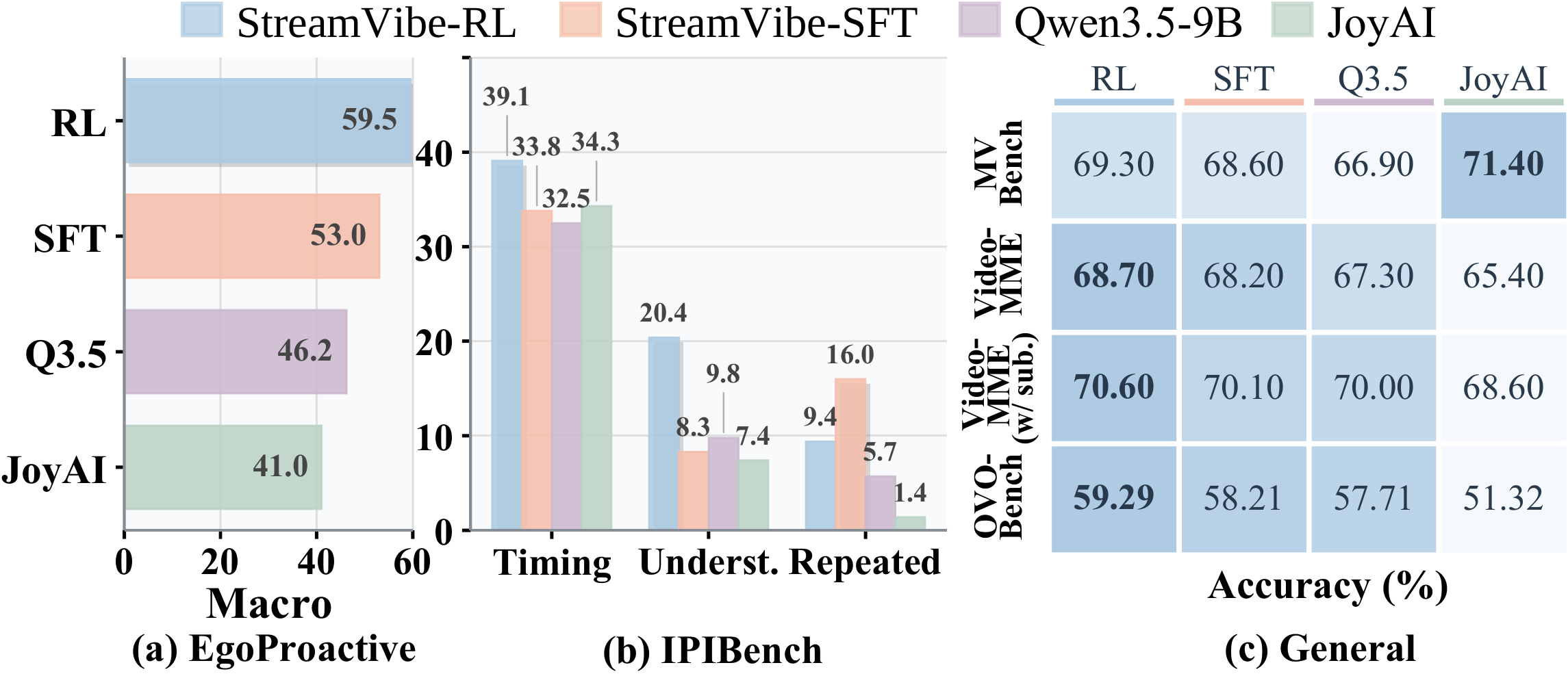}
\caption{Generalization to external benchmarks. (a) Macro score on EgoProactive; (b) submetric scores on IPIBench; (c) accuracy on general video benchmarks, with darker cells indicating higher scores within each row.}
\label{fig:general_bench_radar}
\end{wrapfigure}

Figure~\ref{fig:general_bench_radar}(c) further shows that our model outperforms its base model across all general video benchmarks, indicating that proactive policy learning preserves general video understanding.
% Continue the following analysis beside the figure to avoid an empty left column.

\begin{figure}[!t]
\centering

\includegraphics[width=\linewidth]{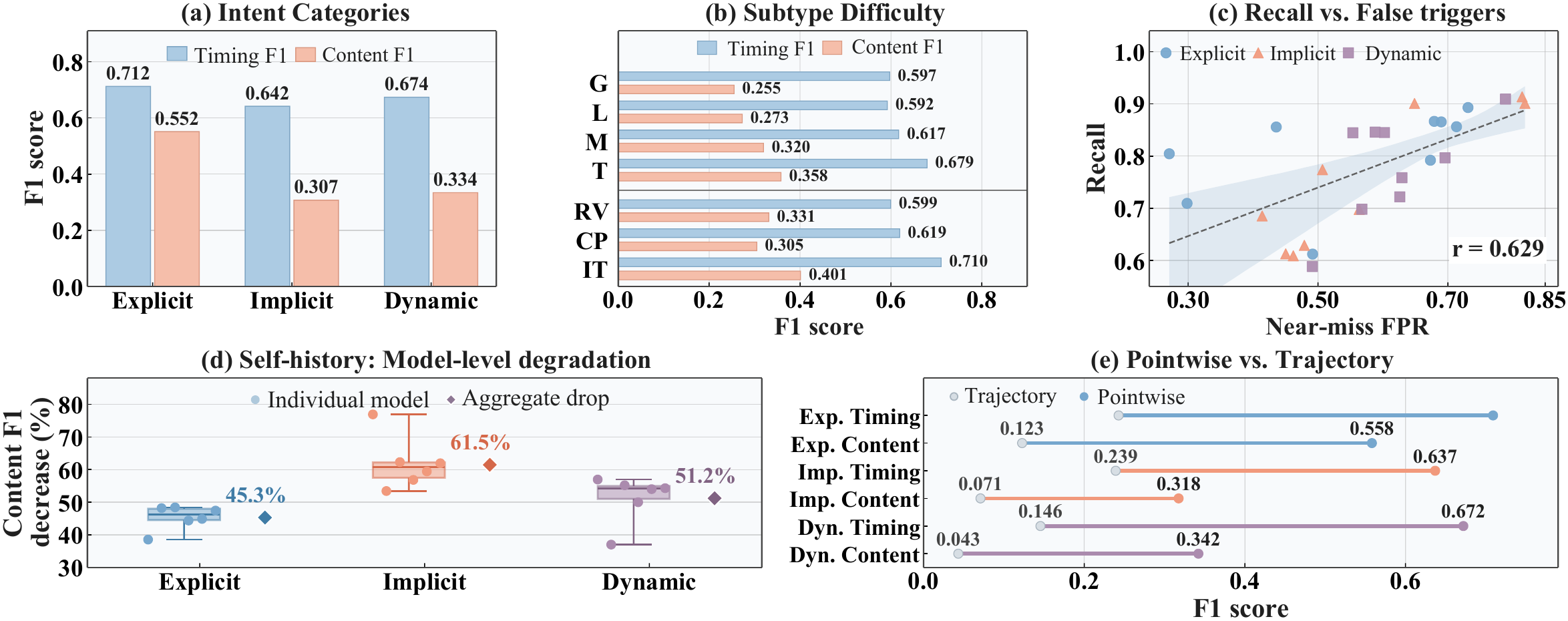}

\caption{\textbf{Diagnostic analysis of proactive agents}. (a)-(b) Timing and Content F1 across intent categories and subtypes; (c) Recall versus near-miss FPR across models; (d) Content F1 degradation under accumulated self-generated history; (e) pointwise versus trajectory F1 per category and metric.}
\label{fig:rq1_diagnostics}

\end{figure}

\enlargethispage{14pt}
\Needspace{3\baselineskip}
\paragraph{$\spadesuit$ RQ2: \textit{Current proactive agents struggle to sustain reliable behavior over continuous interaction.}} Figure~\ref{fig:rq1_diagnostics} and Figure~\ref{fig:rq234_analysis} reveal three main findings. \textit{(i) Intent understanding remains difficult.} Implicit and Dynamic Intents are markedly harder than Explicit Intents, indicating that agents struggle to determine what help is needed when intents are unstated or keep changing. \textit{(ii) Response selectivity is insufficient.} Higher Recall often comes with more false triggers, suggesting that current methods cannot yet balance coverage against unnecessary responses. \textit{(iii) Sustained interaction amplifies these problems.} Timing F1 declines steadily with interaction length; Revision shows a sharper late-stage Recall drop, and Composition exhibits weaker content quality, indicating difficulty in continuously updating intents while maintaining multiple constraints. Correct individual judgments are thus insufficient for stable proactive behavior throughout interaction.
\par

% Keep the run-in heading with its opening text, without reserving a whole paragraph.
\paragraph{$\spadesuit$ RQ3: \textit{Policy learning improves proactive decision-making, while state tracking enhances sustained interaction stability.}} As shown in Figure~\ref{fig:rq234_analysis}(2), SFT makes the model more sensitive to potential assistance opportunities, leading to higher Recall but also more unnecessary triggers. RL further improves decision selectivity, substantially reducing FPR while also improving answer quality. Figure~\ref{fig:rq234_analysis}(3) further shows that, with the policy fixed, adding the Harness continues to suppress unnecessary responses and improves content quality in most settings, with the clearest gains on Dynamic Intents. These results suggest that policy learning mainly improves local proactive decisions, whereas state tracking helps preserve valid intents and task progress over long-horizon interactions, thereby reducing outdated, redundant, or state-inconsistent outputs.

\setlength{\intextsep}{10pt plus 2pt minus 1pt}
\setlength{\parskip}{5pt}
\enlargethispage{14pt}
\begin{figure}[H]
\centering
\captionsetup{skip=5pt}
\setlength{\belowcaptionskip}{0pt}
\includegraphics[width=\linewidth]{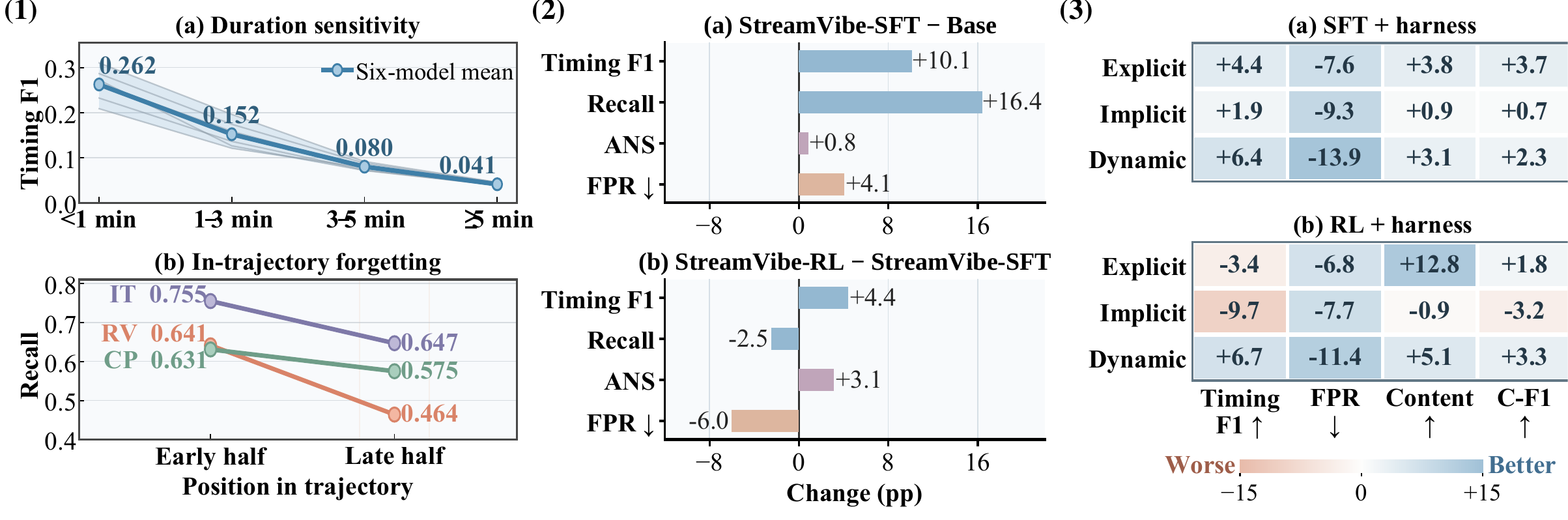}
\caption{\textbf{Effects of interaction length, policy learning and the harness.} (1) Timing F1 versus video duration, and Recall in the early versus late half of the trajectory for each Dynamic Intent subtype. (2) Absolute metric changes of SFT over the base model and of RL over SFT. (3) Gains from adding the Harness to the SFT and RL policies across intent categories. Tolerance: $\tau=2$\,s.}
\label{fig:rq234_analysis}
\label{fig:rq23_diagnostics}
\label{fig:rq3_training}
\label{fig:rq4_harness}
\end{figure}

% Preserve normal paragraph spacing when the analysis figure floats.
\begingroup
\setlength{\parskip}{5pt}
\noindent\textbf{Qualitative Analysis.}\quad Figure~\ref{fig:case_study} illustrates how StreamVibe-VL assists users in representative physical and digital interactions. Across multiple assistance points within the same task, it grounds responses in the current visual evidence and task state, providing assistance at appropriate moments. In contrast, streaming baselines often respond before sufficient evidence appears or fail to update later responses as the task evolves, yielding premature, inconsistent, or redundant assistance. These cases show that our agent keeps both response timing and content consistent throughout sustained interaction. Additional qualitative examples are provided in Appendix~\ref{app:additional-cases}.
\par\endgroup

\begin{figure}[H]
\centering
\captionsetup{skip=3pt}
\includegraphics[width=\linewidth]{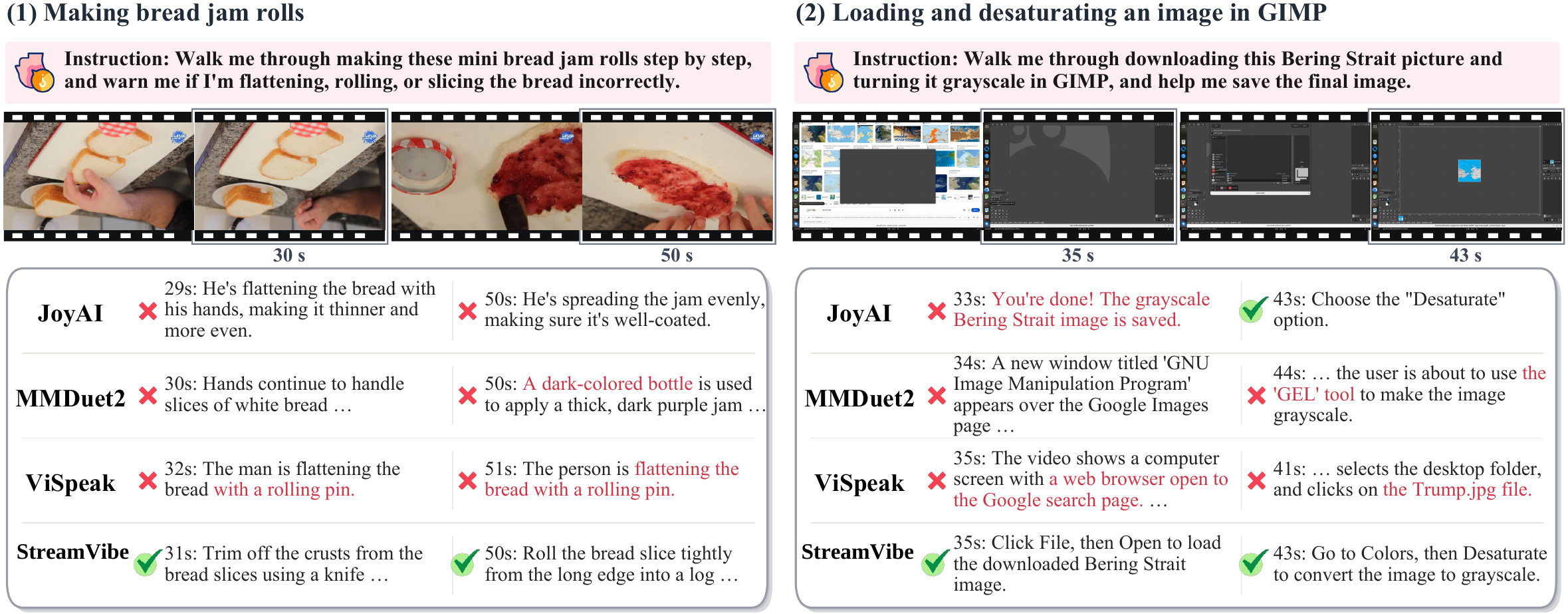}
\caption{\textbf{Case studies of implicit proactive interaction.} Across physical and digital tasks, StreamVibe-VL better aligns response timing and content with the evolving visual evidence and task state, while streaming baselines more often produce premature, missing, or inconsistent assistance.}
\label{fig:case_study}
\end{figure}

% Keep the concluding heading close to its text in the remaining page space.
\titlespacing*{\section}{0pt}{10pt}{4pt}
\section{Conclusion}
We study proactive interaction over streaming video, where user intents may be implicit, evolve over time, and must be tracked alongside a changing environment. We introduce StreamVibe-Bench, a unified benchmark spanning explicit, implicit, and dynamic intents with pointwise and trajectory evaluation; StreamVibe-VL, a proactive streaming policy trained with adaptive streaming SFT and localized trajectory RL; and StreamVibe-Harness, which maintains user intents and task progress as persistent interaction states. Experiments show that our model improves response timing and quality while preserving general video understanding, while explicit state tracking further reduces outdated and redundant responses. More broadly, real-world proactive interaction may involve richer intent evolution and tighter coupling between user needs and environmental changes than captured in our current setting. New observations, task outcomes, and prior assistance may jointly reshape both user needs and the environment, leading to new goals and constraints. Future work could further study such co-evolving interactions, particularly the complex feedback between intents and environmental states over longer horizons.

% We study proactive interaction over streaming video, where user intents may be implicit, evolve over time, and must be tracked alongside a changing environment. We introduce StreamVibe-Bench, a unified benchmark spanning explicit, implicit, and dynamic intents with pointwise and trajectory evaluation; StreamVibe-VL, a proactive streaming policy trained with adaptive streaming SFT and localized trajectory RL; and StreamVibe-Harness, which maintains intents and task progress as persistent interaction state. Experiments show that StreamVibe-VL improves response timing and quality while preserving general video understanding, and that explicit state tracking further suppresses outdated and redundant responses beyond the learned policy.
% However, our framework has not yet been extended to or deployed in more complex and diverse real-world scenarios. Modeling the evolution of user intent remains an avenue for further refinement and extension.

\endgroup
\setlength{\parskip}{6pt}
\setlength{\bibsep}{1pt plus 0.2pt}
\FloatBarrier
\bibliography{iclr2027_conference,dataset_sources}

\appendix
\pretocmd{\section}{\FloatBarrier\suppressfloats[t]}{}{}
\pretocmd{\subsection}{\FloatBarrier\suppressfloats[t]}{}{}
\makeatletter
\setlength{\@fptop}{0pt}
\setlength{\@fpsep}{10pt}
\makeatother
% BEGIN embedded appendix_results.tex
\setlength{\floatsep}{12pt plus 2pt minus 1pt}
\setlength{\textfloatsep}{14pt plus 2pt minus 1pt}
\setlength{\intextsep}{10pt plus 2pt minus 1pt}
\renewcommand{\topfraction}{0.98}
\renewcommand{\bottomfraction}{0.95}
\renewcommand{\textfraction}{0.02}
\renewcommand{\floatpagefraction}{0.82}
\setcounter{topnumber}{4}
\setcounter{bottomnumber}{4}
\setcounter{totalnumber}{6}

\section{Additional Evaluation Results}

\label{sec:additional_results}

We report Pointwise evaluation with gold or model-generated interaction history and Trajectory evaluation with model-generated history. The matching tolerances are $\pm2$, $\pm1$, and $\pm0$\,s; tighter tolerances rescore the same predictions. The following model-comparison tables report StreamVibe-VL without the Harness, followed by separate Harness comparisons. \textbf{Bold} and \underline{underline} denote the best and second-best eligible results in each column; FPR is lower-is-better and all other metrics are higher-is-better.

\subsection{Results Under the Three Evaluation Protocols}
\label{app:three-protocol-results}

Tables~\ref{tab:app_pointwise_gold_tau2} and~\ref{tab:app_pointwise_gold_tau0} report Pointwise evaluation with gold interaction history,
Tables~\ref{tab:app_pointwise_self_tau2} and~\ref{tab:app_pointwise_self_tau0} report Pointwise evaluation with self-generated interaction history,
and Tables~\ref{tab:app_trajectory_tau2} and~\ref{tab:app_trajectory_tau0} report Trajectory evaluation with self-generated interaction history.

\begin{table}[!htbp]
\centering
\setlength{\tabcolsep}{2.8pt}
\renewcommand{\arraystretch}{1.05}
\arrayrulecolor{gray!38}
\resizebox{\ifdim\width>0.99\linewidth 0.99\linewidth\else\width\fi}{!}{%
\begin{tabular}{@{}l!{\color{gray!38}\vrule width 0.3pt}ccccc>{\columncolor{metricshade}}c!{\color{gray!38}\vrule width 0.3pt}ccccc>{\columncolor{metricshade}}c!{\color{gray!38}\vrule width 0.3pt}ccccc>{\columncolor{metricshade}}c@{}}
\toprule
\rowcolor{tableheader}
\rule{0pt}{2.4ex}\textbf{Model} & \multicolumn{6}{c}{\textbf{Explicit}} & \multicolumn{6}{c}{\textbf{Implicit}} & \multicolumn{6}{c}{\textbf{Dynamic}} \\
\rowcolor{tableheader}
\rule{0pt}{2.2ex} & P & R & F1 & FPR & ANS & C-F1 & P & R & F1 & FPR & ANS & C-F1 & P & R & F1 & FPR & ANS & C-F1 \\
\midrule
\rowcolor{tableclosed}\multicolumn{19}{@{}l@{}}{\rule{0pt}{2.25ex}\hspace{0.4em}\textit{Proprietary Models}} \\
Gemini-3.1-Pro & \underline{70.92} & 85.58 & \underline{77.56} & 43.57 & 67.46 & 52.32 & \textbf{62.34} & 68.52 & 65.28 & 41.40 & 45.22 & 29.52 & 62.87 & 84.59 & 72.13 & 58.81 & 51.15 & 36.89 \\
GPT-Realtime-2.1 & 69.67 & 70.95 & 70.30 & 29.83 & 55.92 & 39.31 & 57.78 & 90.09 & \textbf{70.41} & 64.87 & 51.10 & \textbf{35.98} & \textbf{62.99} & 84.47 & \textbf{72.17} & 55.37 & 45.64 & 32.94 \\
Gemini-3.8-Flash & \textbf{81.55} & 80.45 & \textbf{81.00} & 27.09 & 82.63 & \textbf{66.93} & \underline{60.34} & 77.40 & \underline{67.81} & 50.69 & 51.37 & \underline{34.83} & \underline{62.96} & 84.52 & \underline{72.16} & 60.25 & \textbf{60.56} & \textbf{43.70} \\
\rowcolor{tableopen}\multicolumn{19}{@{}l@{}}{\rule{0pt}{2.25ex}\hspace{0.4em}\textit{Open-Source Offline Models}} \\
Qwen3.5-9B & 57.42 & 61.23 & 59.26 & 49.17 & 78.37 & 46.44 & 52.95 & \underline{91.36} & 67.04 & 81.45 & 44.59 & 29.89 & 56.74 & \underline{90.93} & 69.88 & 78.89 & 47.50 & 33.19 \\
Qwen3-VL-8B & 56.38 & \textbf{90.87} & 69.59 & 83.71 & 80.51 & 56.03 & 51.25 & \textbf{95.28} & 66.65 & 91.05 & 47.91 & 31.93 & 54.37 & \textbf{97.56} & 69.83 & 93.88 & 50.91 & 35.55 \\
JoyAI-VL-Interaction & 58.38 & 79.21 & 67.22 & 67.32 & 81.63 & 54.87 & 52.50 & 90.10 & 66.34 & 81.88 & 40.06 & 26.58 & 57.48 & 72.20 & 64.00 & 62.60 & 45.23 & 28.95 \\
\rowcolor{tableopen}\multicolumn{19}{@{}l@{}}{\rule{0pt}{2.25ex}\hspace{0.4em}\textit{Open-Source Streaming Models}} \\
MMDuet2 & 15.98 & 36.50 & 22.23 & \textbf{4.13} & 74.26 & 16.51 & 19.45 & 38.65 & 25.88 & 7.90 & 22.71 & 5.88 & 13.06 & 35.51 & 19.10 & 5.24 & 39.10 & 7.47 \\
Mage-VL & 8.56 & 51.94 & 14.70 & 5.50 & 8.19 & 1.20 & 20.70 & 50.25 & 29.32 & 6.11 & 10.23 & 3.00 & 11.97 & 52.88 & 19.52 & 6.20 & 6.10 & 1.19 \\
LiveStar & 6.84 & 24.95 & 10.74 & 6.27 & 25.43 & 2.73 & 14.96 & 19.95 & 17.10 & \underline{4.81} & 3.60 & 0.62 & 7.29 & 12.92 & 9.32 & \underline{2.86} & 8.05 & 0.75 \\
ViSpeak & 8.41 & 74.03 & 15.10 & 12.67 & 62.98 & 9.51 & 20.66 & 74.71 & 32.37 & 12.75 & 10.19 & 3.30 & 13.51 & 72.32 & 22.77 & 14.19 & 16.85 & 3.84 \\
VideoLLM-Online & 6.33 & 17.66 & 9.32 & \underline{5.32} & 31.88 & 2.97 & 12.29 & 5.04 & 7.15 & \textbf{1.44} & 17.89 & 1.28 & 17.86 & 14.75 & 16.16 & \textbf{1.28} & 8.69 & 1.40 \\
\rowcolor{tableours}\multicolumn{19}{@{}l@{}}{\rule{0pt}{2.25ex}\hspace{0.4em}\textit{\textbf{Ours}}} \\
\textbf{StreamVibe-VL (SFT)} & 60.19 & \underline{89.31} & 71.91 & 73.11 & \underline{85.02} & 61.14 & 55.74 & 69.79 & 61.98 & 56.35 & \underline{53.65} & 33.25 & 57.56 & 79.68 & 66.84 & 69.58 & 57.67 & 38.55 \\
\textbf{StreamVibe-VL (RL)} & 61.59 & 86.66 & 72.01 & 67.91 & \textbf{85.11} & \underline{61.29} & 57.33 & 60.89 & 59.06 & 46.19 & \textbf{55.32} & 32.67 & 58.94 & 75.86 & 66.34 & 62.93 & \underline{59.25} & \underline{39.31} \\
\bottomrule
\end{tabular}}

\caption{\footnotesize \textbf{\textit{Pointwise evaluation with gold interaction history}} on StreamVibe-Bench without the Harness ($\pm2$\,s tolerance).}
\label{tab:app_pointwise_gold_tau2}
\label{tab:pointwise_tau20_full}
\end{table}

\begin{table}[!htbp]
\centering
\setlength{\tabcolsep}{2.8pt}
\renewcommand{\arraystretch}{1.05}
\arrayrulecolor{gray!38}
\resizebox{\ifdim\width>0.99\linewidth 0.99\linewidth\else\width\fi}{!}{%
\begin{tabular}{@{}l!{\color{gray!38}\vrule width 0.3pt}ccccc>{\columncolor{metricshade}}c!{\color{gray!38}\vrule width 0.3pt}ccccc>{\columncolor{metricshade}}c!{\color{gray!38}\vrule width 0.3pt}ccccc>{\columncolor{metricshade}}c@{}}
\toprule
\rowcolor{tableheader}
\rule{0pt}{2.4ex}\textbf{Model} & \multicolumn{6}{c}{\textbf{Explicit}} & \multicolumn{6}{c}{\textbf{Implicit}} & \multicolumn{6}{c}{\textbf{Dynamic}} \\
\rowcolor{tableheader}
\rule{0pt}{2.2ex} & P & R & F1 & FPR & ANS & C-F1 & P & R & F1 & FPR & ANS & C-F1 & P & R & F1 & FPR & ANS & C-F1 \\
\midrule
\rowcolor{tableclosed}\multicolumn{19}{@{}l@{}}{\rule{0pt}{2.25ex}\hspace{0.4em}\textit{Proprietary Models}} \\
Gemini-3.1-Pro & \underline{67.47} & 81.43 & \underline{73.80} & 51.13 & 68.92 & 50.86 & 53.46 & 58.76 & 55.98 & 52.96 & 46.73 & 26.16 & \underline{58.03} & 78.07 & 66.57 & 68.27 & 52.94 & 35.24 \\
GPT-Realtime-2.1 & 63.10 & 64.25 & 63.67 & 42.00 & 57.79 & 36.79 & 50.76 & 79.14 & 61.85 & 77.83 & 51.78 & \underline{32.03} & 55.68 & 74.67 & 63.79 & 69.60 & 46.60 & 29.73 \\
Gemini-3.8-Flash & \textbf{79.04} & 77.97 & \textbf{78.50} & 31.60 & 83.41 & \textbf{65.48} & \underline{53.48} & 68.61 & 60.11 & 61.10 & 52.45 & 31.53 & \textbf{59.01} & 79.21 & 67.63 & 67.94 & \textbf{61.25} & \textbf{41.42} \\
\rowcolor{tableopen}\multicolumn{19}{@{}l@{}}{\rule{0pt}{2.25ex}\hspace{0.4em}\textit{Open-Source Offline Models}} \\
Qwen3.5-9B & 54.78 & 58.42 & 56.54 & 54.27 & 79.23 & 44.80 & 51.40 & 88.69 & 65.08 & 84.61 & 44.91 & 29.23 & 55.39 & \underline{88.77} & \underline{68.22} & 82.03 & 47.61 & 32.48 \\
Qwen3-VL-8B & 55.38 & \textbf{89.25} & 68.35 & 86.65 & 81.16 & 55.47 & 50.52 & \textbf{93.93} & \textbf{65.70} & 92.66 & 47.82 & 31.42 & 54.18 & \textbf{97.22} & \textbf{69.58} & 94.37 & 50.97 & 35.46 \\
JoyAI-VL-Interaction & 57.78 & 78.40 & 66.53 & 68.79 & 81.96 & 54.53 & 51.82 & \underline{88.94} & \underline{65.49} & 83.25 & 40.27 & 26.37 & 56.31 & 70.74 & 62.71 & 64.73 & 45.40 & 28.47 \\
\rowcolor{tableopen}\multicolumn{19}{@{}l@{}}{\rule{0pt}{2.25ex}\hspace{0.4em}\textit{Open-Source Streaming Models}} \\
MMDuet2 & 4.77 & 10.91 & 6.64 & \underline{7.13} & 75.87 & 5.04 & 5.49 & 10.90 & 7.30 & 11.10 & 24.07 & 1.76 & 7.12 & 19.35 & 10.41 & 6.95 & 38.66 & 4.02 \\
Mage-VL & 1.97 & 11.93 & 3.38 & 10.19 & 9.84 & 0.33 & 7.58 & 18.40 & 10.74 & 9.78 & 9.65 & 1.04 & 3.05 & 13.48 & 4.97 & 10.38 & 7.07 & 0.35 \\
LiveStar & 2.95 & 10.75 & 4.63 & 7.93 & 27.01 & 1.25 & 7.31 & 9.75 & 8.36 & \underline{5.98} & 3.85 & 0.32 & 2.61 & 4.63 & 3.34 & \underline{3.73} & 10.36 & 0.35 \\
ViSpeak & 2.16 & 19.01 & 3.88 & 19.12 & 64.28 & 2.49 & 5.52 & 19.96 & 8.65 & 19.06 & 10.99 & 0.95 & 4.86 & 26.02 & 8.19 & 19.10 & 16.76 & 1.37 \\
VideoLLM-Online & 2.53 & 7.07 & 3.73 & \textbf{6.56} & 29.77 & 1.11 & 6.66 & 2.73 & 3.87 & \textbf{1.71} & 19.82 & 0.77 & 15.24 & 12.59 & 13.79 & \textbf{1.51} & 5.52 & 0.76 \\
\rowcolor{tableours}\multicolumn{19}{@{}l@{}}{\rule{0pt}{2.25ex}\hspace{0.4em}\textit{\textbf{Ours}}} \\
\textbf{StreamVibe-VL (SFT)} & 59.83 & \underline{88.77} & 71.48 & 74.09 & \underline{85.14} & \underline{60.86} & 53.46 & 66.95 & 59.45 & 59.72 & \underline{54.21} & \textbf{32.23} & 56.36 & 78.01 & 65.44 & 72.00 & 58.14 & 38.05 \\
\textbf{StreamVibe-VL (RL)} & 61.01 & 85.85 & 71.33 & 69.38 & \textbf{85.24} & 60.80 & \textbf{54.45} & 57.83 & 56.09 & 49.81 & \textbf{55.90} & 31.35 & 57.45 & 73.94 & 64.66 & 65.71 & \underline{59.70} & \underline{38.60} \\
\bottomrule
\end{tabular}}

\caption{\footnotesize \textbf{\textit{Pointwise evaluation with gold interaction history}} on StreamVibe-Bench without the Harness ($\pm0$\,s tolerance).}
\label{tab:app_pointwise_gold_tau0}
\end{table}

\begin{table}[!htbp]
\centering
\setlength{\tabcolsep}{2.8pt}
\renewcommand{\arraystretch}{1.05}
\arrayrulecolor{gray!38}
\resizebox{\ifdim\width>0.99\linewidth 0.99\linewidth\else\width\fi}{!}{%
\begin{tabular}{@{}l!{\color{gray!38}\vrule width 0.3pt}ccccc>{\columncolor{metricshade}}c!{\color{gray!38}\vrule width 0.3pt}ccccc>{\columncolor{metricshade}}c!{\color{gray!38}\vrule width 0.3pt}ccccc>{\columncolor{metricshade}}c@{}}
\toprule
\rowcolor{tableheader}
\rule{0pt}{2.4ex}\textbf{Model} & \multicolumn{6}{c}{\textbf{Explicit}} & \multicolumn{6}{c}{\textbf{Implicit}} & \multicolumn{6}{c}{\textbf{Dynamic}} \\
\rowcolor{tableheader}
\rule{0pt}{2.2ex} & P & R & F1 & FPR & ANS & C-F1 & P & R & F1 & FPR & ANS & C-F1 & P & R & F1 & FPR & ANS & C-F1 \\
\midrule
\rowcolor{tableclosed}\multicolumn{19}{@{}l@{}}{\rule{0pt}{2.25ex}\hspace{0.4em}\textit{Proprietary Models}} \\
Gemini-3.1-Pro & 66.28 & \textbf{82.45} & 73.49 & 33.17 & 40.95 & 30.09 & 68.23 & 70.21 & 69.21 & 29.31 & 29.20 & 20.21 & 73.03 & 80.05 & \underline{76.38} & 28.13 & 36.70 & 28.03 \\
GPT-Realtime-2.1 & 65.42 & 72.84 & 68.93 & 34.54 & 54.10 & 37.29 & 61.47 & 85.03 & 71.36 & 50.95 & \textbf{48.94} & \underline{34.92} & 66.86 & 79.83 & 72.77 & 42.50 & 41.99 & 30.56 \\
Gemini-3.8-Flash & \textbf{89.93} & 75.22 & \textbf{81.92} & 5.79 & 71.88 & \textbf{58.88} & \underline{79.24} & 79.49 & \underline{79.36} & 13.77 & \underline{46.90} & \textbf{37.22} & \textbf{85.98} & \textbf{82.75} & \textbf{84.33} & 7.27 & \textbf{56.27} & \textbf{47.45} \\
\rowcolor{tableopen}\multicolumn{19}{@{}l@{}}{\rule{0pt}{2.25ex}\hspace{0.4em}\textit{Open-Source Offline Models}} \\
Qwen3.5-9B & 66.55 & 29.86 & 41.22 & 8.24 & 68.04 & 28.05 & 56.04 & \underline{89.04} & 68.79 & 67.93 & 22.90 & 15.75 & 62.64 & \underline{80.52} & 70.46 & 51.20 & 32.39 & 22.82 \\
JoyAI-VL-Interaction & 64.78 & 67.33 & 66.03 & 29.64 & 68.08 & 44.95 & 78.54 & \textbf{91.23} & \textbf{84.41} & 14.37 & 21.65 & 18.27 & \underline{83.76} & 59.29 & 69.43 & 4.54 & 33.53 & 23.28 \\
\rowcolor{tableopen}\multicolumn{19}{@{}l@{}}{\rule{0pt}{2.25ex}\hspace{0.4em}\textit{Open-Source Streaming Models}} \\
MMDuet2 & 15.98 & 36.50 & 22.23 & \textbf{4.13} & \textbf{74.26} & 16.51 & 19.45 & 38.65 & 25.88 & 7.90 & 22.71 & 5.88 & 13.06 & 35.51 & 19.10 & 5.24 & 39.10 & 7.47 \\
Mage-VL & 8.56 & 51.94 & 14.70 & 5.50 & 8.19 & 1.20 & 20.70 & 50.25 & 29.32 & 6.11 & 10.23 & 3.00 & 11.97 & 52.88 & 19.52 & 6.20 & 6.10 & 1.19 \\
LiveStar & 6.84 & 24.95 & 10.74 & 6.27 & 25.43 & 2.73 & 14.96 & 19.95 & 17.10 & \underline{4.81} & 3.60 & 0.62 & 7.29 & 12.92 & 9.32 & \underline{2.86} & 8.05 & 0.75 \\
ViSpeak & 8.41 & 74.03 & 15.10 & 12.67 & 62.98 & 9.51 & 20.66 & 74.71 & 32.37 & 12.75 & 10.19 & 3.30 & 13.51 & 72.32 & 22.77 & 14.19 & 16.85 & 3.84 \\
VideoLLM-Online & 6.33 & 17.66 & 9.32 & \underline{5.32} & 31.88 & 2.97 & 12.29 & 5.04 & 7.15 & \textbf{1.44} & 17.89 & 1.28 & 17.86 & 14.75 & 16.16 & \textbf{1.28} & 8.69 & 1.40 \\
\rowcolor{tableours}\multicolumn{19}{@{}l@{}}{\rule{0pt}{2.25ex}\hspace{0.4em}\textit{\textbf{Ours}}} \\
\textbf{StreamVibe-VL (SFT)} & 69.05 & 76.03 & 72.37 & 27.09 & 70.42 & 50.96 & 77.54 & 62.79 & 69.39 & 12.51 & 42.46 & 29.46 & 76.98 & 69.37 & 72.98 & 16.38 & 45.94 & 33.53 \\
\textbf{StreamVibe-VL (RL)} & \underline{73.58} & \underline{76.67} & \underline{75.09} & 17.66 & \underline{72.43} & \underline{54.39} & \textbf{84.28} & 65.75 & 73.87 & 5.00 & 45.71 & 33.77 & 82.70 & 68.45 & 74.90 & 8.05 & \underline{48.53} & \underline{36.35} \\
\bottomrule
\end{tabular}}

\caption{\footnotesize \textbf{\textit{Pointwise evaluation with self-generated interaction history}} on StreamVibe-Bench without the Harness ($\pm2$\,s tolerance).}
\label{tab:app_pointwise_self_tau2}
\end{table}

\begin{table}[!htbp]
\centering
\setlength{\tabcolsep}{2.8pt}
\renewcommand{\arraystretch}{1.05}
\arrayrulecolor{gray!38}
\resizebox{\ifdim\width>0.99\linewidth 0.99\linewidth\else\width\fi}{!}{%
\begin{tabular}{@{}l!{\color{gray!38}\vrule width 0.3pt}ccccc>{\columncolor{metricshade}}c!{\color{gray!38}\vrule width 0.3pt}ccccc>{\columncolor{metricshade}}c!{\color{gray!38}\vrule width 0.3pt}ccccc>{\columncolor{metricshade}}c@{}}
\toprule
\rowcolor{tableheader}
\rule{0pt}{2.4ex}\textbf{Model} & \multicolumn{6}{c}{\textbf{Explicit}} & \multicolumn{6}{c}{\textbf{Implicit}} & \multicolumn{6}{c}{\textbf{Dynamic}} \\
\rowcolor{tableheader}
\rule{0pt}{2.2ex} & P & R & F1 & FPR & ANS & C-F1 & P & R & F1 & FPR & ANS & C-F1 & P & R & F1 & FPR & ANS & C-F1 \\
\midrule
\rowcolor{tableclosed}\multicolumn{19}{@{}l@{}}{\rule{0pt}{2.25ex}\hspace{0.4em}\textit{Proprietary Models}} \\
Gemini-3.1-Pro & 54.69 & \textbf{68.03} & \underline{60.63} & 59.37 & 41.25 & 25.01 & 45.91 & 47.24 & 46.57 & 56.49 & 27.20 & 12.67 & 50.21 & 55.03 & 52.51 & 64.44 & 33.65 & 17.67 \\
GPT-Realtime-2.1 & \underline{57.03} & \underline{63.50} & 60.09 & 51.52 & 55.53 & 33.37 & \textbf{49.81} & \underline{68.90} & \underline{57.82} & 70.04 & \underline{49.80} & \textbf{28.79} & \textbf{54.90} & \underline{65.55} & \textbf{59.75} & 63.23 & 43.11 & \textbf{25.76} \\
Gemini-3.8-Flash & \textbf{72.18} & 60.37 & \textbf{65.75} & 32.78 & \underline{73.90} & \textbf{48.59} & 35.61 & 35.72 & 35.66 & 65.59 & \underline{49.80} & \underline{17.76} & 43.63 & 42.00 & 42.80 & 66.43 & \textbf{58.46} & \underline{25.02} \\
\rowcolor{tableopen}\multicolumn{19}{@{}l@{}}{\rule{0pt}{2.25ex}\hspace{0.4em}\textit{Open-Source Offline Models}} \\
Qwen3.5-9B & 53.43 & 23.97 & 33.09 & 18.94 & 69.09 & 22.86 & \underline{48.11} & \textbf{76.44} & \textbf{59.05} & 82.85 & 23.04 & 13.61 & \underline{52.34} & \textbf{67.28} & \underline{58.88} & 70.42 & 31.84 & 18.75 \\
JoyAI-VL-Interaction & 47.90 & 49.78 & 48.82 & 61.53 & 68.28 & 33.33 & 27.04 & 31.40 & 29.06 & 85.19 & 21.03 & 6.11 & 37.12 & 26.27 & 30.77 & 52.47 & 34.70 & 10.68 \\
\rowcolor{tableopen}\multicolumn{19}{@{}l@{}}{\rule{0pt}{2.25ex}\hspace{0.4em}\textit{Open-Source Streaming Models}} \\
MMDuet2 & 4.77 & 10.91 & 6.64 & \underline{7.13} & \textbf{75.87} & 5.04 & 5.49 & 10.90 & 7.30 & 11.10 & 24.07 & 1.76 & 7.12 & 19.35 & 10.41 & 6.95 & 38.66 & 4.02 \\
Mage-VL & 1.97 & 11.93 & 3.38 & 10.19 & 9.84 & 0.33 & 7.58 & 18.40 & 10.74 & 9.78 & 9.65 & 1.04 & 3.05 & 13.48 & 4.97 & 10.38 & 7.07 & 0.35 \\
LiveStar & 2.95 & 10.75 & 4.63 & 7.93 & 27.01 & 1.25 & 7.31 & 9.75 & 8.36 & \underline{5.98} & 3.85 & 0.32 & 2.61 & 4.63 & 3.34 & \underline{3.73} & 10.36 & 0.35 \\
ViSpeak & 2.16 & 19.01 & 3.88 & 19.12 & 64.28 & 2.49 & 5.52 & 19.96 & 8.65 & 19.06 & 10.99 & 0.95 & 4.86 & 26.02 & 8.19 & 19.10 & 16.76 & 1.37 \\
VideoLLM-Online & 2.53 & 7.07 & 3.73 & \textbf{6.56} & 29.77 & 1.11 & 6.66 & 2.73 & 3.87 & \textbf{1.71} & 19.82 & 0.77 & 15.24 & 12.59 & 13.79 & \textbf{1.51} & 5.52 & 0.76 \\
\rowcolor{tableours}\multicolumn{19}{@{}l@{}}{\rule{0pt}{2.25ex}\hspace{0.4em}\textit{\textbf{Ours}}} \\
\textbf{StreamVibe-VL (SFT)} & 47.87 & 52.70 & 50.17 & 69.48 & 67.37 & \underline{33.80} & 33.89 & 27.44 & 30.33 & 54.36 & 46.02 & 13.96 & 40.67 & 36.65 & 38.56 & 63.87 & 45.61 & 17.59 \\
\textbf{StreamVibe-VL (RL)} & 47.36 & 49.35 & 48.33 & 67.32 & 69.37 & 33.53 & 31.78 & 28.02 & 29.78 & 46.10 & \textbf{50.35} & 14.99 & 39.12 & 37.49 & 38.29 & 57.34 & \underline{49.03} & 18.77 \\
\bottomrule
\end{tabular}}

\caption{\footnotesize \textbf{\textit{Pointwise evaluation with self-generated interaction history}} on StreamVibe-Bench without the Harness ($\pm0$\,s tolerance).}
\label{tab:app_pointwise_self_tau0}
\end{table}

\begin{table}[!htbp]
\centering
\setlength{\tabcolsep}{2.8pt}
\renewcommand{\arraystretch}{1.05}
\arrayrulecolor{gray!38}
\resizebox{\ifdim\width>0.99\linewidth 0.99\linewidth\else\width\fi}{!}{%
\begin{tabular}{@{}l!{\color{gray!38}\vrule width 0.3pt}ccccc>{\columncolor{metricshade}}c!{\color{gray!38}\vrule width 0.3pt}ccccc>{\columncolor{metricshade}}c!{\color{gray!38}\vrule width 0.3pt}ccccc>{\columncolor{metricshade}}c@{}}
\toprule
\rowcolor{tableheader}
\rule{0pt}{2.4ex}\textbf{Model} & \multicolumn{6}{c}{\textbf{Explicit}} & \multicolumn{6}{c}{\textbf{Implicit}} & \multicolumn{6}{c}{\textbf{Dynamic}} \\
\rowcolor{tableheader}
\rule{0pt}{2.2ex} & P & R & F1 & FPR & ANS & C-F1 & P & R & F1 & FPR & ANS & C-F1 & P & R & F1 & FPR & ANS & C-F1 \\
\midrule
\rowcolor{tableclosed}\multicolumn{19}{@{}l@{}}{\rule{0pt}{2.25ex}\hspace{0.4em}\textit{Proprietary Models}} \\
Gemini-3.1-Pro & \underline{32.74} & \textbf{79.21} & \underline{46.33} & 4.60 & \underline{85.60} & \underline{39.66} & 17.73 & 58.78 & 27.24 & 12.80 & \underline{52.12} & \underline{14.20} & 14.23 & 57.82 & \underline{22.84} & 8.92 & \underline{64.95} & \underline{14.83} \\
GPT-Realtime-2.1 & 9.70 & 55.56 & 16.52 & 11.63 & 66.84 & 11.04 & 14.91 & 56.26 & 23.57 & 15.71 & 48.21 & 11.36 & 9.69 & 52.67 & 16.37 & 14.49 & 46.86 & 7.67 \\
Gemini-3.8-Flash & \textbf{47.69} & \underline{79.10} & \textbf{59.50} & \textbf{2.37} & \textbf{87.49} & \textbf{52.06} & \textbf{21.03} & 71.82 & \textbf{32.53} & 12.94 & \textbf{59.96} & \textbf{19.50} & \underline{16.00} & 66.17 & \textbf{25.77} & 8.42 & \textbf{72.48} & \textbf{18.68} \\
\rowcolor{tableopen}\multicolumn{19}{@{}l@{}}{\rule{0pt}{2.25ex}\hspace{0.4em}\textit{Open-Source Offline Models}} \\
Qwen3.5-9B & 7.94 & 52.48 & 13.79 & 14.56 & 44.98 & 6.20 & 9.37 & 47.62 & 15.66 & 19.30 & 16.63 & 2.60 & 4.87 & 51.65 & 8.90 & 20.95 & 27.67 & 2.46 \\
JoyAI-VL-Interaction & 4.39 & 37.96 & 7.87 & 18.87 & 46.83 & 3.69 & 4.67 & \textbf{95.44} & 8.90 & 90.87 & 10.21 & 0.91 & 3.15 & \textbf{76.70} & 6.05 & 64.52 & 21.64 & 1.31 \\
\rowcolor{tableopen}\multicolumn{19}{@{}l@{}}{\rule{0pt}{2.25ex}\hspace{0.4em}\textit{Open-Source Streaming Models}} \\
MMDuet2 & 15.98 & 36.50 & 22.23 & \underline{4.13} & 74.26 & 16.51 & 19.45 & 38.65 & 25.88 & 7.90 & 22.71 & 5.88 & 13.06 & 35.51 & 19.10 & 5.24 & 39.10 & 7.47 \\
Mage-VL & 8.56 & 51.94 & 14.70 & 5.50 & 8.19 & 1.20 & \underline{20.70} & 50.25 & 29.32 & 6.11 & 10.23 & 3.00 & 11.97 & 52.88 & 19.52 & 6.20 & 6.10 & 1.19 \\
LiveStar & 6.84 & 24.95 & 10.74 & 6.27 & 25.43 & 2.73 & 14.96 & 19.95 & 17.10 & \underline{4.81} & 3.60 & 0.62 & 7.29 & 12.92 & 9.32 & \underline{2.86} & 8.05 & 0.75 \\
ViSpeak & 8.41 & 74.03 & 15.10 & 12.67 & 62.98 & 9.51 & 20.66 & \underline{74.71} & \underline{32.37} & 12.75 & 10.19 & 3.30 & 13.51 & \underline{72.32} & 22.77 & 14.19 & 16.85 & 3.84 \\
VideoLLM-Online & 6.33 & 17.66 & 9.32 & 5.32 & 31.88 & 2.97 & 12.29 & 5.04 & 7.15 & \textbf{1.44} & 17.89 & 1.28 & \textbf{17.86} & 14.75 & 16.16 & \textbf{1.28} & 8.69 & 1.40 \\
\rowcolor{tableours}\multicolumn{19}{@{}l@{}}{\rule{0pt}{2.25ex}\hspace{0.4em}\textit{\textbf{Ours}}} \\
\textbf{StreamVibe-VL (SFT)} & 11.55 & 40.66 & 17.99 & 11.40 & 57.97 & 10.43 & 13.13 & 47.91 & 20.61 & 15.59 & 38.06 & 7.84 & 5.38 & 56.76 & 9.83 & 26.13 & 35.88 & 3.53 \\
\textbf{StreamVibe-VL (RL)} & 15.80 & 41.68 & 22.91 & 8.27 & 60.07 & 13.76 & 18.02 & 43.64 & 25.51 & 9.68 & 41.74 & 10.65 & 7.60 & 52.58 & 13.28 & 17.11 & 39.55 & 5.25 \\
\bottomrule
\end{tabular}}

\caption{\footnotesize \textbf{\textit{Trajectory evaluation with self-generated interaction history}} on StreamVibe-Bench without the Harness ($\pm2$\,s tolerance).}
\label{tab:app_trajectory_tau2}
\end{table}

\begin{table}[!htbp]
\centering
\setlength{\tabcolsep}{2.8pt}
\renewcommand{\arraystretch}{0.95}
\arrayrulecolor{gray!38}
\resizebox{\ifdim\width>0.99\linewidth 0.99\linewidth\else\width\fi}{!}{%
\begin{tabular}{@{}l!{\color{gray!38}\vrule width 0.3pt}ccccc>{\columncolor{metricshade}}c!{\color{gray!38}\vrule width 0.3pt}ccccc>{\columncolor{metricshade}}c!{\color{gray!38}\vrule width 0.3pt}ccccc>{\columncolor{metricshade}}c@{}}
\toprule
\rowcolor{tableheader}
\rule{0pt}{2.4ex}\textbf{Model} & \multicolumn{6}{c}{\textbf{Explicit}} & \multicolumn{6}{c}{\textbf{Implicit}} & \multicolumn{6}{c}{\textbf{Dynamic}} \\
\rowcolor{tableheader}
\rule{0pt}{2.2ex} & P & R & F1 & FPR & ANS & C-F1 & P & R & F1 & FPR & ANS & C-F1 & P & R & F1 & FPR & ANS & C-F1 \\
\midrule
\rowcolor{tableclosed}\multicolumn{19}{@{}l@{}}{\rule{0pt}{2.25ex}\hspace{0.4em}\textit{Proprietary Models}} \\
Gemini-3.1-Pro & \underline{19.21} & \textbf{46.49} & \underline{27.19} & 8.43 & \underline{85.54} & \underline{23.26} & \underline{9.01} & 29.88 & \underline{13.85} & 16.13 & \underline{55.51} & \underline{7.69} & 7.74 & 31.46 & 12.42 & 11.71 & \underline{69.94} & \underline{8.69} \\
GPT-Realtime-2.1 & 4.84 & 27.70 & 8.24 & 14.90 & 71.20 & 5.87 & 6.62 & 24.96 & 10.46 & 19.32 & 52.70 & 5.51 & 4.34 & 23.61 & 7.33 & 17.57 & 51.95 & 3.81 \\
Gemini-3.8-Flash & \textbf{27.51} & \underline{45.63} & \textbf{34.33} & \textbf{6.29} & \textbf{88.34} & \textbf{30.33} & \textbf{10.60} & \underline{36.20} & \textbf{16.40} & 17.04 & \textbf{64.64} & \textbf{10.60} & \underline{8.52} & 35.23 & \underline{13.72} & 11.70 & \textbf{76.23} & \textbf{10.46} \\
\rowcolor{tableopen}\multicolumn{19}{@{}l@{}}{\rule{0pt}{2.25ex}\hspace{0.4em}\textit{Open-Source Offline Models}} \\
Qwen3.5-9B & 3.69 & 24.41 & 6.41 & 17.85 & 44.26 & 2.84 & 4.75 & 24.15 & 7.94 & 22.00 & 17.36 & 1.38 & 2.34 & 24.81 & 4.28 & 23.79 & 29.24 & 1.25 \\
JoyAI-VL-Interaction & 3.23 & 27.92 & 5.79 & 20.05 & 41.05 & 2.38 & 4.53 & \textbf{92.53} & 8.64 & 91.20 & 10.04 & 0.87 & 2.92 & \textbf{70.92} & 5.61 & 65.11 & 21.13 & 1.19 \\
\rowcolor{tableopen}\multicolumn{19}{@{}l@{}}{\rule{0pt}{2.25ex}\hspace{0.4em}\textit{Open-Source Streaming Models}} \\
MMDuet2 & 4.77 & 10.91 & 6.64 & 7.13 & 75.87 & 5.04 & 5.49 & 10.90 & 7.30 & 11.10 & 24.07 & 1.76 & 7.12 & 19.35 & 10.41 & 6.95 & 38.66 & 4.02 \\
Mage-VL & 1.97 & 11.93 & 3.38 & 10.19 & 9.84 & 0.33 & 7.58 & 18.40 & 10.74 & 9.78 & 9.65 & 1.04 & 3.05 & 13.48 & 4.97 & 10.38 & 7.07 & 0.35 \\
LiveStar & 2.95 & 10.75 & 4.63 & 7.93 & 27.01 & 1.25 & 7.31 & 9.75 & 8.36 & \underline{5.98} & 3.85 & 0.32 & 2.61 & 4.63 & 3.34 & \underline{3.73} & 10.36 & 0.35 \\
ViSpeak & 2.16 & 19.01 & 3.88 & 19.12 & 64.28 & 2.49 & 5.52 & 19.96 & 8.65 & 19.06 & 10.99 & 0.95 & 4.86 & 26.02 & 8.19 & 19.10 & 16.76 & 1.37 \\
VideoLLM-Online & 2.53 & 7.07 & 3.73 & \underline{6.56} & 29.77 & 1.11 & 6.66 & 2.73 & 3.87 & \textbf{1.71} & 19.82 & 0.77 & \textbf{15.24} & 12.59 & \textbf{13.79} & \textbf{1.51} & 5.52 & 0.76 \\
\rowcolor{tableours}\multicolumn{19}{@{}l@{}}{\rule{0pt}{2.25ex}\hspace{0.4em}\textit{\textbf{Ours}}} \\
\textbf{StreamVibe-VL (SFT)} & 6.47 & 22.79 & 10.08 & 13.49 & 52.78 & 5.32 & 7.62 & 27.78 & 11.96 & 17.91 & 38.43 & 4.60 & 3.92 & \underline{41.33} & 7.16 & 27.76 & 35.18 & 2.52 \\
\textbf{StreamVibe-VL (RL)} & 7.90 & 20.84 & 11.46 & 10.71 & 56.67 & 6.49 & \underline{9.01} & 21.81 & 12.75 & 12.20 & 42.85 & 5.46 & 4.98 & 34.48 & 8.70 & 19.03 & 41.00 & 3.57 \\
\bottomrule
\end{tabular}}

\caption{\footnotesize \textbf{\textit{Trajectory evaluation with model-generated interaction history}} on StreamVibe-Bench without the Harness ($\pm0$\,s tolerance).}
\label{tab:app_trajectory_tau0}
\end{table}

\subsection{General Timing and External Benchmarks}
\label{app:external-benchmarks}

The general-timing results use Pointwise or Trajectory evaluation as indicated. EgoProactive, IPIBench, and OVO-Bench retain their benchmark-specific evaluation protocols.

\begin{table}[!htbp]
\centering
\captionsetup{font=normalsize,skip=3pt}
\scriptsize
\setlength{\tabcolsep}{2.8pt}
\renewcommand{\arraystretch}{0.90}
\arrayrulecolor{gray!38}
\resizebox{\ifdim\width>0.99\linewidth 0.99\linewidth\else\width\fi}{!}{%
\begin{tabular}{@{}l!{\color{gray!38}\vrule width 0.3pt}ccccc>{\columncolor{metricshade}}c!{\color{gray!38}\vrule width 0.3pt}ccccc>{\columncolor{metricshade}}c@{}}
\toprule
\rowcolor{tableheader}
\rule{0pt}{2.4ex}\textbf{Model} & \multicolumn{6}{c}{\textbf{$\tau=2$\,s}} & \multicolumn{6}{c}{\textbf{$\tau=0$\,s}} \\
\rowcolor{tableheader}
\rule{0pt}{2.2ex} & P & R & F1 & FPR & ANS & C-F1 & P & R & F1 & FPR & ANS & C-F1 \\
\midrule
\rowcolor{tableclosed}\multicolumn{13}{@{}l@{}}{\rule{0pt}{2.25ex}\hspace{0.4em}\textit{Proprietary Models}} \\
Gemini-3.1-Pro & \underline{73.78} & 78.71 & 76.17 & 34.17 & 47.57 & 36.23 & 65.06 & 69.40 & 67.16 & 47.97 & 49.82 & 33.46 \\
GPT-Realtime-2.1 & 65.29 & 82.16 & 72.76 & 48.30 & 40.64 & 29.57 & 56.15 & 70.66 & 62.57 & 65.34 & 41.09 & 25.71 \\
Gemini-3.8-Flash & \textbf{75.14} & 79.58 & \textbf{77.30} & 35.71 & 47.59 & 36.79 & \textbf{68.79} & 72.85 & 70.76 & 45.70 & 49.06 & 34.71 \\
\rowcolor{tableopen}\multicolumn{13}{@{}l@{}}{\rule{0pt}{2.25ex}\hspace{0.4em}\textit{Open-Source Offline Models}} \\
Qwen3.5-9B & 52.64 & 83.96 & 64.71 & 88.56 & 61.52 & 39.81 & 51.17 & 81.61 & 62.90 & 92.05 & 61.44 & 38.65 \\
JoyAI-VL-Interaction & 69.34 & 86.15 & \underline{76.84} & 53.41 & 67.11 & 51.57 & \underline{67.44} & 83.80 & \textbf{74.74} & 56.90 & 67.67 & 50.58 \\
\rowcolor{tableopen}\multicolumn{13}{@{}l@{}}{\rule{0pt}{2.25ex}\hspace{0.4em}\textit{Open-Source Streaming Models}} \\
MMDuet2 & 37.06 & 45.76 & 40.95 & 9.17 & 49.16 & 20.13 & 14.98 & 18.50 & 16.55 & 14.40 & 45.18 & 7.48 \\
Mage-VL & 35.96 & 37.99 & 36.95 & \underline{4.84} & 22.73 & 8.40 & 9.64 & 10.18 & 9.90 & \underline{10.18} & 22.58 & 2.24 \\
LiveStar & 23.57 & 40.34 & 29.75 & 12.18 & 28.80 & 8.57 & 12.79 & 21.89 & 16.15 & 15.72 & 34.38 & 5.55 \\
ViSpeak & 28.21 & 84.84 & 42.34 & 22.66 & 33.06 & 14.00 & 12.19 & 36.67 & 18.30 & 31.91 & 32.39 & 5.93 \\
VideoLLM-Online & 45.06 & 17.46 & 25.17 & \textbf{3.40} & 12.54 & 3.16 & 30.65 & 11.88 & 17.12 & \textbf{4.47} & 14.40 & 2.47 \\
\rowcolor{tableours}\multicolumn{13}{@{}l@{}}{\rule{0pt}{2.25ex}\hspace{0.4em}\textit{\textbf{Ours}}} \\
\textbf{StreamVibe-VL (SFT)} & 59.39 & \textbf{96.93} & 73.65 & 94.32 & \textbf{71.32} & \underline{52.53} & 59.22 & \underline{96.66} & 73.44 & 94.72 & \textbf{71.35} & \underline{52.40} \\
\textbf{StreamVibe-VL (RL)} & 60.37 & \underline{96.60} & 74.30 & 85.47 & \underline{71.28} & \textbf{52.96} & 59.90 & \textbf{97.91} & \underline{74.33} & 86.53 & \underline{71.30} & \textbf{53.00} \\
\bottomrule
\end{tabular}}

\caption{\scriptsize \textbf{\textit{Pointwise evaluation with gold interaction history}} on general timing without the Harness.}
\label{tab:app_general_pointwise}

\centering
\setlength{\tabcolsep}{2.8pt}
\renewcommand{\arraystretch}{0.90}
\arrayrulecolor{gray!38}
\resizebox{\ifdim\width>0.99\linewidth 0.99\linewidth\else\width\fi}{!}{%
\begin{tabular}{@{}l!{\color{gray!38}\vrule width 0.3pt}ccccc>{\columncolor{metricshade}}c!{\color{gray!38}\vrule width 0.3pt}ccccc>{\columncolor{metricshade}}c@{}}
\toprule
\rowcolor{tableheader}
\rule{0pt}{2.4ex}\textbf{Model} & \multicolumn{6}{c}{\textbf{$\tau=2$\,s}} & \multicolumn{6}{c}{\textbf{$\tau=0$\,s}} \\
\rowcolor{tableheader}
\rule{0pt}{2.2ex} & P & R & F1 & FPR & ANS & C-F1 & P & R & F1 & FPR & ANS & C-F1 \\
\midrule
\rowcolor{tableclosed}\multicolumn{13}{@{}l@{}}{\rule{0pt}{2.25ex}\hspace{0.4em}\textit{Proprietary Models}} \\
Gemini-3.1-Pro & 29.75 & \underline{91.35} & \underline{44.88} & 26.46 & \underline{68.26} & \underline{30.64} & 20.62 & 63.33 & 31.11 & 31.85 & \underline{68.00} & \underline{21.15} \\
GPT-Realtime-2.1 & 30.17 & 68.86 & 41.96 & 17.84 & 60.90 & 25.55 & 17.84 & 40.72 & 24.81 & 23.25 & 61.91 & 15.36 \\
Gemini-3.8-Flash & \textbf{48.33} & 85.50 & \textbf{61.75} & 12.53 & \textbf{72.78} & \textbf{44.94} & \underline{30.38} & 53.75 & \textbf{38.82} & 18.63 & \textbf{74.77} & \textbf{29.03} \\
\rowcolor{tableopen}\multicolumn{13}{@{}l@{}}{\rule{0pt}{2.25ex}\hspace{0.4em}\textit{Open-Source Offline Models}} \\
Qwen3.5-9B & 24.21 & 31.53 & 27.39 & 5.36 & 34.18 & 9.36 & 9.79 & 12.75 & 11.08 & \underline{8.99} & 32.45 & 3.60 \\
JoyAI-VL-Interaction & 30.00 & 70.61 & 42.11 & 26.05 & 54.85 & 23.10 & 20.74 & 48.82 & 29.11 & 30.27 & 51.49 & 14.99 \\
\rowcolor{tableopen}\multicolumn{13}{@{}l@{}}{\rule{0pt}{2.25ex}\hspace{0.4em}\textit{Open-Source Streaming Models}} \\
MMDuet2 & 37.06 & 45.76 & 40.95 & 9.17 & 49.16 & 20.13 & 14.98 & 18.50 & 16.55 & 14.40 & 45.18 & 7.48 \\
Mage-VL & 35.96 & 37.99 & 36.95 & \underline{4.84} & 22.73 & 8.40 & 9.64 & 10.18 & 9.90 & 10.18 & 22.58 & 2.24 \\
LiveStar & 23.57 & 40.34 & 29.75 & 12.18 & 28.80 & 8.57 & 12.79 & 21.89 & 16.15 & 15.72 & 34.38 & 5.55 \\
ViSpeak & 28.21 & 84.84 & 42.34 & 22.66 & 33.06 & 14.00 & 12.19 & 36.67 & 18.30 & 31.91 & 32.39 & 5.93 \\
VideoLLM-Online & \underline{45.06} & 17.46 & 25.17 & \textbf{3.40} & 12.54 & 3.16 & \textbf{30.65} & 11.88 & 17.12 & \textbf{4.47} & 14.40 & 2.47 \\
\rowcolor{tableours}\multicolumn{13}{@{}l@{}}{\rule{0pt}{2.25ex}\hspace{0.4em}\textit{\textbf{Ours}}} \\
\textbf{StreamVibe-VL (SFT)} & 22.95 & 86.92 & 36.31 & 49.13 & 47.87 & 17.38 & 19.51 & \underline{73.89} & 30.87 & 51.65 & 43.09 & 13.30 \\
\textbf{StreamVibe-VL (RL)} & 25.44 & \textbf{92.28} & 39.88 & 44.94 & 53.80 & 21.46 & 20.42 & \textbf{74.06} & \underline{32.01} & 48.47 & 48.06 & 15.38 \\
\bottomrule
\end{tabular}}

\caption{\scriptsize \textbf{\textit{Trajectory evaluation with self-generated interaction history}} on general timing without the Harness.}
\label{tab:app_general_trajectory}
\label{tab:general_timing}

\centering
\setlength{\tabcolsep}{2.8pt}
\renewcommand{\arraystretch}{0.90}
\arrayrulecolor{gray!38}
\resizebox{\ifdim\width>0.99\linewidth 0.99\linewidth\else\width\fi}{!}{%
\begin{tabular}{@{}l!{\color{gray!38}\vrule width 0.3pt}>{\columncolor{metricshade}}cccc!{\color{gray!38}\vrule width 0.3pt}ccc>{\columncolor{metricshade}}c@{}}
\toprule
\rowcolor{tableheader}
\rule{0pt}{2.4ex}\textbf{Model} & \multicolumn{4}{c}{\textbf{EgoProactive}} & \multicolumn{4}{c}{\textbf{IPIBench}} \\
\rowcolor{tableheader}
\rule{0pt}{2.2ex} & Macro & G-mean & Interrupt & Silent & Timing & Underst. & Repeated & Macro \\
\midrule
\rowcolor{tableclosed}\multicolumn{9}{@{}l@{}}{\rule{0pt}{2.25ex}\hspace{0.4em}\textit{Proprietary Models}} \\
Gemini-3.1-Pro & 58.75 & 58.75 & 59.42 & 58.09 & \underline{53.40} & \underline{33.12} & 7.55 & \underline{31.36} \\
GPT-Realtime-2.1 & 45.28 & 45.28 & \underline{67.09} & 23.46 & 41.52 & 18.25 & 8.02 & 22.60 \\
Gemini-3.8-Flash & \textbf{64.96} & \textbf{64.96} & 66.19 & \underline{63.72} & \textbf{58.77} & \textbf{37.82} & \underline{12.74} & \textbf{36.44} \\
\rowcolor{tableopen}\multicolumn{9}{@{}l@{}}{\rule{0pt}{2.25ex}\hspace{0.4em}\textit{Open-Source Offline Models}} \\
Qwen3.5-9B & 46.20 & 43.20 & 62.60 & 29.80 & 32.50 & 9.80 & 5.70 & 16.00 \\
JoyAI-VL-Interaction & 41.00 & 28.60 & \textbf{70.40} & 11.60 & 34.30 & 7.40 & 1.40 & 14.40 \\
\rowcolor{tableopen}\multicolumn{9}{@{}l@{}}{\rule{0pt}{2.25ex}\hspace{0.4em}\textit{Open-Source Streaming Models}} \\
MMDuet2 & 38.21 & 30.96 & 15.82 & 60.61 & 30.88 & 6.49 & 5.66 & 14.34 \\
Mage-VL & 40.52 & 35.48 & 60.10 & 20.95 & 26.65 & 11.47 & 5.19 & 14.44 \\
LiveStar & 53.62 & 53.16 & 46.58 & 60.67 & 14.65 & 3.40 & 0.47 & 6.17 \\
ViSpeak & 45.21 & 43.84 & 34.17 & 56.26 & 19.89 & 0.43 & 0.00 & 6.77 \\
VideoLLM-Online & 44.71 & 39.14 & 23.10 & \textbf{66.32} & 7.67 & 0.38 & 2.36 & 3.47 \\
\rowcolor{tableours}\multicolumn{9}{@{}l@{}}{\rule{0pt}{2.25ex}\hspace{0.4em}\textit{\textbf{Ours}}} \\
\textbf{StreamVibe-VL (SFT)} & 53.04 & 52.20 & 62.45 & 43.63 & 33.80 & 8.30 & \textbf{16.00} & 19.40 \\
\textbf{StreamVibe-VL (RL)} & \underline{59.50} & \underline{59.50} & 58.90 & 60.00 & 39.10 & 20.40 & 9.40 & 23.00 \\
\bottomrule
\end{tabular}}

\caption{\scriptsize Results on EgoProactive and IPIBench without the Harness.}
\label{tab:bench_ego_ipi}
\end{table}

\begin{table}[!htbp]
\centering
\captionsetup{font=normalsize,skip=3pt}
\scriptsize
\setlength{\tabcolsep}{2.8pt}
\renewcommand{\arraystretch}{0.90}
\arrayrulecolor{gray!38}
\resizebox{\ifdim\width>0.99\linewidth 0.99\linewidth\else\width\fi}{!}{%
\begin{tabular}{@{}l!{\color{gray!38}\vrule width 0.3pt}ccc>{\columncolor{metricshade}}c@{}}
\toprule
\rowcolor{tableheader}
\rule{0pt}{2.4ex}\textbf{Model} & \multicolumn{4}{c}{\textbf{OVO-Bench}} \\
\rowcolor{tableheader}
\rule{0pt}{2.2ex} & BT & RT & FAR & Overall \\
\midrule
% \rowcolor{tableclosed}\multicolumn{5}{@{}l@{}}{\rule{0pt}{2.25ex}\hspace{0.4em}\textit{Proprietary Models}} \\
% Gemini-3.8-Flash & \textbf{72.09} & 78.12 & \textbf{64.96} & \textbf{71.72} \\
\rowcolor{tableopen}\multicolumn{5}{@{}l@{}}{\rule{0pt}{2.25ex}\hspace{0.4em}\textit{Open-Source Offline Models}} \\
Qwen3.5-9B & 53.94 & 63.94 & 55.24 & 57.71 \\
JoyAI-VL-Interaction & 47.09 & \underline{67.99} & 38.87 & 51.32 \\
\rowcolor{tableopen}\multicolumn{5}{@{}l@{}}{\rule{0pt}{2.25ex}\hspace{0.4em}\textit{Open-Source Streaming Models}} \\
MMDuet2 & 47.02 & 61.59 & 37.91 & 48.84 \\
Mage-VL & 46.76 & \textbf{81.12} & 43.50 & \textbf{63.94} \\
LiveStar & 42.25 & 55.12 & 4.29 & 33.88 \\
ViSpeak & \textbf{57.00} & 65.80 & 53.61 & \underline{60.55} \\
VideoLLM-Online & 31.99 & 29.58 & 5.69 & 22.42 \\
\rowcolor{tableours}\multicolumn{5}{@{}l@{}}{\rule{0pt}{2.25ex}\hspace{0.4em}\textit{\textbf{Ours}}} \\
\textbf{StreamVibe-VL (SFT)} & \underline{54.12} & 64.12 & \underline{60.10} & 59.44 \\
\textbf{StreamVibe-VL (RL)} & 53.42 & 64.55 & \textbf{62.00} & 59.29 \\
\bottomrule
\end{tabular}}

\caption{\scriptsize Results on OVO-Bench without the Harness.}
\label{tab:bench_ovo}
\end{table}

\subsection{StreamVibe-Harness}

We compare StreamVibe-VL with and without the Harness under Trajectory evaluation at $\pm2$\,s, using the corresponding base model as a reference. The external-benchmark comparison also reports the matched SFT pair evaluated with and without the Harness.

\begin{table}[!htbp]
\centering
\setlength{\tabcolsep}{2.8pt}
\renewcommand{\arraystretch}{0.95}
\arrayrulecolor{gray!38}
\resizebox{\ifdim\width>0.99\linewidth 0.99\linewidth\else\width\fi}{!}{%
\begin{tabular}{@{}l!{\color{gray!38}\vrule width 0.3pt}ccccc>{\columncolor{metricshade}}c!{\color{gray!38}\vrule width 0.3pt}ccccc>{\columncolor{metricshade}}c!{\color{gray!38}\vrule width 0.3pt}ccccc>{\columncolor{metricshade}}c@{}}
\toprule
\rowcolor{tableheader}
\rule{0pt}{2.4ex}\textbf{Model} & \multicolumn{6}{c}{\textbf{Explicit}} & \multicolumn{6}{c}{\textbf{Implicit}} & \multicolumn{6}{c}{\textbf{Dynamic}} \\
\rowcolor{tableheader}
\rule{0pt}{2.2ex} & P & R & F1 & FPR & ANS & C-F1 & P & R & F1 & FPR & ANS & C-F1 & P & R & F1 & FPR & ANS & C-F1 \\
\midrule
\rowcolor{tableours}\multicolumn{19}{@{}l@{}}{\rule{0pt}{2.25ex}\hspace{0.4em}\textit{\textbf{Ours + Harness}}} \\
\textbf{StreamVibe-VL (SFT) + Harness} & \underline{18.49} & 28.51 & \underline{22.43} & \underline{3.80} & \underline{61.79} & \underline{13.86} & \underline{18.20} & 29.39 & 22.48 & \underline{6.33} & 38.93 & 8.75 & \underline{10.01} & 43.20 & \underline{16.25} & \underline{12.25} & 38.98 & \underline{6.33} \\
\textbf{StreamVibe-VL (RL) + Harness} & \textbf{24.57} & 16.20 & 19.53 & \textbf{1.49} & \textbf{72.83} & \textbf{14.22} & \textbf{21.53} & 29.48 & \underline{24.89} & \textbf{2.02} & \underline{40.86} & \underline{10.17} & \textbf{14.80} & 30.65 & \textbf{19.96} & \textbf{5.68} & \textbf{44.62} & \textbf{8.91} \\
\rowcolor{tableours}\multicolumn{19}{@{}l@{}}{\rule{0pt}{2.25ex}\hspace{0.4em}\textit{\textbf{Ours}}} \\
\textbf{StreamVibe-VL (SFT)} & 11.55 & 40.66 & 17.99 & 11.40 & 57.97 & 10.43 & 13.13 & \textbf{47.91} & 20.61 & 15.59 & 38.06 & 7.84 & 5.38 & \textbf{56.76} & 9.83 & 26.13 & 35.88 & 3.53 \\
\textbf{StreamVibe-VL (RL)} & 15.80 & \underline{41.68} & \textbf{22.91} & 8.27 & 60.07 & 13.76 & 18.02 & 43.64 & \textbf{25.51} & 9.68 & \textbf{41.74} & \textbf{10.65} & 7.60 & \underline{52.58} & 13.28 & 17.11 & \underline{39.55} & 5.25 \\
\rowcolor{tableopen}\multicolumn{19}{@{}l@{}}{\rule{0pt}{2.25ex}\hspace{0.4em}\textit{Open-Source Offline Models}} \\
Qwen3.5-9B & 7.94 & \textbf{52.48} & 13.79 & 14.56 & 44.98 & 6.20 & 9.37 & \underline{47.62} & 15.66 & 19.30 & 16.63 & 2.60 & 4.87 & 51.65 & 8.90 & 20.95 & 27.67 & 2.46 \\
\bottomrule
\end{tabular}}

\caption{\footnotesize \textbf{\textit{Trajectory evaluation}} of StreamVibe-VL with and without the Harness on StreamVibe-Bench ($\pm2$\,s tolerance).}
\label{tab:app_harness_streamvibe}
\end{table}

\begin{table}[!htbp]
\centering
\setlength{\tabcolsep}{2.8pt}
\renewcommand{\arraystretch}{0.95}
\arrayrulecolor{gray!38}
\resizebox{\ifdim\width>0.99\linewidth 0.99\linewidth\else\width\fi}{!}{%
\begin{tabular}{@{}l!{\color{gray!38}\vrule width 0.3pt}ccccc>{\columncolor{metricshade}}c!{\color{gray!38}\vrule width 0.3pt}ccc>{\columncolor{metricshade}}c!{\color{gray!38}\vrule width 0.3pt}>{\columncolor{metricshade}}cccc@{}}
\toprule
\rowcolor{tableheader}
\rule{0pt}{2.4ex}\textbf{Model} & \multicolumn{6}{c}{\textbf{General timing}} & \multicolumn{4}{c}{\textbf{IPIBench}} & \multicolumn{4}{c}{\textbf{EgoProactive}} \\
\rowcolor{tableheader}
\rule{0pt}{2.2ex} & P & R & F1 & FPR & ANS & C-F1 & Timing & Underst. & Repeated & Macro & Macro & G-mean & Interrupt & Silent \\
\midrule
\rowcolor{tableours}\multicolumn{15}{@{}l@{}}{\rule{0pt}{2.25ex}\hspace{0.4em}\textit{\textbf{Ours + Harness}}} \\
\textbf{StreamVibe-VL (SFT) + Harness} & \underline{48.39} & 68.25 & \underline{56.63} & 12.57 & \underline{61.44} & \underline{34.79} & \underline{44.13} & 15.67 & \underline{38.68} & \underline{32.83} & 46.71 & 43.48 & \textbf{63.75} & 29.66 \\
\textbf{StreamVibe-VL (RL) + Harness} & \textbf{53.23} & 67.22 & \textbf{59.41} & \underline{6.84} & \textbf{62.88} & \textbf{37.36} & \textbf{45.34} & \underline{16.88} & \textbf{40.05} & \textbf{35.67} & \underline{48.79} & \underline{44.0} & 3.87 & \underline{29.80} \\
\rowcolor{tableours}\multicolumn{15}{@{}l@{}}{\rule{0pt}{2.25ex}\hspace{0.4em}\textit{\textbf{Ours}}} \\
\textbf{StreamVibe-VL (SFT)} & 22.95 & \underline{86.92} & 36.31 & 49.13 & 47.87 & 17.38 & 34.85 & 13.40 & 28.77 & 25.67 & 35.74 & 23.33 & \underline{62.82} & 8.66 \\
\textbf{StreamVibe-VL (RL)} & 25.44 & \textbf{92.28} & 39.88 & 44.94 & 53.80 & 21.46 & 39.10 & \textbf{20.40} & 29.40 & 23.00 & \textbf{53.04} & \textbf{52.20} & 62.45 & \textbf{43.63} \\
\rowcolor{tableopen}\multicolumn{15}{@{}l@{}}{\rule{0pt}{2.25ex}\hspace{0.4em}\textit{Open-Source Offline Models}} \\
Qwen3.5-9B & 24.21 & 31.53 & 27.39 & \textbf{5.36} & 34.18 & 9.36 & 32.50 & 9.80 & 5.70 & 16.00 & 46.20 & 43.20 & 62.60 & \underline{29.80} \\
\bottomrule
\end{tabular}}

\caption{\footnotesize StreamVibe-Harness comparison on general timing (Trajectory, $\pm2$\,s), IPIBench, and EgoProactive.}
\label{tab:app_harness_external}
\end{table}

% END embedded appendix_results.tex

% The former latency figure is not included: runs use different hardware.
% \begin{figure}[t]
% \centering
% \includegraphics[width=\linewidth]{figure/efficiency_quality_cost.pdf}
% \caption{\textbf{Recorded quality--cost trade-off.} Results use the same 400
% short-form trajectories (200 Explicit Intents and 200 Implicit Intents, with 50 from
% each implicit domain) and 28,159 causal 1-FPS decision steps under
% Trajectory evaluation at $\tau=2$\,s. Crosses denote aggregate estimates,
% small points denote paired sample-bootstrap estimates, and equal-sized
% circles identify methods rather than uncertainty. ANS is conditional on
% judged matched responses (at least 99.9\% coverage for every model) and is
% computed from qualified traces on the same sample IDs. Processing time
% excludes model loading and, for Mage-VL, compatibility
% recovery; unlogged background work is also excluded. The StreamVibe-VL logs
% and dedicated native replays do not constitute a controlled-hardware latency
% benchmark.}
% \label{fig:efficiency}
% \end{figure}

\Needspace{12\baselineskip}
\subsection{Category Difficulty Across Intent Types}
\label{app:category-difficulty}

\begin{figure}[!htbp]
\centering
\includegraphics[width=\linewidth]{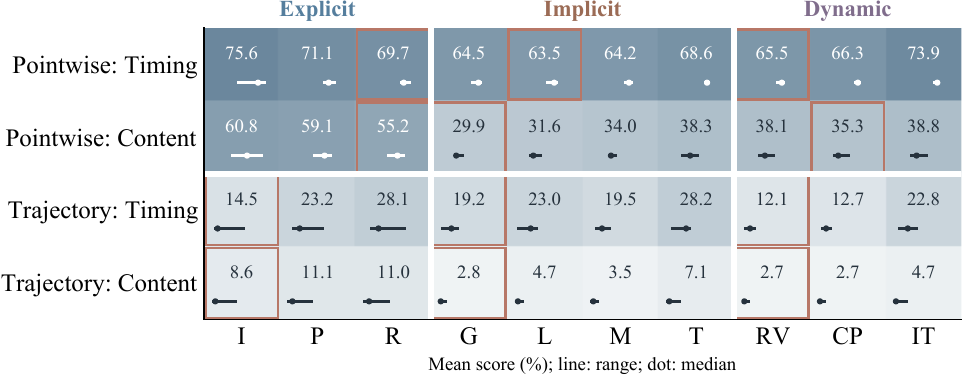}
\caption{\textbf{Category difficulty across StreamVibe-Bench.} Each cell gives
the mean score across all models available in that setting; the line spans the
per-model range and the dot marks the median. Outlines identify the lowest
mean within each intent type and metric. Pointwise uses gold interaction history
($n{=}7$ models), whereas Trajectory uses model-generated interaction history
and also includes native streaming baselines ($n{=}12$ models). All results use
the $\pm2$\,s tolerance.}
\label{fig:category-difficulty}
\end{figure}

\begin{figure}[!htbp]
\centering
\includegraphics[width=\linewidth]{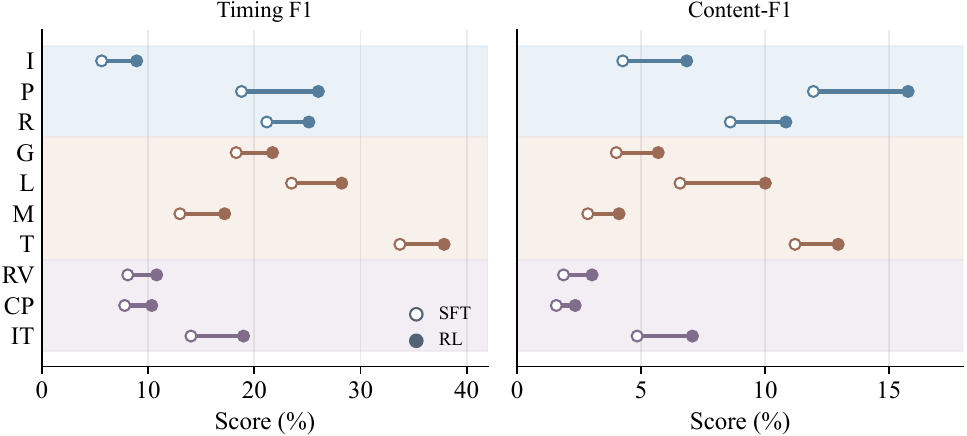}
\caption{\textbf{Absolute subtype profiles of StreamVibe-VL (SFT) and StreamVibe-VL (RL) under
Trajectory evaluation.} Hollow and filled endpoints show SFT and RL,
respectively; the connecting segment exposes both direction and magnitude.
Background bands encode the three intent types.  Scores are percentage points
at $\pm2$\,s.}
\label{fig:subcategory-profiles}
\end{figure}

\begin{figure}[!htbp]
\centering
\includegraphics[width=\linewidth]{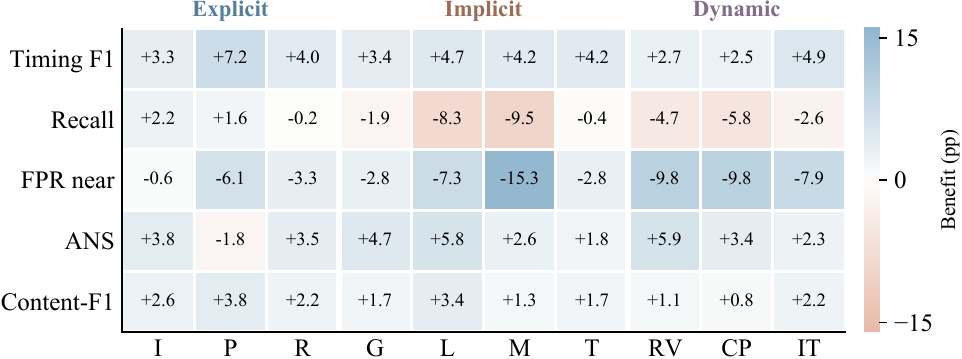}
\caption{\textbf{How RL changes the operating point by subtype.} Cells report
StreamVibe-VL (RL) minus StreamVibe-VL (SFT) in percentage points under Trajectory evaluation.
Numbers retain the metric's natural sign; color encodes benefit, so a negative
$\mathrm{FPR}_{\rm near}$ change is blue.  The visualization jointly exposes
timing, coverage, false interventions, answer quality, and Content-F1.}
\label{fig:subcategory-deltas}
\end{figure}

Building on the overall experimental results above, we further analyze performance across different subtypes to obtain a more fine-grained understanding of proactive interaction capabilities. Specifically, we examine \emph{(i) the relative difficulty of different subtypes for existing systems}, \emph{(ii) how model performance changes across evaluation settings}, and \emph{(iii) how RL shifts response behavior and performance trade-offs across subtypes}. All analyses use a matching tolerance of $\pm2$\,s. Figures~\ref{fig:category-difficulty}--\ref{fig:subcategory-deltas} report subtype-level results under the Pointwise and Trajectory evaluation settings used in Tables~\ref{tab:app_pointwise_gold_tau2} and~\ref{tab:app_trajectory_tau2}. Figure~\ref{fig:category-difficulty} summarizes the average performance of all available models across subtypes under each evaluation protocol to characterize the overall difficulty of different task types. Pointwise includes seven models, while Trajectory additionally includes five native streaming baselines.

\paragraph{Trajectory evaluation reveals substantially greater difficulty under sustained interaction.}
As shown in Figure~\ref{fig:category-difficulty}, performance drops markedly from Pointwise to Trajectory evaluation across all subtypes. Under Pointwise evaluation, Timing F1 remains relatively high at 63.5--75.6, indicating that models can often recognize locally appropriate response moments when evaluated at isolated timestamps. In contrast, Trajectory Timing F1 falls to 12.1--28.2 and Content-F1 to 2.7--11.1, revealing a substantial gap between local perception and sustained interaction. This suggests that maintaining reliable assistance over a full trajectory requires more than recognizing individual events: models must continuously update active intents, track task state, and regulate responses based on their evolving interaction history.

Figure~\ref{fig:category-difficulty} further shows that subtype difficulty changes under sustained interaction. For Explicit intents, I achieves the highest Pointwise Timing F1 at 75.6 but the lowest Trajectory Timing F1 at 14.5, whereas R reaches 28.1. Among Implicit intents, Tool/GUI remains relatively strong, while Guidance becomes more challenging. For Dynamic intents, Revision and Composition underperform Interruption, highlighting the difficulty of continuously revising and composing active requirements. Overall, Pointwise evaluation primarily reflects local recognition and timing, whereas Trajectory evaluation exposes the additional challenges of maintaining coherent and reliable behavior throughout continuous, evolving interaction.

\paragraph{Different subtypes exhibit distinct performance patterns.}
As shown in Figure~\ref{fig:subcategory-profiles}, different subtypes exhibit clear differences in both response timing and content quality. StreamVibe-VL performs best on Tool/GUI, achieving a Timing F1 of 37.9 and a Content-F1 of 13.0, while Monitoring reaches only 17.2 and 4.1, respectively, suggesting that tasks with explicit operational goals are easier to handle reliably than continuous monitoring. For Explicit intents, P and R achieve Timing F1 scores of 26.0 and 25.1, substantially higher than I at 8.9, indicating that single-event cases are more prone to missed or mistimed responses during sustained interaction. For Dynamic intents, Revision and Composition both achieve Timing F1 scores around 10, whereas Interruption reaches 19.0 and also attains higher Content-F1, suggesting that continuously revising or composing existing requirements places greater demands on maintaining the currently valid intent and task state.

\paragraph{RL shifts the policy toward more selective intervention.}

As shown in Figure~\ref{fig:subcategory-deltas}, RL consistently improves Timing F1 and Content-F1 while reducing $\mathrm{FPR}_{\rm near}$ across all subtypes, indicating that its primary effect is to suppress unnecessary triggers and concentrate responses around more appropriate moments. The gains vary across subtypes: P shows the largest Timing F1 improvement of $+7.2$, while Monitoring exhibits the largest FPR reduction of $-15.3$, suggesting that RL is particularly effective in settings prone to over-triggering. RL makes proactive behavior more selective and better aligned with appropriate intervention opportunities across subtypes.

\section{Dataset Construction and Evaluation Details}
\label{app:dataset-details}

This section consolidates the taxonomy, source provenance, training mixture,
evaluation cohorts, and metrics summarized in the main paper.

\subsection{Overview of Evaluation Datasets}
\label{app:evaluation-datasets}

Our evaluation covers proactive interaction and general video understanding.
The former tests when to intervene and whether the response supports the
ongoing task; the latter checks whether proactive policy learning preserves
video perception and temporal reasoning. We describe the datasets below;
protocols and metrics are detailed in Appendix~\ref{app:evaluation-protocols}.

\paragraph{StreamVibe-Bench.}
Our primary benchmark evaluates sustained proactive interaction under
Explicit, Implicit, and Dynamic Intents. Its 4,100 evaluation samples cover
ten subtypes, ranging from responding to specified events to inferring useful
assistance and tracking revised or interleaved requests. Pointwise evaluation
isolates response decisions at individual timestamps, while Trajectory
evaluation tests continuous interaction with model-generated history.
The following subsections describe its taxonomy, sources, and construction.

\paragraph{General-Timing.}
We construct General-Timing as a separate evaluation subset drawn from the
publicly released training data of JoyAI-VL-Interaction~\citep{joyai2026}.
The selected samples are held out from our training: they are used for
neither supervised fine-tuning nor reinforcement learning, and are reserved
exclusively for evaluation. This subset tests response timing and
content quality on additional streaming-video interactions beyond
StreamVibe-Bench. We apply our Pointwise and Trajectory protocols and report results in Tables~\ref{tab:app_general_pointwise}
and~\ref{tab:app_general_trajectory}.

\paragraph{EgoProactive.}
EgoProactive~\citep{kundu2026planwatchrecover} evaluates proactive procedural
assistance from wearable, egocentric videos. Its annotations include
out-of-plan deviations and recovery guidance, requiring an assistant to
recognize task progress, decide when an interruption is useful, and provide
appropriate corrective assistance. We use its official evaluation protocol
to assess proactive assistance in physical task execution.

\paragraph{IPIBench.}
IPIBench~\citep{li2026ipibench} evaluates interactive proactive intelligence
under continuous video streams. It covers proactive monitoring, proactive
task management, and interleaved reactive--proactive requests, including
multi-turn interactions in which users add, modify, or cancel requests.
It complements StreamVibe-Bench by testing the coordination of response
timing and evolving interaction requirements under an external protocol.
We follow its official evaluation setup; results for IPIBench and
EgoProactive are reported in Table~\ref{tab:bench_ego_ipi}.

\paragraph{MVBench.}
MVBench~\citep{li2023mvbench} evaluates video understanding across diverse
temporal tasks that require reasoning beyond a single frame. We use it to
assess whether proactive policy learning preserves general temporal
perception and reasoning, following its official evaluation protocol.

\paragraph{Video-MME.}
Video-MME~\citep{fu2024videomme} assesses video understanding across varied
content domains and video durations. It complements MVBench by testing
broader video comprehension over different temporal scales. We follow its
official evaluation protocol and report the corresponding benchmark score.

\paragraph{OVO-Bench.}
OVO-Bench~\citep{li2025ovobench} evaluates timestamp-aware online video
understanding through backward tracing, real-time perception, and forward
active responding. These settings test whether a model can recall relevant
past events, interpret the current observation, and respond as a stream
unfolds. We retain its official task definitions and evaluation metrics.

\paragraph{OVBench.}
OVBench~\citep{huang2025ovbench} evaluates online video understanding as
visual observations arrive over time. We use it to assess comprehension of
ongoing video streams alongside the proactive interaction benchmarks,
following its official evaluation protocol and reported metrics.

\subsection{Task Taxonomy and Evaluation Sources}

Each example contains a video stream, an initial user instruction, optional
timestamped user turns, and one or more verified evidence points.  An evidence point
stores an acceptable intervention window, a preferred time, and a gold
response.  The ten subtypes in Table~\ref{tab:app-dataset-statistics} describe
what must be tracked rather than merely the visual domain.

\begin{table}[!htbp]
\centering
\small
\setlength{\tabcolsep}{4.2pt}
\renewcommand{\arraystretch}{1.13}
\begin{tabularx}{0.98\textwidth}{@{}llYrrr@{}}
\toprule
\rowcolor{tableheader}
\textbf{Category} & \textbf{Subtype} & \textbf{Operational definition} &
\textbf{Samples} & \textbf{Evidence} & \textbf{Hours} \\
\midrule
\rowcolor{blue!3}
Explicit & I & One requested event or answer; verified Explicit Intents pool & 200 & 200 & 7.5 \\
\rowcolor{blue!3}
 & P & State transition or irreversible point; verified Explicit Intents pool & 200 & 723 & 8.2 \\
\rowcolor{blue!3}
 & R & Repeated occurrences of a requested event; verified Explicit Intents pool & 200 & 794 & 6.4 \\
\midrule
\rowcolor{orange!4}
Implicit & G & Guidance for the next task action & 500 & 3,702 & 26.7 \\
\rowcolor{orange!4}
 & L & Learning and explanation & 500 & 3,287 & 18.4 \\
\rowcolor{orange!4}
 & M & State monitoring for task progress and changes & 500 & 2,701 & 18.9 \\
\rowcolor{orange!4}
 & T & Tool use in software interfaces & 500 & 2,861 & 15.3 \\
\midrule
\rowcolor{purple!4}
Dynamic & RV & Revision replaces an active intent; Dynamic Intents construction pool & 500 & 1,885 & 25.2 \\
\rowcolor{purple!4}
 & CP & Composition adds a intent while preserving prior ones; Dynamic Intents construction pool & 500 & 2,340 & 31.2 \\
\rowcolor{purple!4}
 & IT & Interruption inserts an immediate request while the proactive task remains active; Dynamic Intents construction pool & 500 & 2,053 & 25.6 \\
\midrule
\rowcolor{tableours}
\multicolumn{3}{r}{\textbf{Target balanced evaluation allocation}} &
\textbf{4,100} & \textbf{20,546} & \textbf{183.4} \\
\bottomrule
\end{tabularx}
\caption{\textbf{StreamVibe-Bench target allocation and source coverage.}
``Evidence'' counts proactive gold evidence points; reactive questions are excluded.
Implicit and Dynamic statistics are measured on their balanced manifests.
Explicit evidence-point and duration statistics are proportional estimates from
the qualified I/P/R pool at 200 samples per subtype.}
\label{tab:app-dataset-statistics}
\end{table}

\paragraph{Source coverage.}
Physical-task videos include BARISTA \citep{src_barista},
Assembly101 \citep{src_assembly101}, HoloAssist \citep{src_holoassist},
YouCook2 \citep{src_youcook2}, and Ego-Exo4D \citep{src_egoexo4d}.
Learning sources additionally include ActivityNet \citep{src_activitynet},
Video-MMLU \citep{src_videommlu}, and Video-MMMU \citep{src_videommmu};
monitoring sources include EgoSchema \citep{src_egoschema},
Perception Test \citep{src_perceptiontest}, Charades \citep{src_charades},
and MLVU \citep{src_mlvu}. GUI sources include
AgentNet \citep{src_agentnet}, VideoCUA \citep{src_videocua},
FingerTip \citep{src_fingertip}, GUI-World \citep{src_guiworld},
and GUI-360 \citep{src_gui360}. Source videos are reused with our intent-specific
requests and evidence-point annotations; their original benchmark scores are not
used as StreamVibe-Bench labels.

\paragraph{Annotation and provenance.}
Candidate evidence points are produced from the source video and task record, then
checked by four reviewers for request fidelity, visible evidence, timing, and
assistance content.  Accepted videos are serialized at 1\,FPS; at time $t$, the
record exposes only frames and user turns available by $t$.  The normalized
metadata retains the source collection and video identifier, duration,
subtype, initial and updated requests, evidence-point windows, gold responses,
review record, and negative-probe type.  Dynamic examples additionally retain
the active intent transition, making superseded and newly valid
constraints distinguishable.

\subsection{Human Review Criteria and Inter-Annotator Agreement}
\label{app:human-review}

\paragraph{Review unit and acceptance rule.}
The review unit is a complete candidate sample: the source video, initial
instruction, timestamped user turns, evidence windows, preferred response
times, and reference answers. Four human reviewers independently assess
the same sample against Table~\ref{tab:human-review-rubric}. A reviewer
approves a sample only when all applicable criteria are satisfied; a
required correction counts as non-approval of the current version.
A sample is retained only if all four reviewers approve the same
final version. Thus, three approvals are insufficient. Samples with any
objection must be corrected and reviewed again by all four reviewers, or
discarded; changing an annotation invalidates approval of its earlier
version. Unresolvable ambiguity or insufficient visual evidence requires
exclusion rather than a guessed reference answer.

\begin{table}[!htbp]
\centering
\caption{\textbf{Human review rubric.} Each reviewer checks every applicable
criterion. Acceptance requires four independent approvals of the same
sample version, with no unresolved issue.}
\label{tab:human-review-rubric}
\small
\setlength{\tabcolsep}{5pt}
\renewcommand{\arraystretch}{1.14}
\begin{tabularx}{\textwidth}{@{}P{0.23\textwidth}Y@{}}
\toprule
\rowcolor{tableheader}
\textbf{Criterion} & \textbf{Standard for approval} \\
\midrule
Intent fidelity & The instruction is coherent and compatible with the task.
The assigned Explicit, Implicit, or Dynamic category matches the interaction;
inferred intents are supported by the user's context, not invented. \\
\rowcolor{benchstripe}
Evidence and causality & Every annotated intervention has identifiable visual
or interaction evidence available by its response time. Future frames and
later user turns are not needed to justify the response. \\
Coverage and non-redundancy & Distinct, necessary intervention opportunities
are covered, without duplicate evidence points or repeated assistance after
the relevant intent has already been satisfied. \\
\rowcolor{benchstripe}
Temporal validity & Each response window and preferred timestamp correspond
to a moment when the evidence is sufficient and assistance is still useful;
the preferred time lies within the window and neither anticipates evidence
nor misses the actionable moment. \\
Reference-answer quality & The reference is correct, relevant, and useful for
the active intent. It avoids unsupported details, contradictory advice, and
unnecessary instructions, while remaining consistent with the visible state. \\
\rowcolor{benchstripe}
Dynamic-intent consistency & For Dynamic samples, timestamped revisions,
compositions, and interruptions are coherent. References follow the latest
valid intent and preserve unaffected constraints without acting on superseded
ones; this criterion is not applicable to other categories. \\
Record integrity & The video, source identifier, user-turn timestamps,
evidence windows, and reference answers are complete and aligned. Each
decision refers to an identifiable sample version. \\
\bottomrule
\end{tabularx}
\end{table}

\paragraph{Independent agreement audit.}
Unanimous acceptance is a quality-control rule, not itself evidence of
high inter-annotator agreement. The agreement audit must use the original,
independent judgments on a fixed set of candidate samples \emph{before}
discussion, correction, or acceptance filtering, including candidates that
are not retained. Each reviewer supplies a binary decision, approve or
require revision/reject, under the same rubric. Repeated versions of a
candidate must not be counted as independent samples. The audit should
record the candidate count and decision distribution for each intent type.

\paragraph{Human--human agreement.}
Fleiss' $\kappa$~\citep{fleiss1971agreement} measures chance-corrected
agreement among the four reviewers. For $N$ candidates, let $n_{ic}$ be
the number of reviewers assigning candidate $i$ to decision $c\in\{0,1\}$,
with $\sum_c n_{ic}=4$. Then
\[
\bar P=\frac{1}{N}\sum_{i=1}^{N}
\frac{\sum_c n_{ic}(n_{ic}-1)}{4\cdot3},\qquad
p_c=\frac{\sum_i n_{ic}}{4N},\qquad
\kappa_{\mathrm F}=\frac{\bar P-\sum_c p_c^2}{1-\sum_c p_c^2}.
\]
The audit should additionally report the raw pairwise agreement $\bar P$
and the fraction of candidates with four identical decisions. Computing
$\kappa$ only on retained, unanimously approved samples is invalid as
evidence of general reviewer reliability: filtering enforces agreement,
and a single decision class makes the chance-correction denominator zero.

\paragraph{Gemini--human agreement.}
For this comparison, Gemini-3.8-Flash must independently assign the same
binary review decision to the same frozen candidates, using the rubric
without access to human judgments or post-review corrections. Its original
candidate annotations alone are not binary audit decisions. Compute
Cohen's $\kappa$~\citep{cohen1960agreement} separately between Gemini and
each reviewer, $\kappa_{\mathrm C}^{(r)}=(p_o^{(r)}-p_e^{(r)})/
(1-p_e^{(r)})$, where $p_o^{(r)}$ is observed agreement and $p_e^{(r)}$
is agreement expected from their respective decision marginals. Report
the mean of the four pairwise coefficients and their range, rather than
treating the four human judgments as one independent rater.

\begin{table}[!htbp]
\centering
\caption{\textbf{Inter-annotator agreement.} Human--human agreement is measured by Fleiss' $\kappa$; Gemini--human agreement is the mean of four pairwise Cohen's $\kappa$ coefficients, with ranges in parentheses.}
\label{tab:human-review-agreement}
\small
\setlength{\tabcolsep}{5pt}
\renewcommand{\arraystretch}{1.15}
\begin{tabularx}{\textwidth}{@{}l c >{\centering\arraybackslash}X >{\centering\arraybackslash}X@{}}
\toprule
\rowcolor{tableheader}
\textbf{Audit subset} & \textbf{Candidates} &
\shortstack{\textbf{Human--Human}\\Fleiss' $\kappa\,\uparrow$} &
\shortstack{\textbf{Gemini-3.8-Flash--Human}\\Cohen's $\kappa\,\uparrow$ (mean of four)} \\
\midrule
Explicit & 872 & 0.87 & 0.79\,(0.76--0.82) \\
\rowcolor{benchstripe}
Implicit & 2,946 & 0.84 & 0.75\,(0.72--0.78) \\
Dynamic & 2,187 & 0.82 & 0.73\,(0.70--0.77) \\
\midrule
All candidates & 6,005 & 0.84 & 0.76\,(0.70--0.82) \\
\bottomrule
\end{tabularx}
\end{table}

\paragraph{Analysis.}
As shown in Table~\ref{tab:human-review-agreement}, human--human Fleiss' $\kappa$ ranges from 0.82 to 0.87 across all categories, falling within the ``almost perfect agreement'' band ($>$0.81)~\citep{landis1977agreement}, indicating that the four reviewers maintained highly consistent judgments under the shared rubric. Explicit achieves the highest agreement (0.87), while Dynamic is slightly lower (0.82), consistent with the inherently greater complexity of annotating evolving intents. Gemini--human Cohen's $\kappa$ ranges from 0.73 to 0.79, within the ``substantial agreement'' band (0.61--0.80), with narrow per-pair ranges, indicating that Gemini's review decisions are well aligned with human judgments. Together, these results confirm the reliability and quality of the StreamVibe-Bench annotation pipeline.

\subsection{Training Mixture and Evaluation Cohorts} 
\label{training}

Table~\ref{tab:app-training-sources} reports the SFT data manifest containing 70K task-aligned examples and 768K general streaming examples. Task-aligned Explicit and Implicit examples preserve their original task structure, while the general streaming mixture broadens coverage across visual domains, narration styles, temporal scales, and GUI scenarios. Localized RL starts from a separate balanced 15K seed pool, consisting of 5K tasks each from Implicit Intents, Explicit Intents, and general streaming data.

\begin{table}[!htbp]
\centering
\small
\setlength{\tabcolsep}{4.5pt}
\renewcommand{\arraystretch}{1.13}
\begin{tabularx}{0.96\textwidth}{@{}P{0.21\textwidth}YP{0.10\textwidth}P{0.16\textwidth}@{}}
\toprule
\rowcolor{tableheader}
\textbf{Training component} & \textbf{Recorded source families} & \textbf{Examples} & \textbf{Use} \\
\midrule
Task-aligned Explicit
& Explicit Intents event families
& 35,000
& Streaming SFT \\

\rowcolor{benchstripe}
Task-aligned Implicit
& Guidance, learning, monitoring, and tool/GUI
& 35,000
& Streaming SFT \\

General streaming core
& JoyAI mixture: Live-WhisperX, Shot2Story, TimeLens, GUI-World, VideoGPT+, Charades-Ego, InternVid, ActivityNet, WebVid, and smaller sources
& 763,607
& Streaming SFT \\

\rowcolor{benchstripe}
General streaming auxiliary
& TimeChat-Online, VideoChat-Flash, TimeLens, EgoIT, and StreamO-Instruct
& 4,393
& Streaming SFT \\

\midrule
\rowcolor{tableours}
\multicolumn{2}{r}{\textbf{Full SFT manifest}}
& \textbf{838,000}
& \textbf{1\,FPS streams} \\

Localized RL seed
& Implicit / Explicit / general streaming, 5K each
& 15,000
& Local rollout windows \\

\bottomrule
\end{tabularx}
\caption{\textbf{Training-data provenance.}
The SFT mixture contains 70K task-aligned examples and 768K general streaming examples.
Localized RL uses a separate 15K seed pool, yielding 85K task-oriented training examples in total when combined with task-aligned SFT.
Dataset names indicate source families and do not imply a shared annotation protocol.}
\label{tab:app-training-sources}
\end{table}

\paragraph{General-streaming sources.}
The inherited mixture includes Live-WhisperX~\citep{src_livewhisperx},
Shot2Story~\citep{src_shot2story}, TimeLens~\citep{src_timelens},
VideoGPT+~\citep{src_videogptplus}, Charades-Ego~\citep{src_charadesego},
InternVid~\citep{src_internvid}, WebVid~\citep{src_webvid}, and the
ActivityNet and GUI-World sources cited above. Auxiliary data comes from
TimeChat-Online~\citep{src_timechatonline}, VideoChat-Flash~\citep{src_videochatflash},
TimeLens, EgoIT-99K (released with EgoLife)~\citep{src_egoit}, and
Streamo-Instruct~\citep{src_streamo}.

\subsection{Evaluation Protocols and Metrics}
\label{app:evaluation-protocols}

\begin{table}[!htbp]
\centering
\small
\setlength{\tabcolsep}{4.0pt}
\renewcommand{\arraystretch}{1.13}
\begin{tabularx}{0.98\textwidth}{@{}P{0.17\textwidth}YYP{0.19\textwidth}@{}}
\toprule
\rowcolor{tableheader}
\textbf{Evaluation} & \textbf{What is scored} & \textbf{Reported measures} & \textbf{Judge / scorer} \\
\midrule
Pointwise & Independent probes; supplied history & Precision, recall, F1, $\mathrm{FPR}_{\rm near}$, ANS, Content-F1 & Temporal matcher; Claude Sonnet 4.6~\citep{anthropic2026sonnet46} \\
\rowcolor{benchstripe}
Trajectory & Complete 1\,FPS rollout; generated history & Same six scores at $\pm2$, $\pm1$, $\pm0$\,s & Same matcher and judge \\
General Timing & Streaming transfer cohort & Same timing and content scores & Same protocol \\
\rowcolor{benchstripe}
EgoProactive / IPIBench & Session-level interaction & Macro, G-mean, Interrupt, Silent; Timing, Understanding, Repeated & Official protocols \\
Video suites & MVBench, Video-MME, OVO-Bench, OVBench & Accuracy or official aggregate & Official scorers \\
\bottomrule
\end{tabularx}
\caption{\textbf{Metrics and judges used in the paper.} External benchmarks
retain their official evaluation protocols and are not rescored by the
StreamVibe-Bench content judge.}
\label{tab:app-metrics-judges}
\end{table}

\paragraph{Temporal and content scores.}
At tolerance $\tau$, predictions and evidence points within $\pm\tau$ are matched
chronologically and one-to-one.  Thus
$P=\mathrm{TP}/N_{\rm pred}$,
$R=\mathrm{TP}/N_{\rm gold}$, and F1 is their harmonic mean;
$\mathrm{FPR}_{\rm near}$ measures responses on designated nearby negative
probes.  For each temporally matched response, Claude Sonnet 4.6~\citep{anthropic2026sonnet46} assigns a 0--4
content score $r$ that is normalized to $s_{g,p}=r/4\in[0,1]$.  ANS is its mean over
matched responses.  Content-F1 is
$2\sum_{(g,p)\in\mathcal{M}}s_{g,p}/(N_{\rm pred}+N_{\rm gold})$, so missed
intents and extra interventions remain penalized instead of disappearing from a
conditional answer average.  For Dynamic Intents, ANS and Content-F1 score
proactive responses only; reactive answers remain in the interaction history.

\section{StreamVibe-Harness Implementation Details}
\label{app:harness-details}

This section provides implementation details for the four functional components introduced in Section~\ref{sec:streamvibe-workflow}. StreamVibe-Harness maintains persistent interaction state and organizes causal evidence for the interaction policy. These components jointly support a single interaction process and should not be interpreted as four independent models. We use Qwen3-VL-4B-Instruct~\citep{bai2025qwen3vl} as an auxiliary visual-memory model for memory summarization and consolidation with greedy decoding, while the StreamVibe-VL checkpoint under evaluation serves as the interaction policy. The auxiliary model is used off-the-shelf and receives no additional training.

\begin{table}[htbp]
\centering
\small
\setlength{\tabcolsep}{5pt}
\renewcommand{\arraystretch}{1.08}
\caption{\textbf{Harness configuration.} Temporal parameters correspond to the chunked-streaming setting; benchmark-specific native adapters are described separately below. Retrieval limits denote maximum capacities rather than fixed numbers of items injected into the policy context.}
\label{tab:harness-implementation-config}
\begin{tabular}{@{}p{0.40\linewidth}p{0.54\linewidth}@{}}
\toprule
Component & Setting \\
\midrule
Memory model & Qwen3-VL-4B-Instruct; separate inference process \\
Memory input & At most 64 frames per call; 262,144 pixels per frame \\
Chunk and submission schedule & $T_s=32$ steps; submission lookahead $=8$ steps \\
Consolidation and retention & $M=5$ mid-term summaries; $L=8$ long-term blocks \\
Evidence selection & Up to 3 facts and 3 ordered transitions \\
Compact state evidence & First rendered evidence line, capped at 180 characters \\
Skill recall & 10 fixed Skills; at most 1 selected per context construction \\
Procedural memory & Up to 8 completed steps and 12 temporal anchors \\
\bottomrule
\end{tabular}
\end{table}

\subsection{Interaction State}
\label{app:harness-state}

At time $t$, the interaction state is represented as $\mathcal{S}_t=\langle\mathcal{A}_t,\mathcal{P}_t,\mathcal{R}_t\rangle$, where $\mathcal{A}_t$ stores active intents, $\mathcal{P}_t$ records observed task progress, and $\mathcal{R}_t$ tracks responses that have actually been delivered. State changes are represented as typed events with an identifier, timestamp, source, and payload. A deterministic update function applies valid events to the corresponding state fields and ignores duplicate updates. User instructions may introduce, revise, or retire requests; visual observations may establish task-relevant events; and observed progress or explicit user confirmation may update procedural state. A delivery record is created only after a response has been emitted to the user.

We enforce source-specific update constraints so that information retains its semantic role across the interaction. In particular, a request specifies what should be monitored rather than what has already occurred; a reference plan describes possible future steps rather than completed actions; and an assistant suggestion is not treated as evidence of user progress. Fields without sufficient support remain unresolved. Historical evidence may be exposed to the policy as context, but cannot by itself create, revise, or retire an intent.

Procedural memory retains at most eight completed steps and twelve evidence intervals that serve as temporal retrieval anchors. Progress is updated only when supported by observation or explicit confirmation. For event-monitoring requests, the auxiliary model additionally extracts and normalizes evidence from causal frames within a two-second window. This event parser updates evidence and event state only; it does not generate user-facing responses and therefore does not constitute a second action policy.

\subsection{Intent Lifecycle}
\label{app:harness-lifecycle}

Each active intent is assigned a stable identifier and maintains an independent record of its criterion, associated event, evidence reference, and delivery status. A newly registered criterion begins in an evidence-waiting state. Once sufficient evidence supports a match, the criterion is linked to a concrete event instance. If the policy subsequently responds, the resulting delivery record identifies both the addressed event and the associated intents. This design allows multiple intents to remain active concurrently without treating a response to one intent as completion of the others.

The lifecycle explicitly separates \emph{request validity} from \emph{event coverage}. Answering a one-time question closes that pending question. By contrast, reporting an occurrence under a standing request marks only that event instance as addressed while keeping the request active for future evidence. Closing an event returns the corresponding instance to an evidence-waiting state, whereas explicit user cancellation retires the request itself. These transitions update interaction bookkeeping rather than policy behavior: neither an active intent nor a matched event deterministically triggers a response.

\paragraph{Skill library and selection.}
The Harness maintains ten fixed, read-only Skill records covering visual question answering, multiple-choice answering, event detection, explanation, hazard monitoring, milestone reporting, mistake monitoring, narration, next-step guidance, and stage summarization. Each Skill specifies request cues, evidence to inspect, response and waiting conditions, interaction cadence, and answer form. Skills are reusable behavioral specifications rather than neural models. Each active criterion maintains its own execution instance, including phase, evidence reference, event association, and delivery status.

Skill selection first restricts candidates to those compatible with the current interaction type and requires non-empty observed memory, procedural context, or event context. A pending question prioritizes the question-answering Skill, followed by direct matches to active criterion types. If neither applies, lexical overlap between the request and observed context is used as a fallback, with at least two shared terms required. At most one Skill is selected for each context construction, and selecting no Skill is a valid outcome. For example, next-step guidance is conditioned on observed task progress, whereas event detection tracks a requested transition and whether the corresponding occurrence has already been reported. Importantly, selecting a Skill does not imply that its response condition has been satisfied.

\subsection{Hierarchical Memory}
\label{app:harness-memory}

\paragraph{Storage and consolidation.}
The memory system contains three levels: recent raw visual context, timestamped mid-term summaries, and older consolidated text blocks. Under the chunked-streaming configuration, we use $T_s=32$ steps per chunk, consolidate every $M=5$ mid-term summaries, and retain up to $L=8$ long-term blocks. At one step per second, each chunk corresponds to 32 seconds of video. The auxiliary model processes frames from the short-term buffer in chronological order, uniformly subsampling when the 64-frame input limit is exceeded, and conditions each summary on the current request. Consolidation of older summaries is performed using text-only input.

Mid-term summaries preserve task-relevant object states, quantities, on-screen text, and temporal transitions. Intervals without meaningful change are collapsed, unsupported attributes remain unknown, and the most recently observed state is retained when visually established. Each ordered transition records its temporal interval, actor, action, object, prior and resulting states, temporal relation, and confidence. The mid-term summarization prompt is capped at 220 words, while procedural fields distinguish observed current and completed steps from request-supported references to possible next steps.

Every $M$ consecutive mid-term summaries are consolidated into one long-term block using a 300-word prompt budget. Consolidation removes redundant static descriptions while preserving timestamps, distinct actions, temporal order, and supported procedural information. Each mid-term summary belongs to exactly one non-overlapping consolidation group. Once the number of long-term blocks reaches the capacity $L$, older blocks are evicted in first-in, first-out order.

To preserve causality under asynchronous execution, summarization is submitted before the chunk boundary using only frames already observed at submission time. Frames arriving afterward remain in the next short-term buffer, preventing gaps in temporal coverage. Mid-term inputs remain available while long-term consolidation is pending. If consolidation fails or produces an empty output, their concatenation is retained instead, preserving the available evidence without assuming that generated summaries are lossless.

\paragraph{Indexing and retrieval.}
Summary facts are indexed using normalized text together with stable identifiers, source intervals, and lexical terms. Repeated observations of the same fact update its observation interval rather than creating duplicate entries. Ordered transitions are indexed separately to preserve event sequence. Request descriptions and reference-plan fields are excluded from the factual index, and response records remain separate from visual evidence.

At retrieval time, the Harness constructs a focus string from the pending question, current request, and relevant procedural state. Candidate entries are ranked primarily by lexical overlap with this focus and secondarily by recency, with additional relevance assigned to entries overlapping task-relevant temporal anchors. Facts and ordered transitions are selected independently, with up to three candidates from each category and at most six entries before context projection. If no candidate matches the focus, the most recent entries are retained as a fallback. Selected transitions are reordered chronologically before projection. For native event-monitoring adapters, target retrieval is further restricted to at most three matching evidence lines, excluding negative or uncertain statements. No embedding model or learned retriever is used.

The retrieval capacity is intentionally decoupled from the amount of evidence exposed to the policy. In the compact lifecycle projection, only the first rendered evidence line is retained and truncated to 180 characters; this line may contain a joined excerpt of multiple selected facts. For a pending question, the bounded fact and transition blocks may instead be exposed directly. Retrieved items are always treated as timestamped historical context, and current observations take precedence whenever they supersede an earlier state.

\subsection{State Integration}
\label{app:harness-integration}

Before each policy call, the Harness compiles the relevant state fields and retrieved evidence into a compact textual context and combines it with the causal video input,
\[
\tilde{c}_t = \phi(c_t, \mathcal{S}_t, m_t),
\]
where $c_t$ denotes the causal interaction context, $\mathcal{S}_t$ the persistent interaction state, and $m_t$ the retrieved memory. The projected context explicitly distinguishes pending questions, observed progress, event status, delivery records, uncertainty, and historical evidence. Inactive or empty fields are omitted, and the full memory store is not appended by default. Within a context epoch, the header remains fixed while subsequent observations and actions are appended, allowing the policy to reuse the shared prompt prefix.

\paragraph{Skill Projection for Policy Inference.}
Although each Skill stores a complete behavioral specification, the policy receives only the subset required for the current interaction. For single-criterion guidance, monitoring, and explanation, the projected context includes the selected Skill's execution phase, evidence reference, response basis, and a compact cadence description. Question-answering and multi-criterion contexts expose only the answer-form field, while single-criterion narration retains a compact symbolic representation. The complete Skill specification is therefore not copied into the prompt, and Skill metadata does not replace policy generation. Request changes update the interaction type causally; after an intervening question is answered, the context of the standing request is restored.

The interaction policy $\pi$ remains the sole component that decides whether to remain silent or respond and, when responding, generates the user-facing text. Once a non-empty response is delivered, its timestamp and intent/event associations are committed to $\mathcal{R}_t$, while its exact text is preserved in the interaction history. Subsequent context projections may retain the most recent response verbatim and compress older responses into action markers. Assistant outputs are excluded from visual summarization. Response verification, cadence gating, and post-generation rewriting are disabled. Consequently, this feedback path records delivered assistance for future state construction without introducing new visual evidence or modifying the policy output.

\section{Qualitative Examples and Case Studies}
\label{app:qualitative-examples}

This section presents annotated examples from all ten task subtypes, followed
by additional comparisons of model responses during sustained interaction.

\subsection{Representative Samples from Every Subtype}
\label{app:representative-samples}

\begin{figure}[!htbp]
\centering
\includegraphics[width=0.87\linewidth,trim=0 387.78 0 0.0,clip]{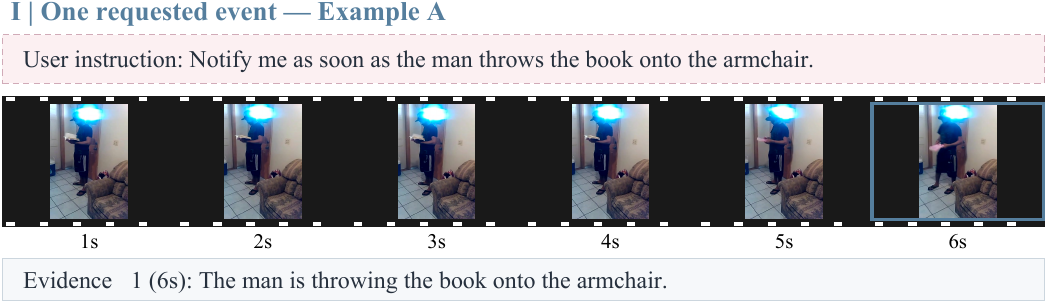}
\caption{\textbf{Explicit intent, single event (I), example A.}}
\label{fig:app-sample-I-A}
\end{figure}

\begin{figure}[!htbp]
\centering
\includegraphics[width=0.87\linewidth,trim=0 421.45 0 0.0,clip]{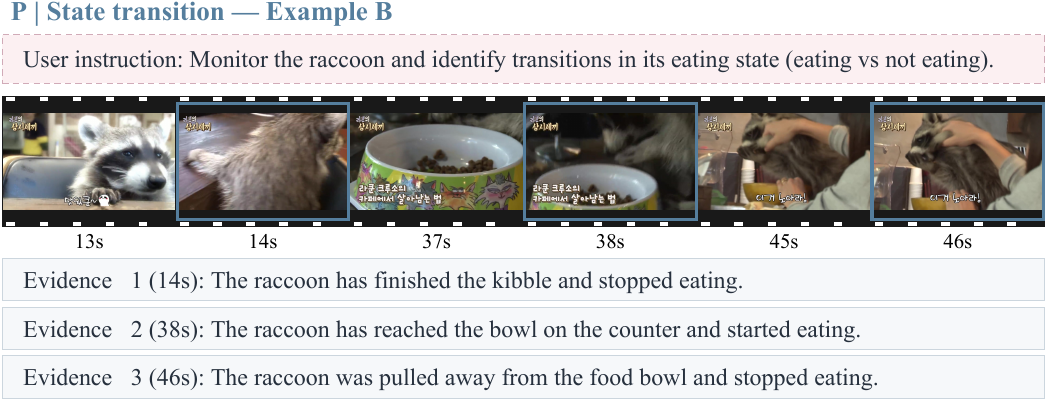}
\caption{\textbf{Explicit intent, state transition (P), example B.}}
\label{fig:app-sample-P-B}
\end{figure}

\begin{figure}[!htbp]
\centering
\includegraphics[width=0.87\linewidth,trim=0 528.43 0 0.0,clip]{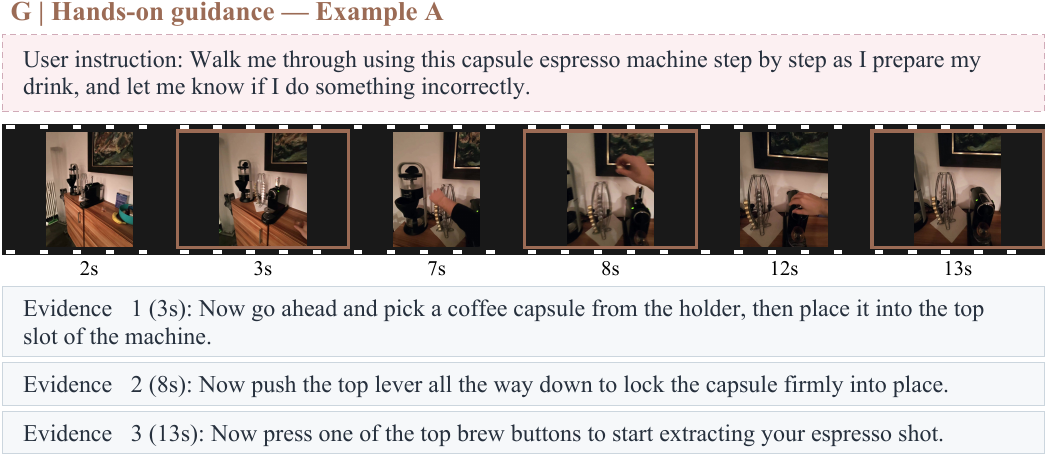}
\caption{\textbf{Implicit intent, guidance (G), example A.}}
\label{fig:app-sample-G-A}
\end{figure}

\begin{figure}[!htbp]
\centering
\includegraphics[width=0.87\linewidth,trim=0 501.8 0 0.0,clip]{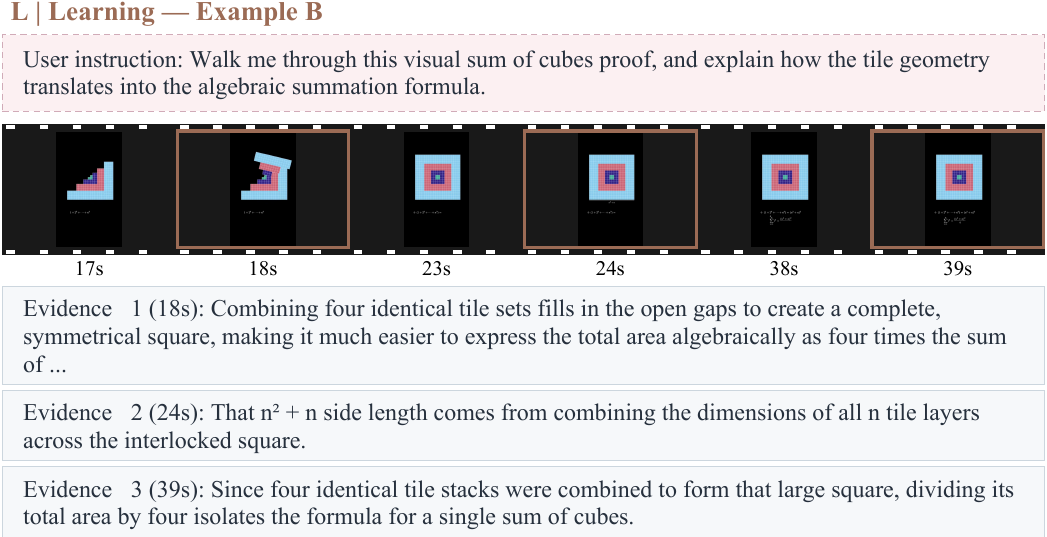}
\caption{\textbf{Implicit intent, learning (L), example B.}}
\label{fig:app-sample-L-B}
\end{figure}

\begin{figure}[!htbp]
\centering
\includegraphics[width=0.87\linewidth,trim=0 517.8 0 0.0,clip]{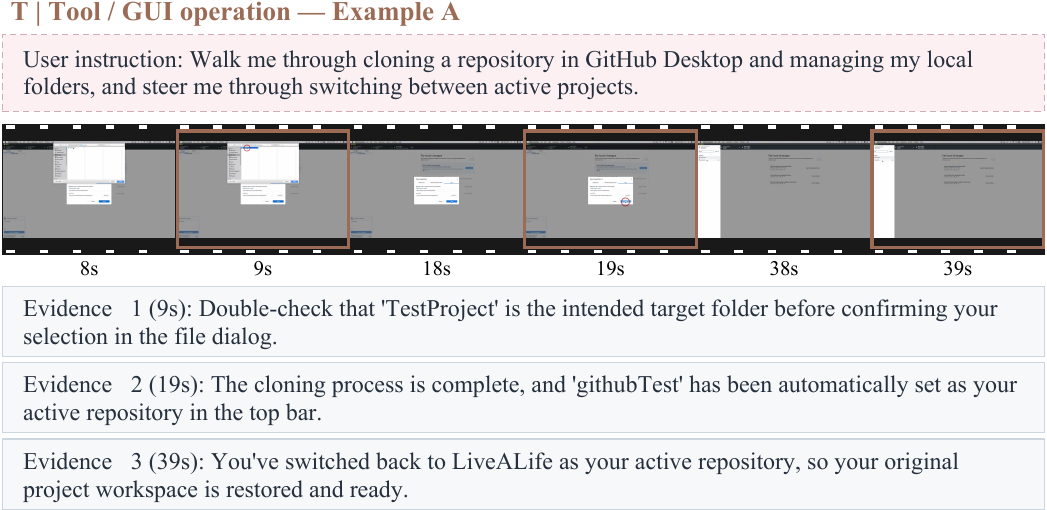}
\caption{\textbf{Implicit intent, tool / GUI (T), example A.}}
\label{fig:app-sample-T-A}
\end{figure}

\begin{figure}[!htbp]
\centering
\includegraphics[width=0.87\linewidth,trim=0 507.45 0 0.0,clip]{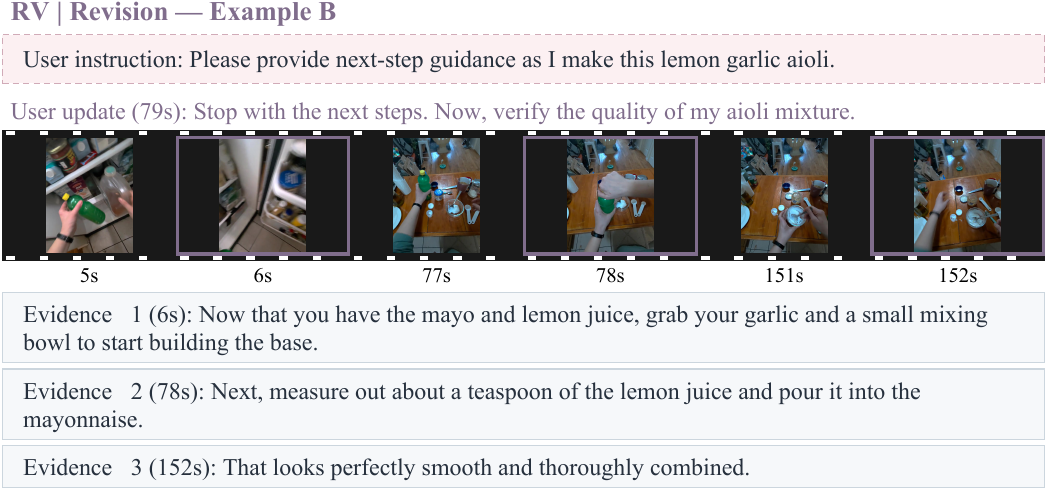}
\caption{\textbf{Dynamic intent, revision (RV), example B.}}
\label{fig:app-sample-RV-B}
\end{figure}

\begin{figure}[!htbp]
\centering
\includegraphics[width=0.87\linewidth,trim=0 286.84 0 0.0,clip]{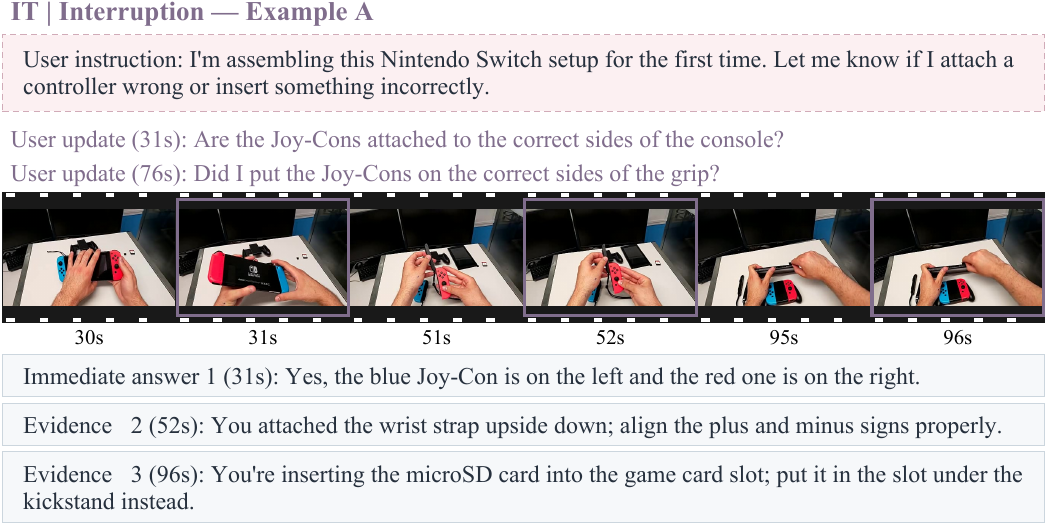}
\caption{\textbf{Dynamic intent, interruption (IT), example A.}}
\label{fig:app-sample-IT-A}
\end{figure}

Figures~\ref{fig:app-sample-I-A}--\ref{fig:app-sample-IT-B} show two
verified samples per subtype. Each full-width sample contains the original
instruction, a chronological six-frame filmstrip, and timestamped reference
assistance. Three annotated interventions are shown per sample, except for
single-event I, whose definition permits only one gold evidence point. Each pair
of adjacent frames shows the second before an intervention and its annotated
time; gaps between pairs are indicated by the timestamps. The outlined frame
marks the intervention. Dynamic examples include the user updates applicable
to the displayed timeline. 

\begin{figure}[!htbp]
\centering
\includegraphics[width=0.87\linewidth,trim=0 223.87 0 163.91,clip]{figure/dataset_examples/filmstrip_sheet_01.pdf}
\caption{\textbf{Explicit intent, single event (I), example B.}}
\label{fig:app-sample-I-B}
\end{figure}

\begin{figure}[!htbp]
\centering
\includegraphics[width=0.87\linewidth,trim=0 0.0 0 327.82,clip]{figure/dataset_examples/filmstrip_sheet_01.pdf}
\caption{\textbf{Explicit intent, state transition (P), example A.}}
\label{fig:app-sample-P-A}
\end{figure}

\begin{figure}[!htbp]
\centering
\includegraphics[width=0.87\linewidth,trim=0 211.72 0 210.73,clip]{figure/dataset_examples/filmstrip_sheet_02.pdf}
\caption{\textbf{Explicit intent, repeated event (R), example A.}}
\label{fig:app-sample-R-A}
\end{figure}

\begin{figure}[!htbp]
\centering
\includegraphics[width=0.87\linewidth,trim=0 0.0 0 420.45,clip]{figure/dataset_examples/filmstrip_sheet_02.pdf}
\caption{\textbf{Explicit intent, repeated event (R), example B.}}
\label{fig:app-sample-R-B}
\end{figure}

\begin{figure}[!htbp]
\centering
\includegraphics[width=0.87\linewidth,trim=0 264.72 0 237.75,clip]{figure/dataset_examples/filmstrip_sheet_03.pdf}
\caption{\textbf{Implicit intent, guidance (G), example B.}}
\label{fig:app-sample-G-B}
\end{figure}

\begin{figure}[!htbp]
\centering
\includegraphics[width=0.87\linewidth,trim=0 0.0 0 501.46,clip]{figure/dataset_examples/filmstrip_sheet_03.pdf}
\caption{\textbf{Implicit intent, learning (L), example A.}}
\label{fig:app-sample-L-A}
\end{figure}

\begin{figure}[!htbp]
\centering
\includegraphics[width=0.87\linewidth,trim=0 264.9 0 277.89,clip]{figure/dataset_examples/filmstrip_sheet_04.pdf}
\caption{\textbf{Implicit intent, monitoring (M), example A.}}
\label{fig:app-sample-M-A}
\end{figure}

\begin{figure}[!htbp]
\centering
\includegraphics[width=0.87\linewidth,trim=0 0.0 0 514.8,clip]{figure/dataset_examples/filmstrip_sheet_04.pdf}
\caption{\textbf{Implicit intent, monitoring (M), example B.}}
\label{fig:app-sample-M-B}
\end{figure}

\begin{figure}[!htbp]
\centering
\includegraphics[width=0.87\linewidth,trim=0 267.89 0 263.9,clip]{figure/dataset_examples/filmstrip_sheet_05.pdf}
\caption{\textbf{Implicit intent, tool / GUI (T), example B.}}
\label{fig:app-sample-T-B}
\end{figure}

\begin{figure}[!htbp]
\centering
\includegraphics[width=0.87\linewidth,trim=0 0.0 0 514.8,clip]{figure/dataset_examples/filmstrip_sheet_05.pdf}
\caption{\textbf{Dynamic intent, revision (RV), example A.}}
\label{fig:app-sample-RV-A}
\end{figure}

\begin{figure}[!htbp]
\centering
\includegraphics[width=0.87\linewidth,trim=0 267.71 0 253.73,clip]{figure/dataset_examples/filmstrip_sheet_06.pdf}
\caption{\textbf{Dynamic intent, composition (CP), example A.}}
\label{fig:app-sample-CP-A}
\end{figure}

\begin{figure}[!htbp]
\centering
\includegraphics[width=0.87\linewidth,trim=0 0.0 0 493.47,clip]{figure/dataset_examples/filmstrip_sheet_06.pdf}
\caption{\textbf{Dynamic intent, composition (CP), example B.}}
\label{fig:app-sample-CP-B}
\end{figure}

\begin{figure}[!htbp]
\centering
\includegraphics[width=0.87\linewidth,trim=0 0.0 0 269.85,clip]{figure/dataset_examples/filmstrip_sheet_07.pdf}
\caption{\textbf{Dynamic intent, interruption (IT), example B.}}
\label{fig:app-sample-IT-B}
\end{figure}

% BEGIN embedded appendix/additional_case_studies.tex
\subsection{Additional Qualitative Case Studies}
\label{app:additional-cases}

\begin{figure}[!htbp]
\centering
\includegraphics[width=\linewidth,height=0.43\textheight,keepaspectratio]{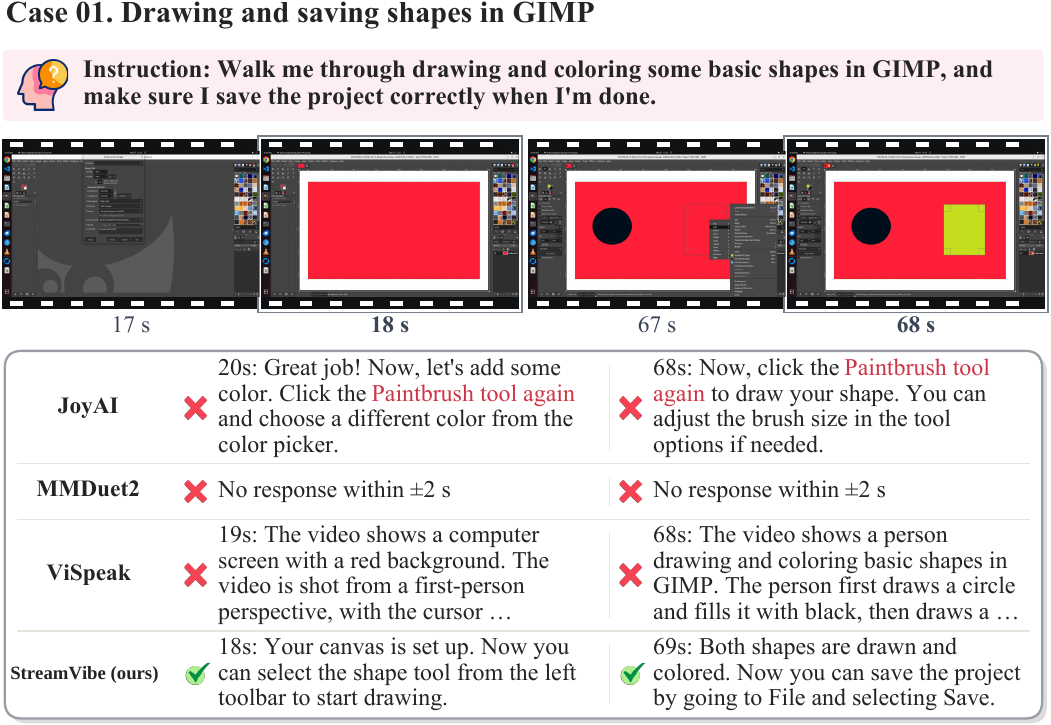}
\caption{\textbf{Additional case study 01.} Drawing and saving shapes in GIMP. Model responses at two annotated evidence points, matched within $\pm2$\,s.}
\label{fig:appendix-cases-01}
\end{figure}

\begin{figure}[!htbp]
\centering
\includegraphics[width=\linewidth,height=0.43\textheight,keepaspectratio]{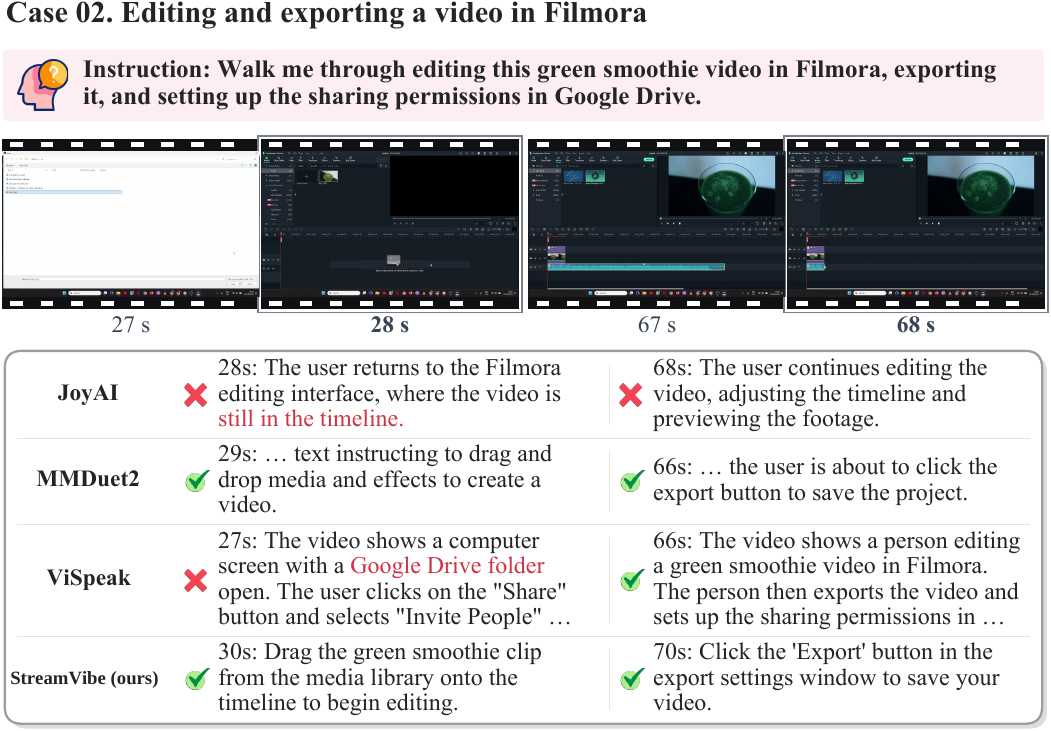}
\caption{\textbf{Additional case study 02.} Editing and exporting a video in Filmora. Model responses at two annotated evidence points, matched within $\pm2$\,s.}
\label{fig:appendix-cases-02}
\end{figure}

\begin{figure}[!htbp]
\centering
\includegraphics[width=\linewidth,height=0.43\textheight,keepaspectratio]{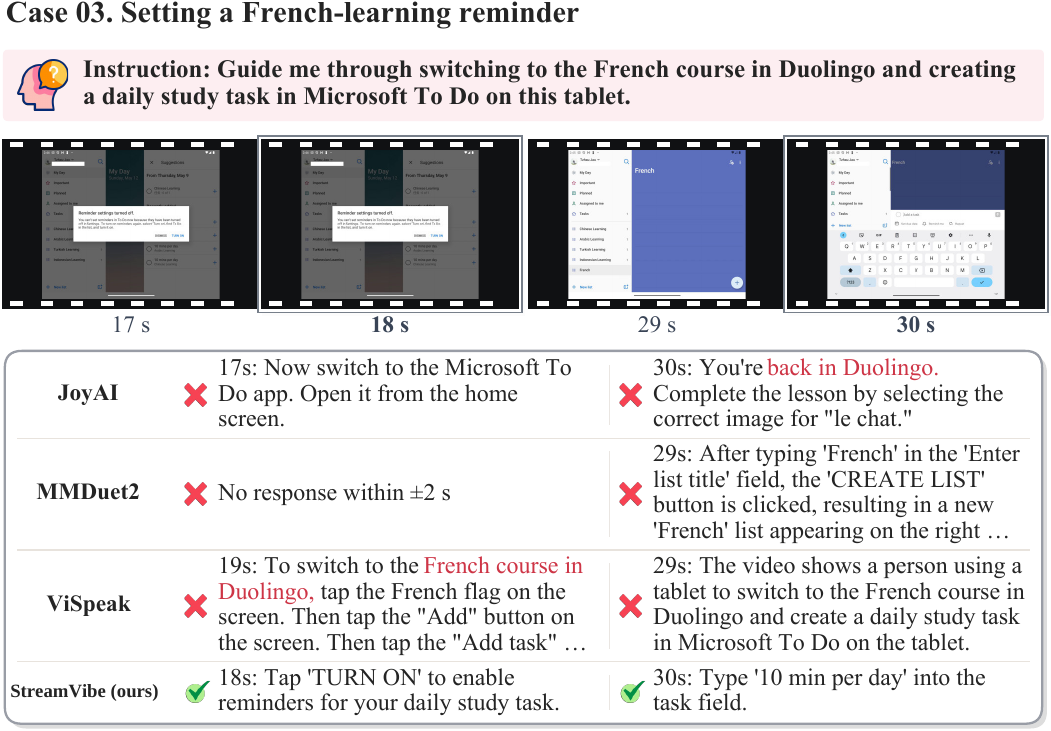}
\caption{\textbf{Additional case study 03.} Setting a French-learning reminder. Model responses at two annotated evidence points, matched within $\pm2$\,s.}
\label{fig:appendix-cases-03}
\end{figure}

\begin{figure}[!htbp]
\centering
\includegraphics[width=\linewidth,height=0.43\textheight,keepaspectratio]{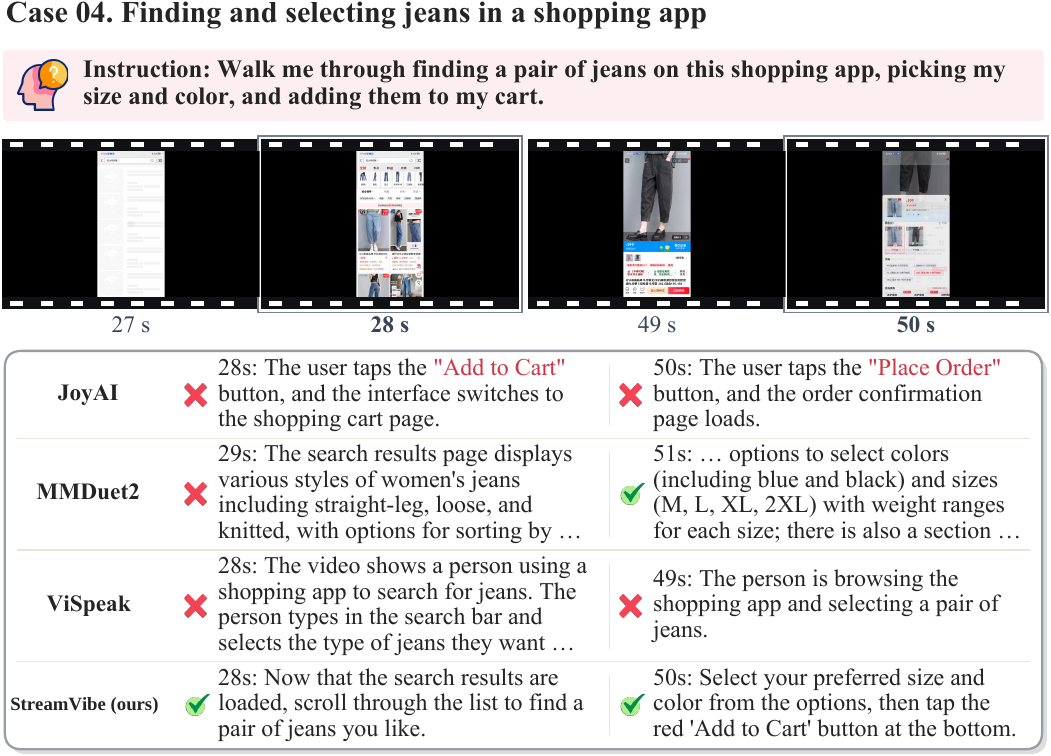}
\caption{\textbf{Additional case study 04.} Finding and selecting jeans in a shopping app. Model responses at two annotated evidence points, matched within $\pm2$\,s.}
\label{fig:appendix-cases-04}
\end{figure}

\begin{figure}[!htbp]
\centering
\includegraphics[width=\linewidth,height=0.43\textheight,keepaspectratio]{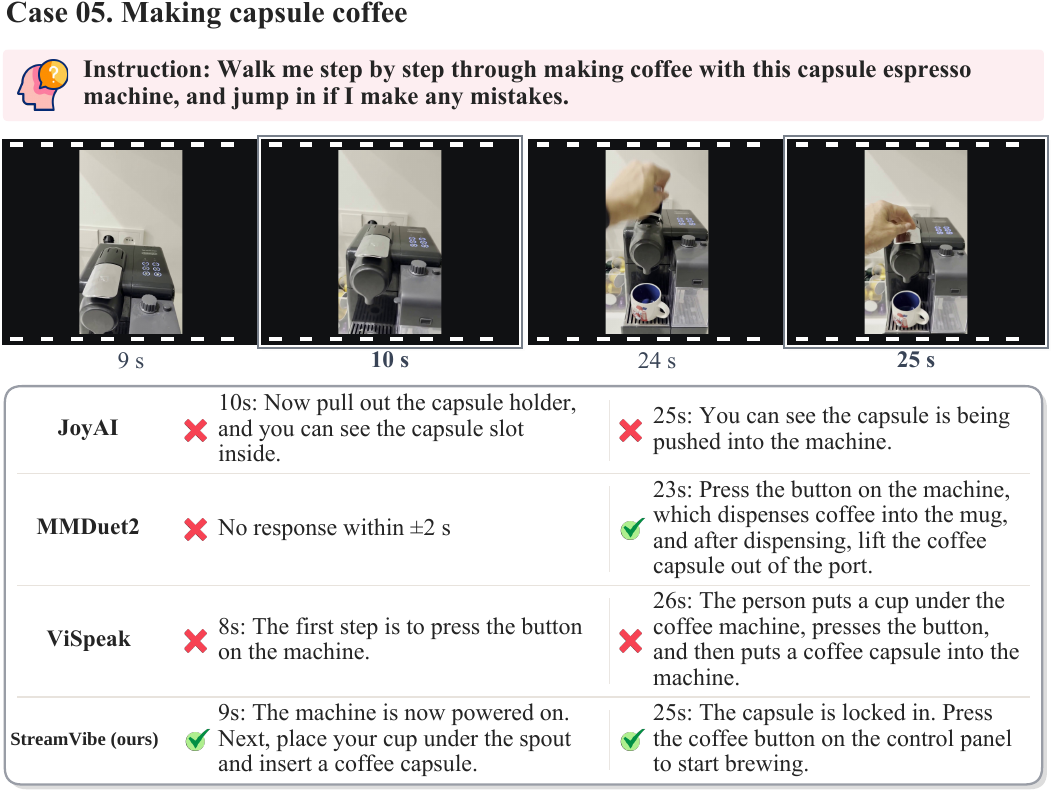}
\caption{\textbf{Additional case study 05.} Making capsule coffee. Model responses at two annotated evidence points, matched within $\pm2$\,s.}
\label{fig:appendix-cases-05}
\end{figure}

\begin{figure}[!htbp]
\centering
\includegraphics[width=\linewidth,height=0.43\textheight,keepaspectratio]{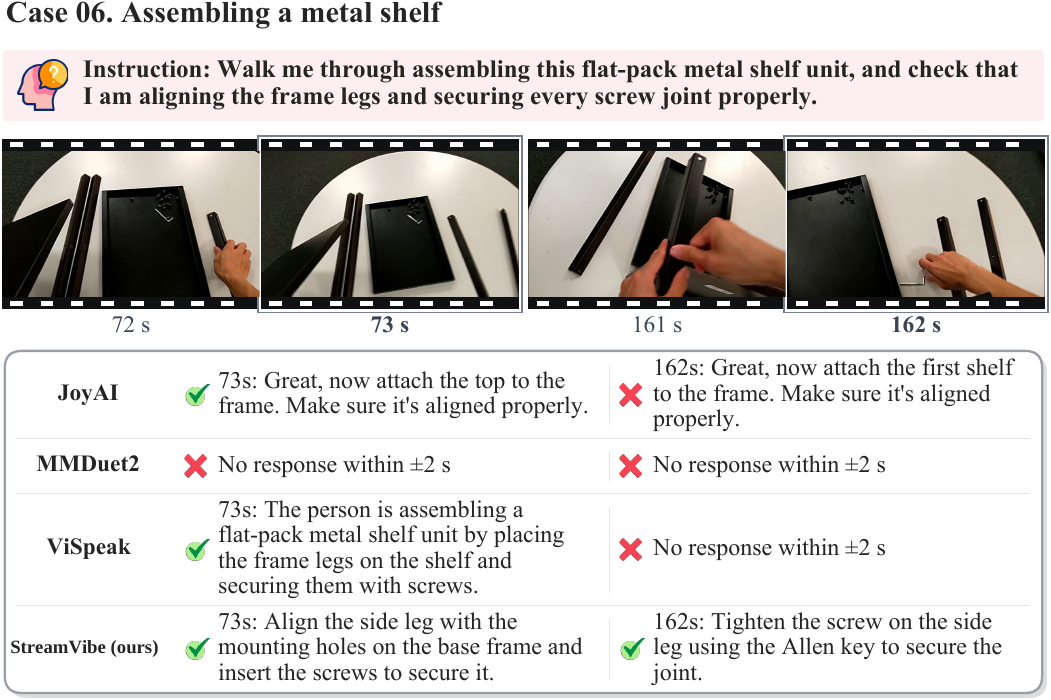}
\caption{\textbf{Additional case study 06.} Assembling a metal shelf. Model responses at two annotated evidence points, matched within $\pm2$\,s.}
\label{fig:appendix-cases-06}
\end{figure}

\begin{figure}[!htbp]
\centering
\includegraphics[width=\linewidth,height=0.43\textheight,keepaspectratio]{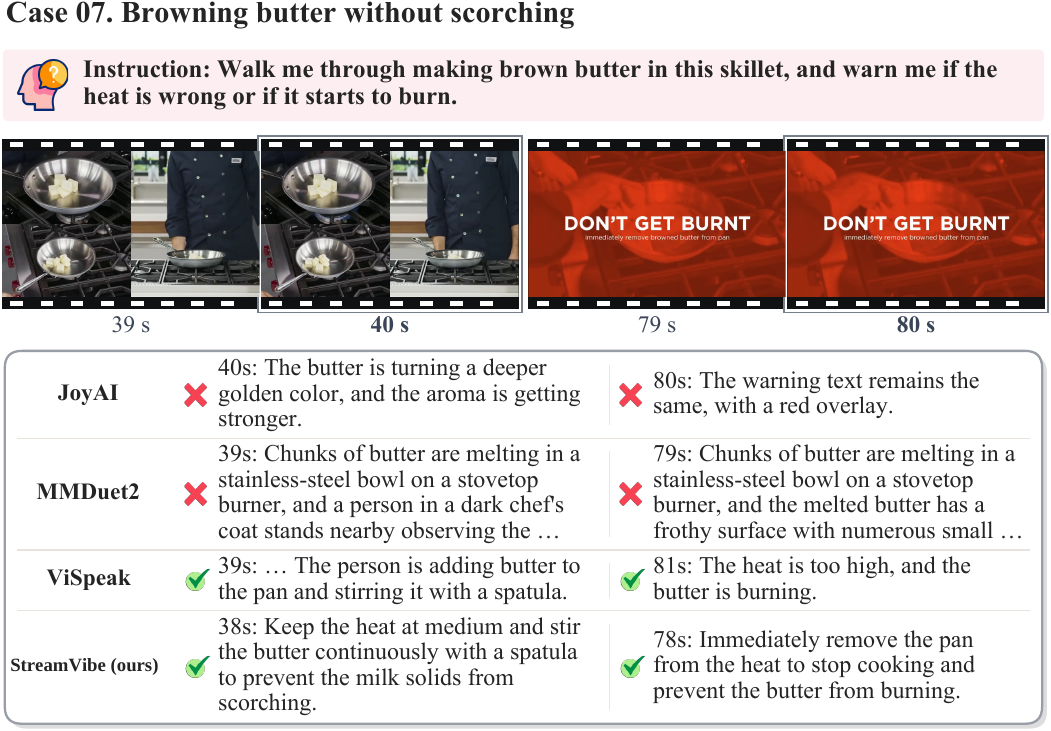}
\caption{\textbf{Additional case study 07.} Browning butter without scorching. Model responses at two annotated evidence points, matched within $\pm2$\,s.}
\label{fig:appendix-cases-07}
\end{figure}

\begin{figure}[!htbp]
\centering
\includegraphics[width=\linewidth,height=0.43\textheight,keepaspectratio]{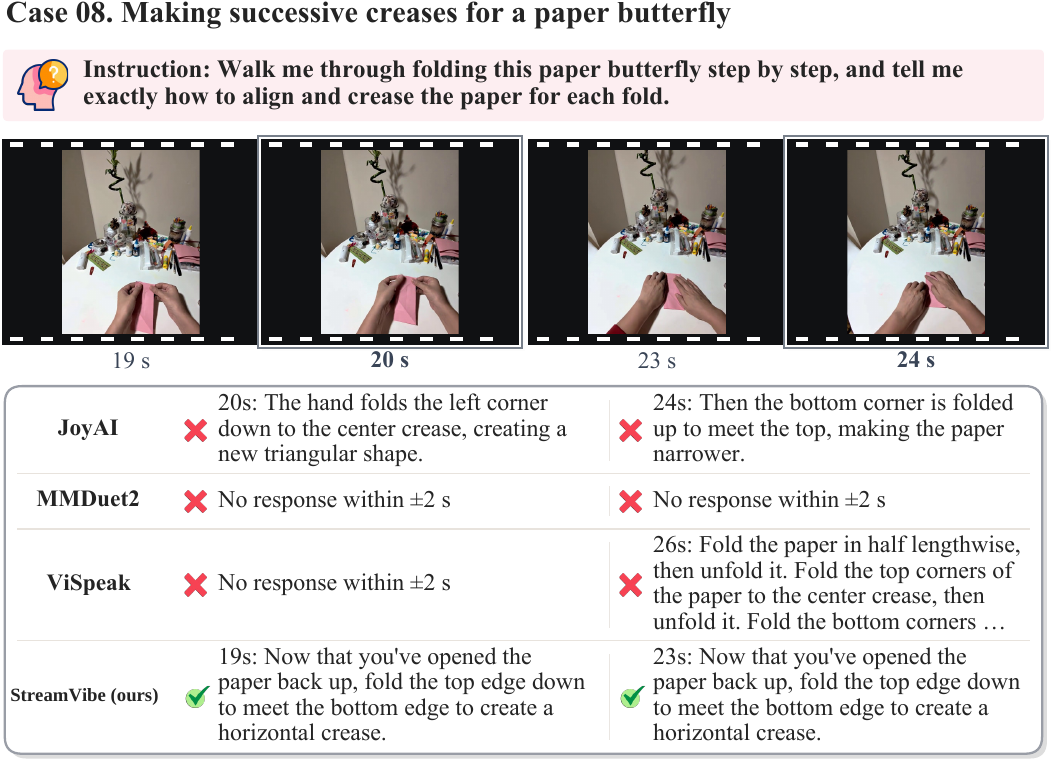}
\caption{\textbf{Additional case study 08.} Making successive creases for a paper butterfly. Model responses at two annotated evidence points, matched within $\pm2$\,s.}
\label{fig:appendix-cases-08}
\end{figure}

Figures~\ref{fig:appendix-cases-01}--\ref{fig:appendix-cases-08} compare responses during sustained interaction. StreamVibe uses task progress, interface state, and user goals to give timely, concrete guidance. Baselines more often respond too early or too late, miss key evidence, repeat earlier advice, or produce guidance inconsistent with the current interface state.

\end{document}

%% file: math_commands.tex
\usepackage{amsmath,amsfonts,amssymb,bm}

\def\eqref#1{equation~\ref{#1}}
\def\1{\bm{1}}

\DeclareMathAlphabet{\mathsfit}{\encodingdefault}{\sfdefault}{m}{sl}
\SetMathAlphabet{\mathsfit}{bold}{\encodingdefault}{\sfdefault}{bx}{n}